\documentclass[sigconf]{acmart}

\usepackage{graphicx,amsmath,multirow,bm,xspace,url}
\usepackage{amsthm}
\usepackage{boldline}
\newtheorem{theorem}{Theorem}
\newtheorem{definition}{Definition}
\usepackage{amsfonts}
\usepackage{fancyhdr}
\usepackage{soul}
\usepackage{balance}
\usepackage{array}
\usepackage{capt-of}
\usepackage{wrapfig}
\usepackage{textcomp, gensymb}
\usepackage{tabularx}
\newcolumntype{Y}{>{\centering\arraybackslash}X}
\usepackage{booktabs}
\usepackage{siunitx}
\usepackage{soul}
\usepackage{ifthen}
\usepackage[noend, linesnumbered]{algorithm2e}
\usepackage{hyphenat}
\usepackage[super]{nth}
\usepackage{subcaption}

\usepackage{breakurl}
\usepackage{etoolbox}
\usepackage{xurl}

\SetAlFnt{\fontfamily{ptm}}
\SetFuncSty{algtimes}
\SetCommentSty{commentfont}
\SetKwIF{If}{ElseIf}{Else}{if}{}{else if}{else}{end if}

\newcommand{\para}[1]{{\bf \noindent #1 \hspace{3pt}}}
\renewcommand{\paragraph}[1]{\para{#1}}
\newenvironment{packed_itemize}{
	\begin{list}{\labelitemi}{\leftmargin=1.em}
		\setlength{\itemsep}{1pt}
		\setlength{\parskip}{0pt}
		\setlength{\parsep}{0pt}
		\setlength{\headsep}{0pt}
		\setlength{\topskip}{0pt}
		\setlength{\topmargin}{0pt}
		\setlength{\topsep}{0pt}
		\setlength{\partopsep}{0pt}
	}{\end{list}}
\newenvironment{packed_enumerate}{
	\begin{enumerate}{\leftmargin=1.em}
		\setlength{\itemsep}{1pt}
		\setlength{\parskip}{0pt}
		\setlength{\parsep}{0pt}
		\setlength{\headsep}{0pt}
		\setlength{\topskip}{0pt}
		\setlength{\topmargin}{0pt}
		\setlength{\topsep}{0pt}
		\setlength{\partopsep}{0pt}
	}{\end{enumerate}}
\makeatletter
\newcommand{\removelatexerror}{\let\@latex@error\@gobble}
\makeatother

\makeatletter
\def\tagform@#1{\maketag@@@{(\ignorespaces{Eq.\ #1}\unskip\@@italiccorr)}}
\renewcommand{\eqref}[1]{\textup{{\normalfont(\ref{#1}}\normalfont)}}
\makeatother

\newcommand\name{{VeriLight}}

\newboolean{showComments}
\setboolean{showComments}{false}
\definecolor{aqua}{rgb}{0.0, 1.0, 1.0}
\definecolor{purple}{rgb}{0.49, 0, 1}
\definecolor{lightpurple}{rgb}{0.8, 0.6, 1}
\definecolor{Yellow}{rgb}{1.0, 1.0, 0}
\definecolor{green}{rgb}{0.089, 0.877, 0.434}

\ifthenelse{\boolean{showComments}}{

    \newcommand{\hs}[1]{{\color{purple} #1}}
    \newcommand{\htodo}[1]{\sethlcolor{lightpurple}\hl{\bf[Hadleigh TODO] #1}}
    
}{

    \newcommand{\xtodo}[1]{}
    \newcommand{\htodo}[1]{}
    \newcommand{\hs}[1]{}
    \newcommand{\xz}[1]{}   
    \newcommand{\cc}[1]{}
}

\AtBeginDocument{%
  }

\copyrightyear{2025}
\acmYear{2025}
\setcopyright{cc}
\setcctype{by-nc-nd}
\acmConference[CCS '25]{Proceedings of the 2025 ACM SIGSAC Conference on Computer and Communications Security}{October 13--17, 2025}{Taipei, Taiwan}
\acmBooktitle{Proceedings of the 2025 ACM SIGSAC Conference on Computer and Communications Security (CCS '25), October 13--17, 2025, Taipei, Taiwan}\acmDOI{10.1145/3719027.3765112}
\acmISBN{979-8-4007-1525-9/2025/10}

\begin{document}

\title{Combating Falsification of Speech Videos with Live Optical Signatures (Extended Version)} 

\author{Hadleigh Schwartz}
\orcid{0009-0001-6151-2240}
\affiliation{%
  \institution{Columbia University}
  \city{New York}
    \state{NY}
  \country{USA}
}

\author{Xiaofeng Yan}
\orcid{0009-0008-0722-7650}
\affiliation{%
  \institution{Columbia University}
  \city{New York}
  \state{NY}
  \country{USA}
}

\author{Charles J. Carver}
\orcid{0000-0002-6664-1893}
\affiliation{%
  \institution{Massachusetts Institute of Technology}
  \city{Cambridge}
  \state{MA}
  \country{USA}
}

\author{Xia Zhou}
\orcid{0000-0002-2852-9024}
\affiliation{%
  \institution{Columbia University}
  \city{New York}
    \state{NY}
  \country{USA}
}

\begin{abstract}
High-profile speech videos are prime targets for falsification, owing to their accessibility and influence. This work proposes \name, a low-overhead and unobtrusive system for protecting speech videos from visual manipulations of speaker identity and lip and facial motion. Unlike the predominant purely digital falsification detection methods, \name\ creates dynamic physical signatures at the event site and embeds them into all video recordings via imperceptible modulated light. These physical signatures encode semantically-meaningful features unique to the speech event, including the speaker's identity and facial motion, and are cryptographically-secured to prevent spoofing. The signatures can be extracted from any video downstream and validated against the portrayed speech content to check its integrity. Key elements of \name\ include (1) a framework for generating extremely compact (i.e., 150-bit), pose-invariant speech video features, based on locality-sensitive hashing; and (2) an optical modulation scheme that embeds >200 bps into video while remaining imperceptible both in video and live.
Experiments on extensive video datasets show \name\ achieves AUCs $\geq$ 0.99 and a true positive rate of 100\% in detecting falsified videos. Further, \name\ is highly robust across recording conditions, video post-processing techniques, and white-box adversarial attacks on its feature extraction methods. A demonstration of \name\ is available at {\texttt{\href{https://mobilex.cs.columbia.edu/verilight}{https://mobilex.cs.columbia.edu/verilight}}}.

 \end{abstract}



\maketitle

 \section{Introduction}
In the early days of video technology, high-profile speeches were some of the first events to be shared over the new communication medium~\cite{first_video_speeches}. Influential figures capitalized on its unique persuasive power~\cite{Donovan_Scherer_2005}, and  video has since been a staple of information exchange. Today, this exchange faces a flood of falsified videos of high-profile speeches created to spread disinformation and discord.

This paper focuses on combating falsification of two salient aspects of a speech event: the speaker's identity, and their lip and face movements, which are directly tied to speech content. These elements are particularly persuasive and semantically-rich~\cite{sudar2021} and have been targeted in numerous incidents~\cite{nancy19,hadid23,swinney2024,biden23_draft,biden24_lipsync, harris24_slow, harris24_slur}. Today, realizing these falsifications is easier than ever. An attacker may use any open-source model or online deepfake tool to generate fictitious videos of their victim.
Even simple edits achievable on most smartphones, such as changing playback speed, can greatly alter a portrayal (e.g., widely-circulated videos claiming to show Nancy Pelosi and Kamala Harris delivering speeches intoxicated~\cite{nancy19, harris24_slow}). 
Once disseminated online, these videos 
create serious political, financial, and social risks. 
Unfortunately, current technologies for detecting falsified speech videos have failed to counter these threats.

\begin{figure}[t!]
    \centering
    \includegraphics[width=\columnwidth]{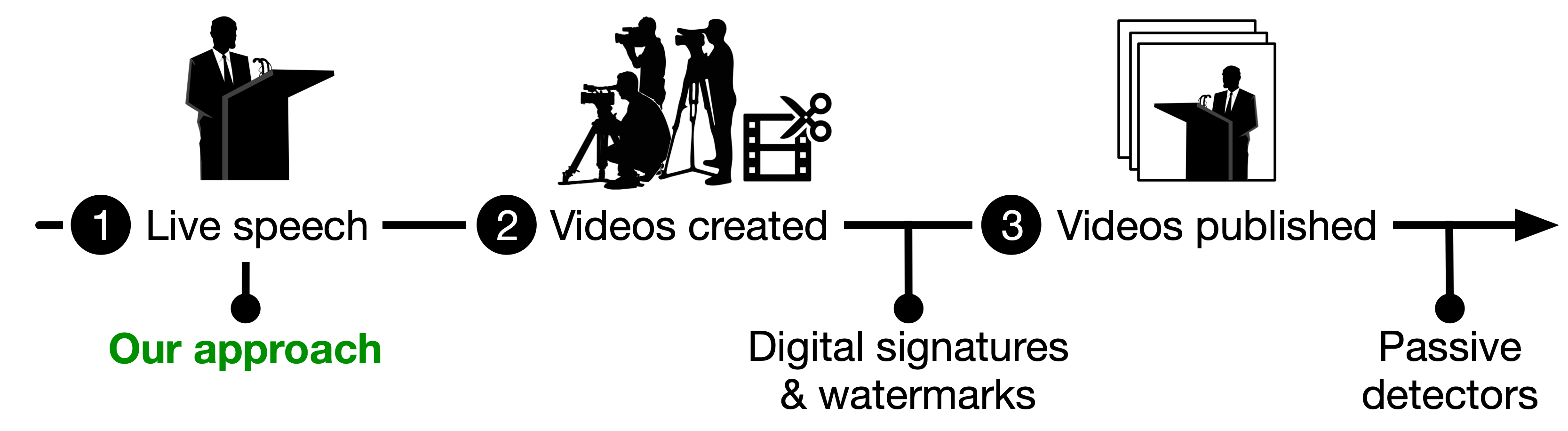}
    \vspace{-0.25in}
    \caption{We initiate protection of speech videos at the earliest possible stage of a video's lifetime: the physical scene of the speech.}
    \label{fig:related}
\end{figure}

Existing technologies can be characterized based on the stage of a video's lifetime they initiate protection (Figure~\ref{fig:related}). A majority are \emph{passive}, operating only \emph{after videos are published} by searching content for artifacts introduced during falsification~\cite{spotfornow,ciftci2020fakecatcher}. These visual flaws, however, are diminishing with advances in generative AI, and passive detectors are increasingly evaded~\cite{pulsedit, carlini2020evading}. 

In response, efforts have increasingly shifted to protecting videos at \emph{their digital creation}. Such methods either watermark generated content~\cite{synthid} or tag videos with digitally-signed credentials upon capture~\cite{truepic, liu24, liu2022vronicle, leica_cam}, generation, or editing~\cite{adobe_cai}. While promising, these approaches face several barriers stemming from their reliance on cooperation from all recording and editing parties. \textbf{(1)} Methods adding verification information to fake content are unlikely to be adopted by malicious actors. With deepfake and editing technologies increasingly democratized, attackers can easily create videos lacking watermarks or credentials and then claim they are real. \textbf{(2)} Capture-time approaches require the use of specialized apps~\cite{truepic} or hardware~\cite{liu24, liu2022vronicle,leica_cam}, which cannot be guaranteed in the large audiences of public speech events. \textbf{(3)} Because digital signatures are bound to a video's low-level representation (i.e., its pixel values), they must be regenerated when any post-processing techniques are applied. These techniques, such as compression and transcoding, are exceedingly common in video sharing workflows. Digital signature methods thus assume user cooperation at each stage of a video's lifetime. As evident today, without such uniform compliance, an indistinguishable mix of real and fake unsigned content is produced, creating competing narratives of the truth.

This work studies a complementary physical approach to protecting speech videos, seeking to shift protection agency from recording parties to speakers themselves. We envision speakers deploying a device that creates signatures \emph{physically} at the speech site, such that they are naturally embedded into \emph{all real} recordings. These physical signatures encode semantically-meaningful features (e.g., representations of speaker lip motion) that are unique to the event and consistent across recording device and position. Legitimate videos inherently pass validation against the signature, while falsified videos possess diverging features and are thus detected. Further, physical signatures are \emph{cryptographically-secured}, preventing their forgery. While digital signature methods require all parties to independently sign their videos, here, cryptographic data is generated only once, on the deployed physical signature creation device.\footnote{We use the term \emph{physical signature} to convey the use of secure, auxiliary data to verify content. Physical signatures are not, however, created in the same way as digital signatures.}

\begin{figure}[t]
    \centering
    \includegraphics[width=.8\columnwidth]{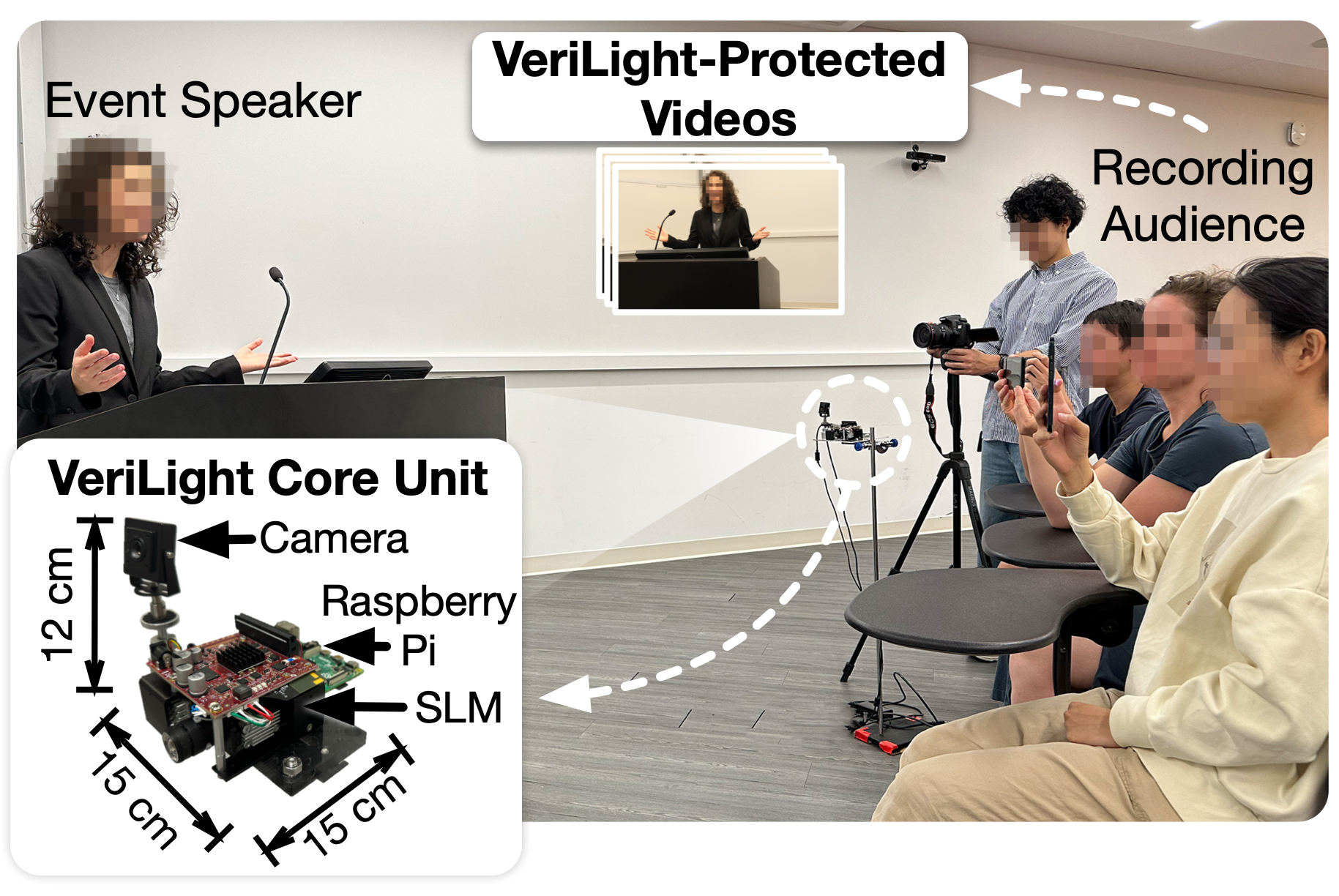}
    \vspace{-.1in}
    \caption{\name\ protecting a live speech. The low-cost core unit creates signatures encoding the event and embeds them into \emph{all} videos by projecting imperceptible light with a spatial light modulator (SLM).}
    \label{fig:teaser}
    \vspace{-.1in}
\end{figure} 

\begin{figure*}[t]
    \centering
    \vspace{-0.1in}
    \includegraphics[width=0.8\textwidth]{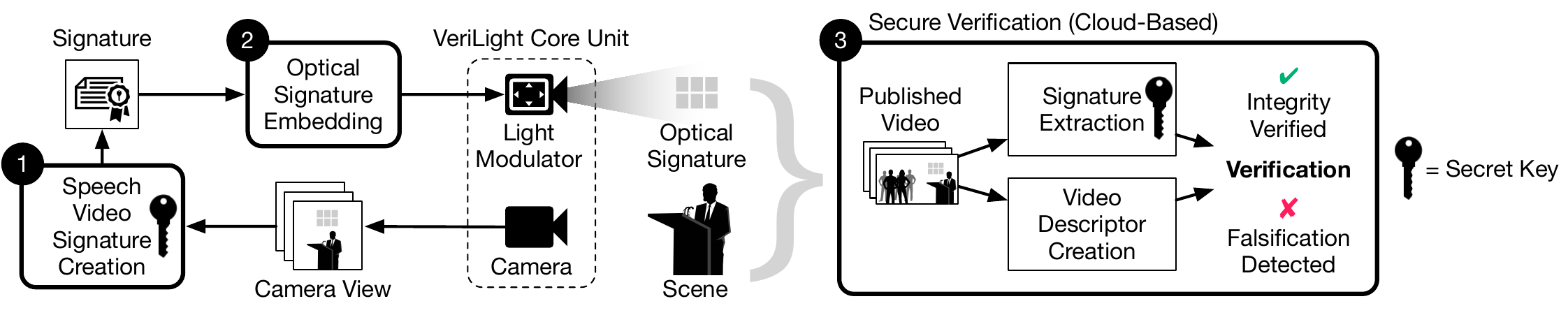}
    \vspace{-0.15in}
    \caption{Overview of \name's modules and workflow, including: (1) speech video signature creation; (2) optical signature embedding during the live speech; (3) verification of a later published video of the speech by the video verification module.}
    \label{fig:sysoverview}
\end{figure*}

Physical signatures provide several benefits in the context of live speeches.
\textbf{(1)} They inherently protect all videos at the event without demanding device cooperation, thus creating a canonical version of the speech reinforced by each recording attendee.
\textbf{(2)} Since physical signatures capture higher-level features rather than low-level pixel values, they remain valid after benign edits that preserve the video's semantic content (e.g., compression).

This paper demonstrates \name, a physical signature platform that disseminates signatures via modulated light at the speech site (Figure \ref{fig:teaser}). A speaker places a low-cost \name\ core unit, serving as a trustworthy third-party witness, at the site. The core unit continually extracts semantically-meaningful and robust visual features specific to the speaker's identity and face and lip motion (referred to as a video descriptor), and uses its private key to generate a message authentication code (MAC). The descriptor and its MAC make up the signature. \name\ then encodes the signature data as optical modulations that remain \emph{imperceptible} both live and in videos, supporting the platform's broader adoption. These optical modulations nonetheless manifest in all recordings as decodable pixel-level signals. A published video can be verified at any point in its lifetime by extracting the optical signatures and comparing recovered descriptors to those computed on the portrayed speech. 

Two main technical challenges arise in realizing \name. \textbf{(1)} 
The limited frame rate of typical cameras results in a low embedding data capacity (hundreds of bits per second). Thus, descriptors must be highly compact yet highly expressive and consistent across camera positions. \textbf{(2)} Optical signature embedding must balance conflicting goals of imperceptibility, robustness, and data capacity.

We address these challenges with the following contributions. \textbf{(1)} We propose a framework based on locality-sensitive hashing (LSH) to compress pose-invariant, semantically-meaningful speech video feature vectors to just 150 bits while preserving their performance. This framework supports diverse feature vectors
and is independent of the signature dissemination modality. 
\textbf{(2)} We design a spatio-temporal light modulation scheme that boosts bandwidth across \emph{all} RGB cameras while remaining imperceptible and resilient against common video post-processing techniques. To the best of our knowledge, this is the first scheme for embedding invisible information into videos from within the environment, taking an important step towards practical physical signatures. \textbf{(3)} We fabricate a \name\ prototype and examine its performance on 257 minutes of speech video captured with our core unit deployed and over 1,300 pairs of real and deepfaked videos.
Additionally, we assess its imperceptibility and robustness in diverse recording scenarios. \textbf{(4)} We evaluate \name's robustness against extensive countermeasures, including two white-box adversarial attacks aiming to generate fake videos that evade detection.

We summarize our key findings below:
\begin{packed_itemize}
\item \name's LSH framework supports over 100-fold reduction in the representation size of generic speech video features while maintaining their verification performance.
\item \name\ attains Areas Under Curves (AUCs) $\geq$ 0.99 and a true positive rate of 100\% in detecting falsifications of speaker identity and face and lip motion. In challenging scenarios where as little as 1.35~s of a video is modified, \name\ achieves an AUC of $\geq$ 0.90, a 40\% gain over the best passive detector baseline evaluated.
\item \name's semantically-meaningful video descriptors are robust to varied countermeasures and white-box adversarial attacks.
\item \name\ supports recording with any RGB camera, at viewing angles up to 60$\degree$ and distances up to 3~m, even when videos are captured with no optical zoom.
\item \name\ achieves error-free signature data extraction and verification of videos recorded in extensive indoor and outdoor environments, as well as after common video post-processing methods such as compression, transcoding, and filter application.
\item User studies and perceptual metrics confirm optical signatures are imperceptible live and in video in varied deployment settings.
\end{packed_itemize}

\para{Artifact availability} Our code and guides on reproducing our hardware are available on Github~\cite{verilight_github} and Zenodo~\cite{verilight_zenodo}. %
\section{Background and Related Work}
\label{sec:related}

\subsection{Video Falsification}
\label{subsec:deepfake_creation}

This work combats visual falsifications of speaker identity and face and lip motion, which determine delivered content. Such falsifications can be made using traditional techniques like frame rate modification, trimming, or cropping. Increasingly, they are achieved via the following types of deepfakes. \emph{(1) Face reenactments} use a source video to drive the facial movements of a target image.
For example, an attacker may use frames from a real speech video as targets to create reenactment deepfakes wherein the speaker's lips are reanimated to match fabricated audio (e.g., ~\cite{biden24_lipsync}).  
\emph{(2) Identity swaps} 
replace a victim's face with that of another identity while retaining the victim's facial motion and expression;
\emph{(3) Complete face synthesis} generates fictitious faces (i.e., non-existent identities). In our case, this form of fake has the same end effect as an identity swap deepfake, as it changes the perceived identity of the speaker.

\subsection{Preventing and Detecting Fake Videos}
We organize existing techniques for preventing and detecting falsified videos into four categories as below. We discuss mechanisms exclusively targeting AI-generated videos (i.e., deepfakes), as well as broader media authenticity initiatives.

\para{Passive detectors} These techniques analyze videos for evidence of tampering and typically are designed to detect deepfakes. They hone in on high-level physical inconsistencies \cite{li2018ictu,hu2021exposing,ciftci2020fakecatcher, qi2020deeprhythm,agarwal2021auraldynamics,mittal2020emotions,agarwal2020phoneviseme, chugh2020not, haliassos2021lips, zhou2021joint} (e.g., unnatural lip movements), biometric incongruities~\cite{agarwal2023watch, agarwal2020appearancebehavior, agarwal2019protecting, cozzolino2021id} (e.g., lack of identity-specific head movements), or pixel-level artifacts learned by models (often convolutional neural networks) trained on deepfakes~\cite{spatial_freq_det, artifact_det_and_sim,blending_artifacts,lugstein2021prnu,  luo2021generalizing, liu2021spatial, yan2023ucf, xu2023tall, Wang_2023_CVPR,spotfornow, afchar2018mesonet, rossler2019faceforensics++, multiattentional_nn, chen2021defakehop,bonettini2021video, masi2020two, cao2022end,nguyen2019capsule}.

Unfortunately, passive detectors are increasingly evaded and even hijacked to improve the quality of fake videos. For instance,~\cite{pulsedit} bypasses remote photoplethysmography (rPPG)-based detection methods~\cite{ciftci2020fakecatcher} by generating faces with realistic rPPG signals. Adversarial attacks have conquered several passive detectors~\cite{carlini2020evading}. 

\para{Digital signing and watermarking} These active techniques add information to digital content at its creation to enable immediate or downstream verification. Emerging frameworks append cryptographically-signed provenance metadata to files at recording time~\cite{liu24, truepic, leica_cam} or upon editing~\cite{liu2022vronicle, adobe_cai}.
Digital watermarks directly embed verification signals into real~\cite{neekhara2022facesigns, qureshi2021detecting, wang2021faketagger} or synthetic~\cite{synthid} media. Overall, these digital methods require the cooperation of all video sources throughout the information ecosystem, since they add data on a per-video level. Unfortunately, adversaries are deincentivized from participating in these frameworks and can find alternatives to create synthetic media lacking such data.

\para{Live QR codes} Two prior works display dynamic QR codes by a speaker to disseminate verification information. Both only address falsification of speech audio. Critch~\cite{critch2022wordsig} proposes to display QR codes that encode a speech's transcript but does not evaluate the concept. In~\cite{shahid2023my}, time-frequency features of audio signals are encoded. Because of these features' lower-level nature, they lack robustness across key recording conditions (e.g., their accuracy is below 90\% at distances beyond \SI{2}{ft} and in the presence of ambient noise) and are not shown to be pose-invariant or adversarially robust. 
Our work instead focuses on visual falsifications and proposes a generic methodology for compressing pose-invariant, semantically-meaningful features. Our features are robust against diverse ambient conditions and evaluated adversarial attacks.

Independent of the verification features they carry, QR codes impose a strict tradeoff between obtrusiveness and supported camera positions. Prior works show a QR code must be over 20 x 20 cm in size to ensure it is readable by all devices within  \SI{3}{m} and 45\degree~\cite{caria2019exploring}. While using optical zoom can boost recording range, this simply fills more of the view with the QR code. Thus, QR code-based systems force speakers to accept either reduced protection robustness or large flickering patterns at the scene -- a critical barrier to adoption.

\para{Light-based methods}
A concurrent work~\cite{michael2025noise} combats video falsification by modulating scene illumination with secret, pseudo-random signals.
By correlating video with these signals, the authors generate images that visualize tampering. These visualizations have the advantage of capturing a broad range of spatial and temporal manipulations, but they must be manually assessed 
by a knowledgeable analyst. Further, because the encoded signals are not specific to scene content, the method is susceptible to replay attacks, wherein valid signals from real videos may be copied onto fakes. \name\ prevents such attacks by encoding cryptographically-secure, adversarially robust, and speech-specific features in modulations. It automates video analysis and outputs concrete integrity decisions.

Another related line of work employs active illumination from screens to verify that video chat participants are real~\cite{gerstner2022detecting, shangrealtime, livescreen, facerevelio}. These systems follow a challenge-response model, where facial appearance is analyzed with respect to dynamic screen light. They are thus constrained to video chat scenarios.

\section{Preliminaries}
\label{sec:pre}

\begin{figure*}[t]
    \centering
     \begin{subfigure}{0.7\textwidth}
        \includegraphics[width=\textwidth]{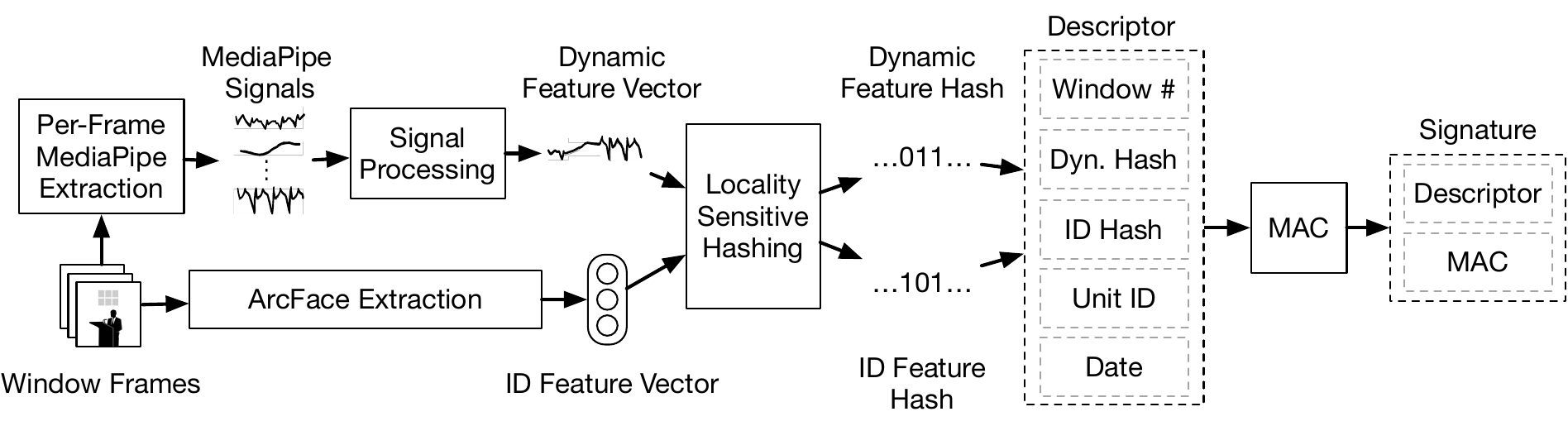}
        \caption{Generating a signature for a window of video frames. }
        \label{fig:digest}
    \end{subfigure}
    \hfill
     \begin{subfigure}{0.28\textwidth}
        \includegraphics[width=\textwidth]{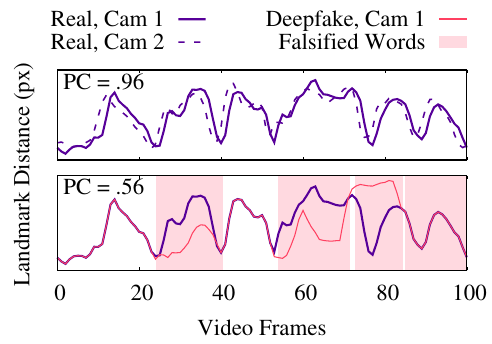}
        \caption{Features from real vs. falsified videos.}
        \label{fig:mediapipe_signals}
    \end{subfigure}
    \vspace{-0.15in}
    \caption{Assembly of \name\ signatures. (a) The signature consists of feature vectors hashed via LSH, plus a MAC. (b) A FaceMesh signal from three speech videos: two real videos from cameras at different yaws, and a reenactment deepfake changing four speech words. The Pearson correlation (PC) between videos' signals reflects the similarity of their speech content.}
\end{figure*}

\name\ is a physical and proactive approach to protecting live speech videos from visual falsification of speaker identity and lip and face movements, which reflect delivered content.
To match the visual nature of these falsifications, we embed physical signatures with light -- the sensing modality of vision. In this section, we describe \name's design and threat model, and then discuss the technical challenges it overcomes.

\subsection{System Overview}
\label{subsec:sys_overview}

Figure~\ref{fig:sysoverview} overviews the \name\ design, comprising three modules. The \name\ core unit, deployed at the site of a live speech, consists of a camera observing the event and a spatial light modulator positioned to project light onto any approximately planar surface in the immediate vicinity of the speaker (e.g., a small portion of a podium, wall, backdrop, or curtain). Notably, each of the speech environments portrayed in recent high-profile falsification incidents~\cite{nancy19,hadid23,swinney2024,biden23_draft,biden24_lipsync, harris24_slow, harris24_slur} possess such a surface. We  believe it is rare to find a high-profile speech event where this is not the case.

For each window of camera video frames, the signature creation module running on the core unit extracts a descriptor from frames. The descriptor contains semantically-meaningful features capturing speaker identity and lip and face motion, and additional provenance metadata such as a window timestamp. The core unit then generates a message authentication code (MAC) for the descriptor with its secret key. A descriptor and its MAC comprise a signature. The optical signature embedding module encodes the signature data as modulated light projected into the scene in the subsequent window of time.
This naturally embeds signatures into real recordings.

To verify a published video of a speech, the verification module extracts optical signatures from video frames and validates the MACs to confirm descriptors' integrity. It then examines whether the semantically-meaningful features recovered from the descriptor match those of the portrayed speech. To ensure that the secret key used to validate MACs is secured, we envision the verification module as a secure cloud service queried by users and media platforms. As such, secret keys will never be exposed to third-party devices. We elaborate on \name's use of secret keys and MACs in \S\ref{subsec:encryption}.

\subsection{Threat Model}
\label{subsec:threat}
Our threat model focuses on three entities: video producers, video verifiers, and attackers. Our focus is preventing attacks wherein falsified videos purport to be real. We do not address the inverse, wherein an attacker claims real content is fake.

\textbf{Video producers} create and disseminate \emph{legitimate} videos of the speech. This group includes viewers recording at the event and non-malicious parties re-distributing videos. Audience members can record using any RGB camera with a frame rate $\geq$ 24 FPS and resolution $\geq$ 1080p. We assume that the speaker’s face is visible in recordings at a maximum viewing angle of 60$\degree$ and distance of 3~m. Videos can be saved using common codecs (e.g., H.265, MPEG-4)
and post-processed with compression, transcoding, and filter application. 
We assume cameras remain still throughout the speech
but discuss reasonable solutions to avoid this requirement in ~\S\ref{sec:limitations_future}.

\textbf{Video verifiers} are individuals or media platforms 
who seek to confirm a video's integrity by passing it to the verification module.

\textbf{Attackers} disseminate falsified videos claiming to portray the speech event. We assume they possess white-box knowledge of \name\ and significant computational power, and carry out attacks \emph{after} a speech has taken place in an attempt to spread fake videos or undermine \name\ verification. We counter injection of interfering light at the speech site (\S\ref{subsec:other_countermeasures}), but further on-site attacks, such as tampering of \name's hardware, are out of scope. Concretely, attackers can perform any combination of the following:
\begin{packed_enumerate}
    \vspace{-.05in}
    \item Falsify the speaker's lip and face motion and/or identity via visual edits (optionally via joint audio manipulation), using traditional techniques or deepfake models.
    \item Access all \name\ algorithms and models, including weights.
    \item Create arbitrary completely synthetic speech videos.
    \item Incorporate valid signatures from other \name-protected videos into generated fake videos, i.e., replay attack.
    \item Modify video speed or re-order legitimate video frames.
    \item Remove or manipulate a video's embedded signatures.
    \item Digitally add to any video pixel-level signals that mimic the optical embedding scheme.
    \vspace{-.05in}
\end{packed_enumerate}
We assume that our private key is securely held out of reach of the adversary. Attacks on cryptographic primitives are out of scope.

\para{Real-world limitations} 
\name\ is but one technical approach to achieving physical signatures and is not without limitations. \textbf{(1)} It requires deployment of the \name\ core unit at speeches. \textbf{(2)} Because physical signatures are based upon a set of extracted event features, they do not address falsification of features outside this set.
We develop a prototype leveraging visual features of speaker identity and lip and face motion. The prototype thus does not protect against falsifications of facial attributes (e.g., makeup) or non-facial elements (e.g., clothing, background). \name's flexible LSH framework supports rich feature sets, but these features must be preselected based on anticipated attacks. \textbf{(3)} Videos must contain the full optical signature projection region to enable verification; as with digital signatures, videos lacking intact optical signatures are viewed as untrustworthy. Using a projection region close to the speaker's face or even configuring a small but visible projection border as a cue for filmers can help ensure signature inclusion.

While we believe that several of the aforementioned limitations can be addressed via further research (\S\ref{sec:limitations_future}), we ultimately view \name\ not as a panacea to fake speech videos, but rather a complement to passive detectors and digital signature methods. In particular, digital signature methods achieve provable security by requiring consistent user and recording-device cooperation to bind pixel values to credentials. \name\ and physical signatures more broadly trade provable security for flexibility, scalability, and a shift of protection agency (and efforts) from audience to speaker.

\subsection{Design Challenges}
\label{subsec:challenges}
\para{Compact and pose-invariant video descriptor} Video descriptors must be extracted in real-time and capture the speaker's identity and lip and facial motion.
Furthermore, they must be pose-invariant, so that any recording can be verified. Unfortunately, the optical embedding channel offers limited bandwidth, constraining the descriptor size. This is because a receiver's sampling rate must be at least double the modulation frequency (i.e., the Nyquist rate) for data to be decodable. Commodity cameras typically operate at
only 24-30 frames per second (FPS). Thus, achieving embedding bandwidths of even hundreds of bits per second (bps) is challenging. Existing speaker analysis methods output large features (e.g., hundred-dimension embeddings for single images~\cite{Wiles18a} or seconds of audio~\cite{wang2018attention}), significantly exceeding this limit.

\para{Signature embedding and extraction} Optical signature embedding and extraction contend with a tradeoff between robustness and imperceptibility. While large optical modulations bolster robustness by increasing the signal to noise ratio (SNR) of embedded signals, such fluctuations are highly perceptible at the scene and in recordings. Our modulation and extraction techniques must be both minimally obtrusive and robust to noise that may be introduced during capture and post-processing. Unlike in the digital 
realm where pixels can be directly modified to covertly and reliably embed information, we must \emph{anticipate} how light \emph{injected} into the scene will induce varying degrees of perceptibility and robustness.

We next present the design of \name's three modules, which jointly address the above challenges.

\section{Speech Video Signature Creation}
\label{sec:digest}

We propose a feature-agnostic framework for creating compact, pose-invariant speech video signatures in real-time (Figure \ref{fig:digest}). We build off existing computer vision tools to extract semantically-meaningful visual features crafted to address the attacks outlined in \S\ref{subsec:deepfake_creation}. Then, we use a technique known as locality-sensitive hashing (LSH) to compress the high-dimensional features to hundreds of bits (within the embedding data capacity) while preserving their verification functionality. Unlike cryptographic hash functions, which output highly different hashes for inputs with minor differences, LSH maps similar inputs to similar hashes. This enables \name\ to validate legitimate videos despite minor feature differences inevitably arising from recording condition and feature extraction variance. While prior works in video fingerprinting similarly apply LSH~\cite{yang2004hierarchical, khelifi2017perceptual}, our framework is unique in producing features that are simultaneously real-time, pose-invariant, and speech-specific.

\subsection{LSH-Based Descriptor Framework}
\label{subsec:lsh}
LSH is a technique for reducing data dimensionality while preserving approximate distances between data points. An LSH scheme consists of a function $H : \mathbb{R}^n \mapsto \{0, 1\}^k$ such that $D(H(\vec{u}), H(\vec{v}))$ estimates $sim(\vec{u}, \vec{v})$. $D$ is the Hamming distance, $sim$ is a similarity metric defined for $\vec{u}, \vec{v}$, and $k$ is a configurable hash size. An LSH scheme does not preserve the exact similarity of inputs but rather provides a probabilistic guarantee that similar inputs are mapped to similar hashes; using a larger hash size $k$ increases this probability.

We use the cosine similarity LSH scheme~\cite{charikar2002similarity}, denoted $H_{cos}$. $H_{cos}$ outputs $k$-bit hashes such that $D(H_{cos}(\vec{u}), H_{cos}(\vec{v}))$ estimates $\Theta(\vec{u}, \vec{v})$. Here, $\Theta(\vec{u}, \vec{v})$ is the angle between $\vec{u}$ and $\vec{v}$. 

The cosine similarity LSH scheme is appealing for our application for two reasons. \textbf{(1)} While larger hash sizes always improve the hash's accuracy in estimating cosine similarity, this relationship is \emph{independent} of the dimensionality of input vectors. This differs from Principle Component Analysis, where information loss is proportional to dimensionality reduction. 
In our use case, this means that our initial feature vectors can be arbitrarily high-dimensional, so long as their similarity is measured by cosine similarity.
We confirm this by deriving a closed-form equation for the effect of hash size on our features' performance (Theorem \ref{theorem:k_perf}).
\textbf{(2)} $H_{cos}$ can estimate Pearson correlation, a popular measure of time series similarity.
This is because the Pearson correlation of two time series is equivalent to their cosine similarity after zero-meaning.
This property is key to computing our dynamic features, which must capture temporal speech characteristics.

Equipped with $H_{cos}$, we can separately address the challenges of descriptor robustness and size constraint. Below, we describe our high-dimensional visual feature vectors, which are hashed to serve as the descriptor verification data.

\subsection{Semantically-Meaningful Video Descriptors}
\label{subsec:feature_extraction}
To address falsifications of speaker identity and lip and face motion, we extract two visual feature vectors: a biometric-based \emph{identity feature vector} and a temporal \emph{dynamic feature vector}. The LSH framework, however, supports varied features, as discussed in~\S\ref{sec:limitations_future}.

\para{Identity feature vector} The identity feature vector is used to verify the speaker identity in a published video to protect against identity swap falsifications. Neural network face embedding models are the gold-standard for extracting visual identity information. They map face images to vectors in a high-dimensional embedding space, where distance corresponds to face similarity. Conveniently, state-of-the-art face embedding models utilize cosine similarity as their distance metric. We employ a pre-trained ArcFace~\cite{deng2019arcface} model~\cite{arcface_impl}, which outputs a 512-dimensional vector. We pass ArcFace crops of the face obtained from a pre-trained face detector~\cite{ultra_light_face_detector}.

\para{Dynamic feature vector} 
The dynamic feature vector protects against falsifications of delivered content by ensuring that a speaker's face and lip motion have not been modified. %
We use MediaPipe FaceMesh~\cite{mediapipe}, a model for real-time face image analysis, to distill a window of video frames into a signal capturing 
spatio-temporal visual characteristics of the speaker. As shown in Figure~\ref{fig:mediapipe_signals}, the  similarity of two speech videos can be quantified as the Pearson correlation of their corresponding signals. We find that these simple signals strongly protect against varied falsifications (\S\ref{subsec:digest_robustness}). They also are more compact yet comparable in robustness to other speech features, as discussed in~\S\ref{sec:limitations_future}. 

Thus, given a window of $n$ video frames, we run FaceMesh on each frame to obtain its 52 blendshape scores -- pose-invariant coefficients representing facial expressions -- and 478 facial landmarks, which we align to a canonical view for pose-invariance. We find that FaceMesh produces accurate output for frames captured up to $60\degree$ off-axis from the speaker (\S\ref{subsec:digest_robustness}). 
We concatenate the values of 5 distances between landmarks around the lips and 11 blendshapes into separate $n$-sample signals, which together capture global facial motion and fine-grained lip motion. We identify this set of features as optimal via forward sequential feature selection~\cite{FERRI1994403} on all blendshape and distance signals, using a comprehensive multi-camera dataset (\S\ref{subsec:digest_robustness}). Finally, we smooth and standardize all signals and concatenate them into our $16n$-dimensional feature vector.

\subsection{MAC Generation and Key Management}
\label{subsec:encryption}
Our descriptor consists of both feature hashes, a window number, a core unit identifier, and a creation date (Figure~\ref{fig:digest}). The core unit generates a HMAC-SHA256 MAC for each descriptor using its secret key, ensuring the integrity and authenticity of embedded data. We refer to a descriptor and its MAC as a speech video signature.

We secure our data via MACs as opposed to public key encryption because public key schemes produce ciphertexts exceeding our embedding bandwidth.\footnote{For equivalent authenticity and integrity guarantees, RSA-based digital signatures are over 10 times larger than any HMAC-SHA MACs.} We discuss approaches to increasing bandwidth to facilitate public key encryption in \S\ref{sec:limitations_future}.

To create and validate MACs, \name\ must establish a secret key shared by core units and the verification service. This can be done using Diffie-Hellman key exchange~\cite{diffie2022new}, a scheme enabling two parties to generate a shared secret key over an insecure channel using their own public-private key pairs. \name\ may employ Diffie-Hellman in one of two ways. In the first, it may require each core unit owner to use their own public-private key pair to participate in secret key exchange with the cloud-based verification service (\S\ref{subsec:sys_overview}). During this process, \name\ can additionally authenticate the owner by validating their public key's digital certificate~\cite{cert_pki}. This process creates a unique secret key for each core unit, securely associated with a unit's identifier and owner. In the second approach, \name\ may maintain the same key across all core units, refreshing as needed. Our prototype assumes a secret key has already been initialized in one of these manners, since the involved key exchange and certificate technologies are well-established.

\section{Optical Signature Embedding}
\label{sec:encoding}
After obtaining the signature for a window, the optical signature embedding module projects light encoding the signature data into the scene. 
Prior works most relevant to this task explore light-based~\cite{rollinglight,  cotting2004embedding} or screen-camera~\cite{disco, unseencode, tera, vrcode, chromacode, inframe++, kazutake_opticalwatermarking3, hilight} communication. 
They achieve imperceptibility at the scene while maximizing the visibility of modulated light in video for \emph{real-time} decoding. These methods are inapplicable in our case, wherein we seek imperceptibility both live and in video, and decoding is performed downstream on videos rather than at capture time. Further, several
are only compatible with rolling-shutter cameras~\cite{disco, rollinglight, vrcode}.

To address these issues, we propose three design elements, illustrated in Figure~\ref{fig:embedding_overview}. \textbf{(1)} We modulate light spatially and temporally to boost embedding bandwidth. The temporal modulation operates at low frequencies (e.g., 3-6 Hz) to accommodate commodity cameras frame rates and all shutter modes. We leverage an amplitude-modulating spatial light modulator (SLM) -- an optical device that controls the intensity of emitted light in both space and time -- to introduce small amounts of carefully-crafted light onto a planar surface in the immediate vicinity of the speaker. 
Our design maintains imperceptibility both live and in videos by exploiting the human visual system's low sensitivity to small fluctuations in light intensity occurring in small regions~\cite{colordiscriminationvsize} and for short durations~\cite{colordiscriminationvtime}. \textbf{(2)} We apply concatenated error correction coding to the signature data to ensure its reliable recovery from videos and enhance its resilience against video post-processing.
\textbf{(3)} We develop an adaptive embedding algorithm which continually tunes the emitted light to respond to environmental changes and balance embedding imperceptibility and robustness.

\subsection{Concatenated Error Correcting Code}
\label{subsec:code}
Our concatenated error-correcting code~\cite{chromacode} consists of two simpler codes: an outer Reed-Solomon (RS) code and an inner convolutional code.  
Raw signature data first goes through the RS coder, which adds $n - k$ parity bytes to the $k$-byte signature to form an $n$-byte codeword. RS can
correct  errors in up to $\lfloor{(n-k)/2}\rfloor$ bytes in a codeword. The codeword is then passed through a convolutional coder, yielding our final coded data. To later recover the RS codeword, we perform soft decision Viterbi decoding. The soft decoder
takes each bit's distance to 1 or 0 to compute the corrected sequence. This aids in decoding signals that are consistently noisy. Unlike RS codes, convolutional code correction strength is dependent on error positions and can thus be unstable. 
Thus the two codes complement each other to greatly improve embedding robustness. The soft decoding corrects a majority of errors. The RS code guarantees to address all remaining errors up to its correction strength.

\subsection{Spatio-Temporal Light Modulation}
\label{subsec:modulation}

Coded data is translated into a series of bitmaps, which are projected in sequence by an SLM equipped with red, green, and blue LEDs.
The SLM accepts RGB bitmaps, where pixel values in a color channel are proportional to LED intensities. Thus, bitmap values determine the composition of light hitting surface regions, which in turn determines the regions' appearance live and pixel values in videos.

\begin{figure}[t]
    \centering
    \includegraphics[width=0.9\columnwidth]{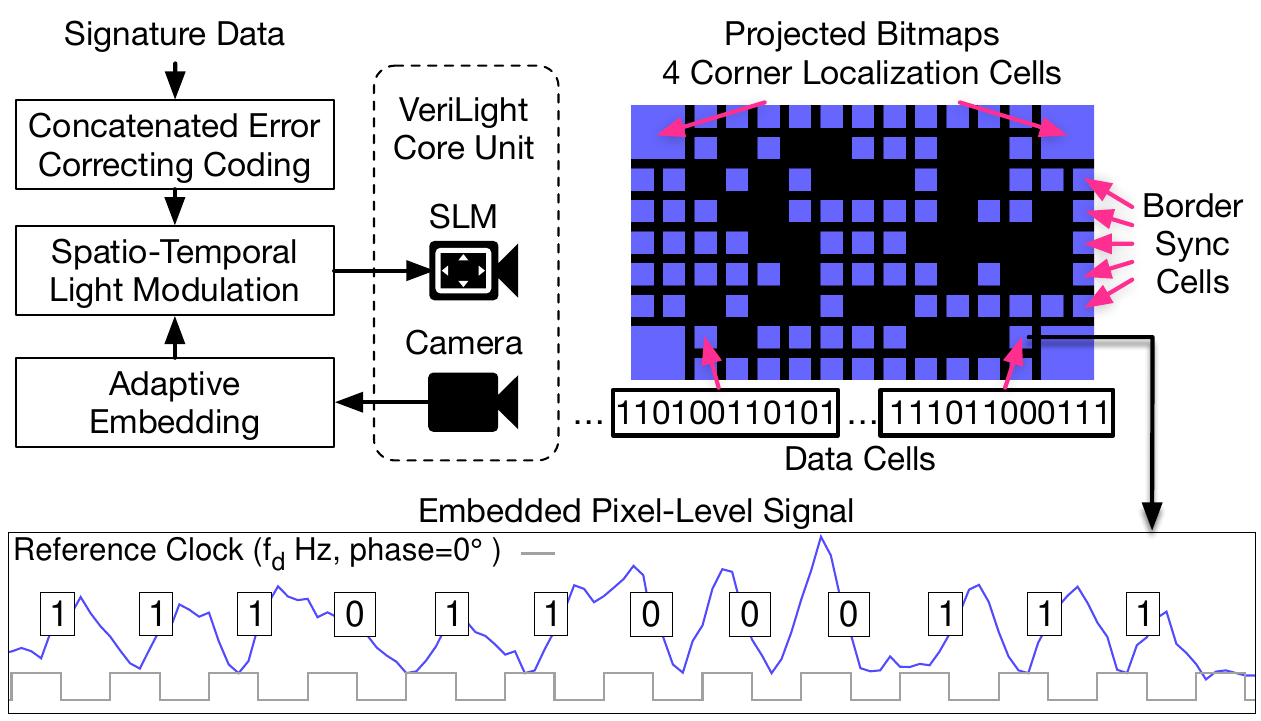}
    \vspace{-.1in}
    \caption{(Top) Overview of optical signature embedding, alongside a bitmap used by our modulation scheme. (Bottom) Resulting signal in a video, belonging to the last data cell. The reference clock, equivalent to our synchronization signal, is used to illustrate our BPSK scheme.}
    \label{fig:embedding_overview}
    \vspace{-.1in}
\end{figure}

We propose a spatially multiplexed modulation scheme. We divide each bitmap into a set of \emph{cells} (blocks of pixels), as shown in Figure \ref{fig:embedding_overview}. Each cell is independently modulated in time to produce optical signals. In a given bitmap, a cell $j$ can be either "on" or "off." When off, its color is set to black ($\text{RGB}(0, 0, 0)$), corresponding to zero emitted light. When on, its color is set to $c_{SLM}^j = \text{RGB} (R^j_{SLM}, G^j_{SLM}, B^j_{SLM})$, as determined in \S\ref{sec:adaptive_embedding}. 
We employ three types of cells, each with specific modulation behaviors.

\para{(1) Data cells} Most cells are data cells, modulated using Binary Phase Shift Keying (BPSK) to carry the raw bits of the signature. BPSK encodes bits by shifting the phase of a carrier signal between two states:  $0^{\circ}$ and  $180^{\circ}$. In our case, each bit is encoded via the display of a cell in two consecutive bitmaps. A 0 is conveyed by a cell value of $c_{SLM}^j$ followed by RGB(0,0,0) (i.e., phase of $0^{\circ}$), and vice versa for a 1. 
We employ BPSK for its balance of data rate and robustness to noise.
Data cells are modulated at $f_d$~Hz by displaying bitmaps at a refresh rate of $2f_d$~Hz. Thus, each data cell carries $m * f_d$ bits for a $m$~s modulation time. To embed $b$ bits, we assign contiguous chunks of $m * f_d$ bits to each data cell. 

\para{(2) Synchronization cells} Synchronization cells make up the border of the bitmap and facilitate demodulation of data cells. They are consistently modulated at the data frequency $f_d$~Hz with a phase of $0^{\circ}$. This provides the necessary reference signal for BPSK-demodulation of the  data cells. Synchronization cells are also used to determine the start and end of each window, as further described in \S\ref{sec:verification}. For an embedding window of $n$ seconds, we modulate all types of cells for $m (< n)$ seconds and leave the remaining $n- m $ as a downtime to facilitate determining the start of each window. 

\para{(3) Localization cells} The corner cells of a bitmap are localization cells. They are consistently modulated at $f_l$~Hz as a beacon of the optical signature's presence and location in recordings, critical for verification.
Their larger size and distinctive frequency distinguish them from other cells, supporting downstream localization (\S\ref{sec:verification}).

\begin{figure}[t]
\removelatexerror
\begin{algorithm}[H]
\caption{Adaptation performed after completion of a window to set upcoming SLM bitmap cell colors and intensities. $\phi_{max}$ and $\beta_{max}$ configure the trade-off between robustness and imperceptibility. $\delta$ controls the adaptation speed. A higher $\delta$ value corresponds to a more aggressive response to BER spikes.}
\label{alg:adaptation}
\small
\DontPrintSemicolon
\SetKwFunction{main}{Adapt}
\SetKwFunction{ber}{\bf BER}
\SetKwFunction{perc}{\bf CIEDE2K}
\SetKwFunction{colorsel}{\bf Equation1}
\SetKwProg{Fn}{Function}{:}{}
\KwIn{$V$: Window's core unit video frames \newline $I$ : required intensity for each SLM bitmap cell $j$}
\KwOut{$c_{SLM}$ : RGB value for each SLM bitmap cell $j$ \newline  $I$ :  updated required intensity for SLM cells.}
\Fn{\main{$V, I$}}{ 
    Let $\Phi_{max}$ be the perceptibility threshold\;
    Let $\beta_{max}$ be the BER threshold\;
    Let $\delta$ be intensity increment/decrement value\;
    $d \gets$ Extract data embedded in $V$\;
    $\beta \gets$ \ber{$d$}\; 
    \If{$\beta \geq \beta_{max}$}{
        $I \gets$ Increment all $I^j$ by $\delta$\; \label{glob_inc}
    }
    \ForEach{cell $j$}{
        $c^j_{p_{on}}, c^j_{p_{off}} \gets$ Patch $j$ color w/ and w/o SLM light\;
        \If{$\beta < \beta_{max}$ and \perc{$c^j_{p_{on}}, c^j_{p_{off}}$} $\geq \Phi_{max}$}{
            $I^j \gets I^j - \delta$\;
        }
        $c^j_{SLM} \gets $ \colorsel{$c^j_{p_{off}}, I^j$}\;   \label{color_eq} 
    }   
    \Return $c_{SLM}, I$\;
}
\end{algorithm}
\end{figure}

\subsection{Adaptive Embedding}
\label{sec:adaptive_embedding}

The key idea of adaptive embedding is to set each bitmap cell $j$'s color to optimize the illumination of its corresponding patch $j$ on the projection surface. While a cell's SNR is strictly determined by emitted light intensity (i.e., the sum of RGB channel emissions), its perceptibility is also influenced by its color (i.e., the light's relative channel values). For a given intensity, light is most imperceptible when its color matches that of a patch in the absence of SLM light.

Thus, we propose an intensity-guided adaptive embedding method (Algorithm \ref{alg:adaptation}), which continually adapts the cell intensities $I^j$ required for sufficient SNR and then optimizes cell colors $c_{SLM}^j$ under this constraint. Prior to \name's deployment, we perform a short, one-time calibration enabling the core unit to map SLM bitmap pixels to its camera's pixels. Upon completion of a window, \name\ runs Algorithm \ref{alg:adaptation} in parallel with ongoing modulations. It assesses robustness and perceptibility based on the completed windows' video, and accordingly increments or decrements $I^j$. To quantify robustness, the core unit runs data extraction on its past window's video (\S\ref{sec:verification}) and computes the bit error rate (BER). The perceptibility of each cell $j$ is measured as the perceived difference between patch $j$'s color with and without SLM light, approximated via the CIEDE2000 color difference formula~\cite{sharma2005ciede2000}. Then, $c_{SLM}^j$ is chosen to reduce perceptibility while satisfying the intensity requirement $I_j$:
\begin{equation}
\label{eq:color_match_eq}
\vspace{-.in}
\begin{aligned}
    & c_{SLM}^j = (\alpha*R^j_{p_{off}}, \alpha*G^j_{p_{off}}, \alpha*B^j_{p_{off}}), \\
    \text{subj. to } & \alpha*R^j_{p_{off}} +  \alpha*G^j_{p_{off}} + \alpha*B^j_{p_{off}} = I^j, 
\end{aligned}
\vspace{-.05in}
\end{equation}
where $R^j_{p_{off}}, G^j_{p_{off}}, B^j_{p_{off}}$ are the RGB values of $c_{p_{off}}^j$.

Note that, because data is distributed across all data cells, all $I^j$ are incremented when BER is above $\beta_{max}$ (Line \ref{glob_inc}). Further, $I^j$ may be incremented even when CIEDE2000$(c_{p_{on}}^j, c_{p_{off}}^j)$ exceeds $\Phi_{max}$, consciously prioritizing robustness.

\section{Video Integrity Verification}
\label{sec:verification}

 \begin{figure}[t]
    \centering
    \begin{subfigure}[t]{0.18\textwidth}
        \includegraphics[width=\textwidth]{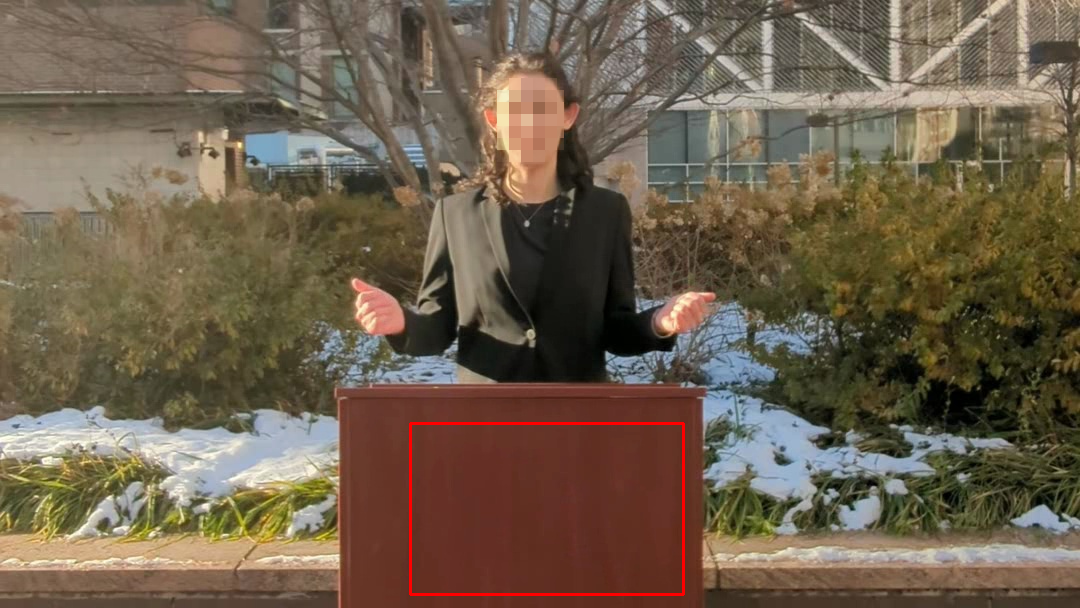}
    \end{subfigure}\hspace{.1em} 
    \begin{subfigure}[t]{0.18\textwidth}
        \includegraphics[width=\textwidth]{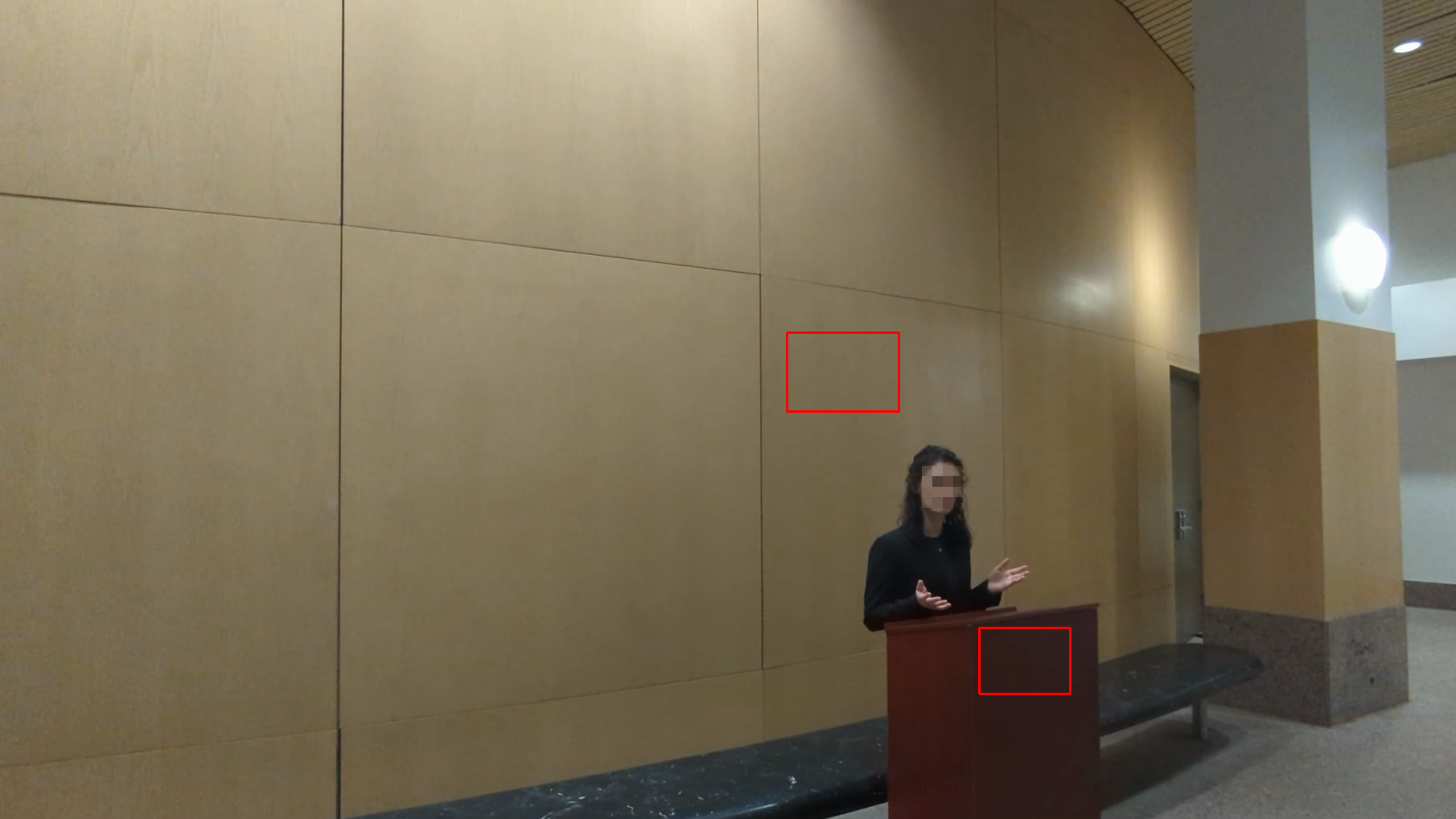}
    \end{subfigure} 
    
    \begin{subfigure}[t]{0.18\textwidth}
        \includegraphics[width=\textwidth]{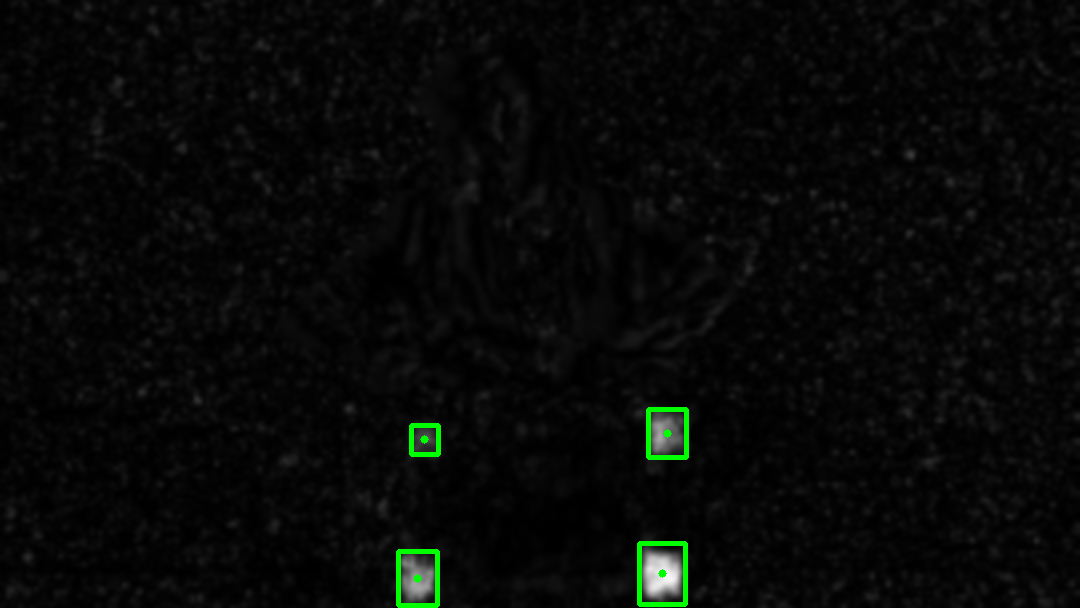}
    \end{subfigure}\hspace{.1em} 
    \begin{subfigure}[t]{0.18\textwidth}
        \includegraphics[width=\textwidth]{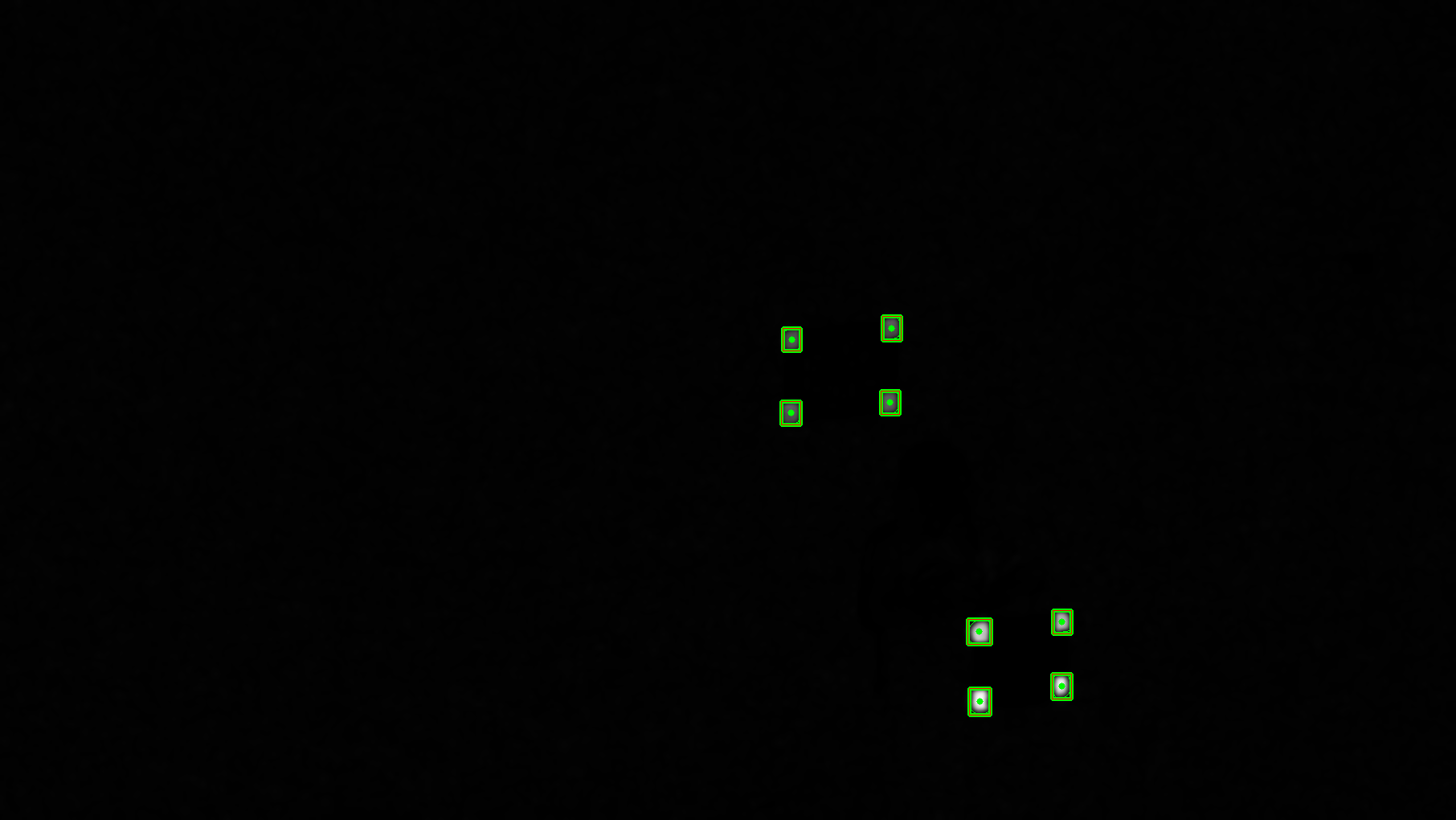}
    \end{subfigure}
    \caption{Embedded signatures are localized in any recording -- including those where signatures are replicated at the scene (top right) -- by condensing the video into a heatmap highlighting localization cells (bottom). Projection regions are outlined in red for visualization purposes.}
    \label{fig:loc_config} 
    \vspace{-0.1in}
\end{figure}

The last \name\ module verifies video integrity. It localizes optical signals in videos without prior knowledge of the projection surface, robustly recovers low-SNR data, and accurately assesses video integrity regardless of recording parameters.

\para{Optical signature localization}
To locate imperceptible optical signatures in a video, we leverage known properties of localization cells to create a \emph{heatmap} in which they light up in any scene (Figure \ref{fig:loc_config}). Since localization cells are modulated at $f_l$ (\S\ref{subsec:modulation}), the pixel values at these cells also exhibit oscillations at $f_l$. Thus, we apply a Fourier transform to each pixel-level signal in the video. We use a subset (e.g., 800) of the frames for this step for efficiency. We record each pixel's power at $f_l$ and normalize it by the noise at other frequencies. This yields our heatmap, where a pixel's brightness is proportional to its normalized power at $f_l$. We then detect the localization cells in the heatmap via contour detection~\cite{suzuki1985topological}. If fewer than four contours are detected, \name\ reports a verification failure.

Next \name\ determines the mapping between pixels in SLM bitmaps and the published video, allowing it to examine embedded cell signals. Any camera's view of the projected bitmaps is related to the bitmap itself via a homography, a projective transform that maps between two planes~\cite{agarwal2005survey}. Computing a homography requires a set of at least four corresponding points in each plane. We use our localization cells as these correspondences and apply the resulting homography to all frames to directly align video and SLM pixels.

\para{Signature data extraction} 
Having obtained the homography mapping bitmap cells to video pixels, \name\ extracts a signal for each cell by taking its average pixel intensity across frames. Building off the scheme described in \S\ref{subsec:modulation}, data is recovered from these signals as follows. First, \name\ determines the start and end of all embedding windows by finding periods of downtime in the synchronization cell signals. Second, it smooths and detrends data cell signals to remove noise and counter any gradual intensity shifts induced by camera auto-white-balance and auto-exposure. It then demodulates these signals per-window and passes its predictions to the concatenated error corrector to recover the signature data.

\para{MAC validation} \name\ extracts the descriptors from all signatures and validates them against their MACs. A MAC mismatch suggests a window's embedded data has been corrupted by tampering or decoding errors. \name\ flags windows with corrupt descriptors as untrustworthy, as their content cannot be verified.

\para{Descriptor comparison}
A video's integrity is determined by comparing recovered descriptors to those computed on portrayed content. For each window $i$, \name\ downsamples the video to the core unit framerate and extracts its 
feature hashes (\S\ref{sec:digest}). Recovered window numbers are ensured to be consecutive, and hashes are compared to their counterparts in the descriptor recovered from window $i + 1$.\footnote{For the final $n$ seconds of a speech video to be verifiable, it must be followed by a window of embedding only. Our implementation configures $n$ to be 4.5 seconds; increasing the embedding bandwidth can allow shorter windows (\S\ref{sec:limitations_future}).}
Both the identity feature and dynamic feature hashes are compared via Hamming distance. The identity feature hash is computed for \emph{all frames} to ensure its consistency throughout the video. An integrity decision is made using the maximum identity and dynamic hash distances across windows. If either exceeds a configured decision threshold, the video is deemed falsified.
\section{Prototype Implementation}
\label{sec:prototype}
\begin{figure}[t]
    \vspace{-.1in}
    \centering
    \includegraphics[width=0.9\columnwidth]{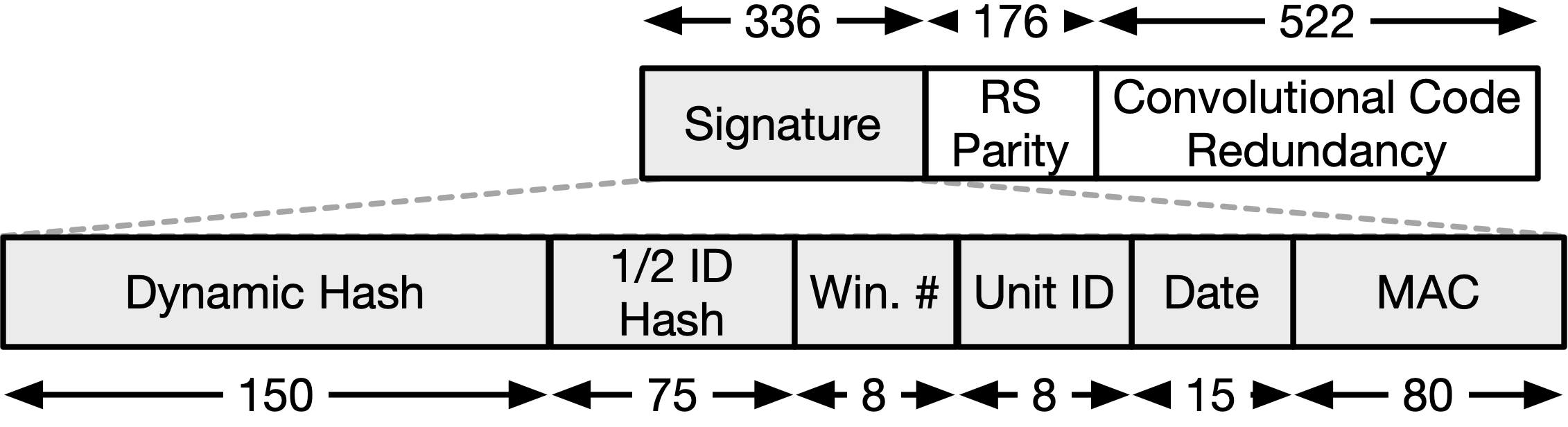}
    \vspace{-.1in}
    \caption{Data embedded per window in our implementation (sizes in bits). The signature includes the descriptor and MAC.}
    \label{fig:impl_payload}
\end{figure}

\para{Software and hardware} We fabricate a \name\ prototype using readily-available, affordable components. The core unit (Figure~\ref{fig:teaser}) consists of an SLM (Texas Instruments  DLPDLCR230NPEVM, \$299), an RGB camera (8MP IMX179 Arducam) running at \SI{24}{FPS}, and a Raspberry Pi 4B for controlling the SLM. 
Code for real-time signature creation and adaptive embedding runs on a MacBook Pro 14, using CPU only. 
Deployment and verification are automated in single scripts, minimizing required operational expertise. Verification is run on the MacBook (offline) for our proof-of-concept. We leave its integration into a cloud service to future work. The average time to verify a 30~s video is 86~s, which can be significantly reduced by using GPU and parallelizing per-frame analysis. We empirically set our descriptor comparison decision thresholds using data from \S\ref{subsec:overall_performance}.

\para{Optical embedding parameters} We utilize 640 x 360 pixel SLM bitmaps, partitioned into a 16 x 9 grid of cells. This corresponds to 87 data cells and 32 synchronization cells, in addition to our four larger localization cells. We set $f_d$ to be \SI{3}{Hz}, as we found that BER increases significantly at higher frequencies. Finally we set $f_l$ to be 6~Hz, and our window duration to be 4.5~s, consisting of .5~s of downtime followed by 4~s of modulation. Thus we can embed 12 bits per data cell, i.e., 12*87=1044 bits per window. We configure Algorithm \ref{alg:adaptation} with $\beta_{max} = 0$, $\phi_{max} = 5$, and $\delta = 5$.

\para{Signature parameters} Figure \ref{fig:impl_payload} shows the composition of our signatures. We choose hash sizes of 150 bits, as both mathematical analysis (Theorem \ref{theorem:k_perf}) and empirical findings (\S\ref{subsec:digest_robustness}) indicate further increases in hash size yield diminishing returns. 
We distribute each identity feature hash over two windows' descriptors, such that the recovered identity feature hash is updated every other window during verification. This leaves space for more RS parity while preserving functionality, as a speaker's identity cannot vary across windows at a physical event. We use a 128-bit secret key to compute HMAC-SHA256 MACs, which we truncate to 80 bits to minimize MAC size while maintaining reasonable security~\cite{hmac_rfc}.\footnote{For improved resilience to forgery, RS parity can be reduced to fit further MAC bits.}

\begin{table*}[t]
    \small
     \begin{tabularx}{\textwidth}{@{}p{1.7cm}|YYYYYYYYYYYY@{}}
        \hlineB{2}
        \vspace{0.25em}
         & \multicolumn{11}{c}{\bf{Detector}} \\ 
          \bf{Deepfake\newline Model} & 
     	Meso4\newline\cite{afchar2018mesonet}&
         Xception\newline\cite{rossler2019faceforensics++} & Capsule\newline\cite{nguyen2019capsule} & Efficient\newline\cite{tan2019efficientnet}  & \phantom{a}SRM\newline\cite{luo2021generalizing} & \phantom{a}SPSL\newline\cite{liu2021spatial}  &  
         \phantom{a}Recce\newline\cite{cao2022end} & 
         \phantom{a}UCF\newline\cite{yan2023ucf}  & 
         \phantom{a}TALL\newline\cite{xu2023tall}  & 
         AltFreeze.\newline\cite{Wang_2023_CVPR} & 
         \bf{Ours}\\
        \hline
        DaGAN (R) & 0.52 & 0.73 & 0.59 & 0.67 & 0.70 & 0.68 & 0.68 & 0.72 & 0.61 & 0.50 & \bf{0.99}\\
        SadTalker (R) & 0.62 & 0.72 & 0.71 & 0.68 & 0.70 & 0.70 & 0.66 & 0.71 & 0.62 & 0.45 & \bf{0.99}\\
        FOMM (R) & 0.64 & 0.83 & 0.73 & 0.80 & 0.79 & 0.74 & 0.77 & 0.80 & 0.69 & 0.43 & \bf{0.99}\\
        TalkLip (R) & 0.83 & 0.95 & 0.86 & 0.95 & 0.96 & 0.93 & 0.94 & 0.92 & 0.80 & 0.34 & \bf{0.99}\\\
        FSGAN (I) & 0.66 & 0.88 & 0.69 & 0.79 & 0.84 & 0.88 & 0.83 & 0.83 & 0.79 & 0.57 & \bf{1.00}\\
        \hlineB{2}
    \end{tabularx}
    \caption{AUC scores achieved by \name\ and ten state-of-the-art passive detectors on our end-to-end video dataset. Our dataset includes both reenactment (R) and identity swap (I) deepfakes. Best performing method is bolded.}
    \label{tab:aucs}
    \vspace{-.14in}
\end{table*}

\section{Protection Performance Evaluation}
\label{subsec:overall_performance}

We evaluate \name's ability to detect tampering of speaker identity and lip and face motion achieved via deepfakes and basic editing. All studies were approved by our Institutional Review Board.

\subsection{Falsification with Deepfake Models}
\label{subsec:deepfake-eval}
To demonstrate \name's protection performance across speakers and deepfake models, we first collect a large-scale video dataset of speeches delivered with our core unit present. We then generate extensive identity swap and reenactment deepfakes, and examine our verification module's ability to differentiate real and fake videos. 

\para{Data collection} 
We construct our own dataset because existing deepfake video datasets were of course not recorded with \name\ deployed. 
Moreover, they lack \emph{pairs} of real and fake videos, which would otherwise be needed to emulate \name's comparisons of recovered descriptors to portrayed content. 

To collect authentic videos, we invited 20 participants (11 male and 9 female, ages 18 to 54) to read aloud six paragraphs (roughly 33~s each) while our core unit was deployed. Paragraphs were sourced from the Presidential Deepfakes Dataset (PDD)~\cite{sankaranarayanan2021presidential} and displayed on a monitor. Participants were recruited through flyers and emails within our institution and each compensated \$10. The core unit was positioned \SI{1.5}{m} away from the participant and \SI{2}{m} away from a white wall.  We utilized a 100 x 70 cm portion of the wall for projection, thoroughly investigating other setups in \S\ref{sec:robustness}. We simultaneously recorded on four cameras placed around the core unit: a Google Pixel 6A, an iPhone 14 in ProRes mode, a webcam (Mokose 4K), and a DSLR camera (Canon EOS 60D). In total, we recorded 257~min of speech across 474 videos.

\para{Deepfake generation} For each original video, we use FSGAN~\cite{nirkin2019fsgan} to generate an identity swap deepfake where the speaker face is supplanted with that of a randomly selected alternative identity. We generate reenactment deepfakes using four state-of-the-art models: DaGAN~\cite{hong2022depth}, First Order Motion Model (FOMM) \cite{siarohin2019first}, TalkLip~\cite{wang2023talklip} and SadTalker~\cite{zhang2023sadtalker}. The reenactment deepfakes modify the speaker's facial movements to reflect their delivery of a different speech from PDD. While TalkLip exclusively modifies the lip region based on driving audio, FOMM, DaGAN, and SadTalker include face images as driving input to also modify expression. These varied reenactment scopes test the coverage of our dynamic features. 

For all deepfakes, we only modify the facial region, leaving the rest of the scene (including the projection surface) as-is. This emulates a realistic attack scenario in which an attacker creates a convincing falsified video by maintaining the video's context while changing speech content. We generate 1,883 reenactment deepfakes (753~min) and 473 identity swap deepfakes (261~min) total.

\para{Metrics} We input all videos to the verification module and record the Hamming distances between computed and recovered dynamic and identity hashes, as well as the module's final decision on video integrity. We quantify performance using recall (i.e., true positive rate) and Area Under Curve (AUC), standard metrics for evaluating binary classifiers.
An AUC of 1 indicates perfect separation of positive and negative class scores. 
We use hash distances as our class scores. Since reenactment deepfakes maintain identity but change content, we compute the AUC for reenactment detection using only dynamic feature hash distances. Similarly, we use identity feature hash distances for identity swaps. The recall is based on \name's final binary decisions and thus considers both distances. 

\para{Passive detector comparison} We benchmark \name\ against 10 state-of-the-art passive detection models (Table \ref{tab:aucs}), spanning a range of whole-video and frame-level methods. The goal of these comparisons is to ensure our created dataset is sufficiently challenging to fairly evaluate \name. 
We choose these models as they are top performers in the comprehensive DeepfakeBench benchmark~\cite{DeepfakeBench_YAN_NEURIPS2023}. Specifically, each ranks within the top-3 methods for at least three datasets assessed, indicating effective cross-domain performance. We use implementations provided via DeepfakeBench for all models. Each was trained on the FaceForensics++ dataset~\cite{rossler2019faceforensics++}, which contains both identity swap and reenactment deepfakes. We do not fine-tune the models on our data, since \name\ requires no deepfake-specific fine-tuning. 
Further, in practice, passive detectors do not have \emph{a priori} knowledge of the origins of their input.

\para{Overall results} 
As shown in Table \ref{tab:aucs}, \name\ achieves AUCs above 0.99 for all deepfake models and outperforms passive detectors by 37\% on average.  \name\ has a recall of 100\%, indicating it detects every one of the over 2,000 fake videos in our dataset. It additionally exhibits generalizability and explainability.

Among all reenactment deepfakes, only one possessed an identity hash distance above the decision threshold. Among all identity swap deepfakes, only 23 possessed an anomalous dynamic hash distance. Thus, our identity and dynamic features effectively isolate identity and motion-specific video elements, respectively. As a result, \name\ can report the \emph{type} of video falsification detected.

We also see that \name's performance generalizes across deepfake models. We attribute this to our descriptors' focus on higher-level, semantically-meaningful visual features (e.g., temporal lip movement patterns, facial characteristics). These features are guaranteed to vary with content (deepfake-generated or not), unlike the low-level, model-specific artifacts sought by passive detectors. 

While the passive detector failures cannot be diagnosed, as they stem from black-box neural networks, \name's results are quite explainable. Its inaccuracies are overwhelmingly false positives (real videos labeled fake) triggered by high dynamic hash distances. These cases are caused by sporadic FaceMesh inaccuracies, which degrade the Pearson correlation between dynamic feature vectors. We consider alternative features not relying on FaceMesh in \S\ref{sec:limitations_future}. 

\para{Signature extraction failures} Out of 2,400 inputted videos, \name\ could not localize the signature in 36 (6 original, and their 30 deepfake counterparts), all corresponding to one participant's session. While a fraction of videos that passed localization had corrupted extracted signatures, each had sufficient intact signatures to enable a final verification decision. All such failures can be resolved by configuring the tradeoff between SNR and imperceptibility (\S\ref{sec:adaptive_embedding}).

\subsection{Other Falsification Attacks}
Beyond the above deepfake falsifications, \name\ inherently addresses other common falsification techniques and attacks.

\para{Speech speed modification}
Attacks modifying only the playback speed of a video~\cite{nancy19, harris24_slow} are achieved by either changing a video's frame rate or duplicating/dropping frames to change the effective content speed. Both approaches alter the structure and frequency of embedded signals (e.g., halving their frequency to achieve a 0.5x slow-down), triggering localization and demodulation failures. A knowledgeable attacker may preserve the playback speed of only the optical signature region; however, this will desynchronize signatures and speech content, causing a conflict of dynamic features.

\para{Video clipping and splicing} Removing portions of video or splicing together clips (e.g., to re-order the speaker's words) changes the progression of the speakers' face movements, captured by dynamic features. \name\ thus prevents such edits. Non-consecutive window numbers would expose clever re-ordering of intact windows.

\para{Signature injection or manipulation} An attacker may try to embed a signature that complements their modified or synthetic content by digitally injecting or manipulating pixel signals. Without \name's secret key, though, they fundamentally cannot.
Copying other videos' signatures fails as they are highly event-specific.

\section{Protection Robustness Evaluation}
\label{sec:robustness}

The previous section evaluated \name's protection performance under a single recording and attack configuration. We now delve into its performance across a broad range of practical attack and recording conditions. We separately evaluate each condition's impact on descriptor and optical signature embedding performance.

\para{Summary of results} \name's descriptor extraction and optical embedding modules both support recording at up to 60$\degree$ off axis and 3~m away from the speaker and projection surface, with supported range further extended when optical zoom is employed. Descriptors enable detection of content falsifications as fine-grained as $\leq$ 1.35~s of a window and generalize across hundreds of evaluated identities. Finally, descriptors and optical signatures are resilient to varied post-processing techniques, including compression and transcoding. 

\subsection{Descriptor Robustness}
\label{subsec:digest_robustness}

We evaluate our descriptor identity and dynamic features in terms of their pose-invariance and generalization across speakers. We explore dynamic features' sensitivity to fine-grained reenactment deepfakes, wherein an attacker modifies only a portion of a window.

\para{Multi-pose and fine-grained deepfake datasets}
To evaluate our identity features across poses and subjects at a large scale, we turn to the Labeled Faces in the Wild dataset (LFW)~\cite{huang2008labeled}. LFW images are captured at extensive angles and distances in unconstrained environments.  Our experiments span all 1,680 LFW individuals.

To evaluate our dynamic features' pose invariance and sensitivity to fine-grained modifications, we build a dataset of speech videos simultaneously captured from extensive angles and distances. We then generate pinpointed reenactment deepfakes of these videos.

Specifically, we construct a multi-camera rig of six synchronized 1080p webcams positioned across two distances (1.5~m and 3~m) and three angles (0$\degree$, 45$\degree$, and 60$\degree$ from the speaker\footnote{We assume performance is symmetrical about the speaker face and thus do not mirror the configuration at angles $<0\degree$.}). We record nine participants as they read aloud four paragraphs (roughly 15-30~s each) sourced from the popular acoustic-phonetic corpus TIMIT~\cite{garofolo1993timit}.

For each of these 216 real videos, we create three deepfake counterparts by changing individual words in the paragraphs. We replace each targeted word's \emph{exact} portion of video with a reenactment deepfake of the same duration portraying the speaker uttering a different word from TIMIT. A SadTalker, FOMM, DaGAN, and TalkLip version is created for each case. 
We arrive at 2,808 videos (2,592 deepfakes) of varied camera positions and falsification granularities.

\para{Metrics} 
We report AUC scores for verification of all LFW subjects using identity features. For dynamic features, we report \emph{per-window} AUC scores under 
various percentages of modified content, by duration. 
In computing these AUCs, positive class scores are distances between real video windows and their fake counterparts. Negative scores are distances between dynamic features extracted from windows of the same scene, shot from different camera positions.  

\begin{figure}[t]
    \centering
    \includegraphics[width=\columnwidth]{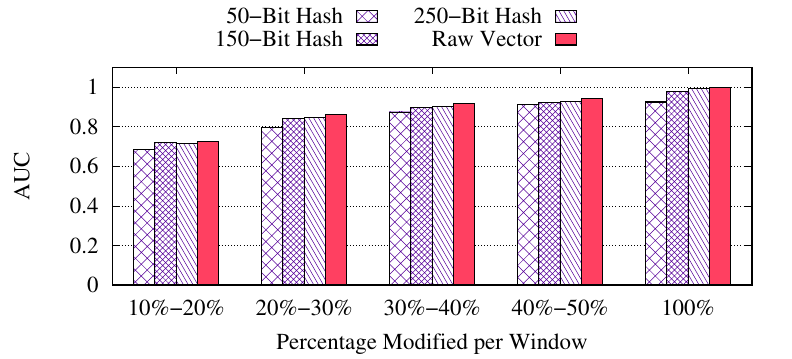}
        \vspace{-.22in}
        \caption{Window-level AUC scores achieved by dynamic features across modification granularities and hash sizes.}
        \vspace{-.1in}
        \label{fig:dyn_hist}
\end{figure}

\para{Sensitivity to fine-grained modifications}
Figure \ref{fig:dyn_hist} shows our dynamic feature AUCs across modification granularities, while Table \ref{tab:dyn_aucs} summarizes the AUCs of passive detectors. \name's 150-bit dynamic hashes score an AUC of .98 for fully-falsified windows and AUCs $\geq$ 0.90 for modification percentages $\geq$ 30. This is a 40\% gain over the best performing passive detector on the 30-40\% bin.

We observe that AUC drops with decreasing modification percentage. Windows with minor modifications may exhibit dynamic feature signals dominated by the similarities between remaining clean content, causing false negatives. For the 10-20\% bin, \name\ AUC drops to 0.72. 
Notably, this modification rate corresponds to a highly specific attack, in which as little as 0.45~s of words within one 4.5~s window are \emph{precisely} supplanted. If any introduced reenactments are even a few frames longer or shorter than content they are replacing, all subsequent frames in the video are shifted; this has the effect of modifying 100\% of content in subsequent windows.

Ultimately, detection capability is inevitably dependent on the degree of modification. Indeed, existing passive detectors struggle to temporally localize finer-grained falsifications~\cite{cai2022you}. \name\ maintains reasonable performance and inherently localizes falsifications on the resolution of windows. We explore the potential of other dynamic features to counter subtle falsifications in \S\ref{sec:limitations_future}.

\begin{figure}[t]  
   \begin{minipage}[t]{0.45\textwidth}
        \centering
        \begin{subfigure}[t]{0.25\textwidth}
            \includegraphics[width=\textwidth]{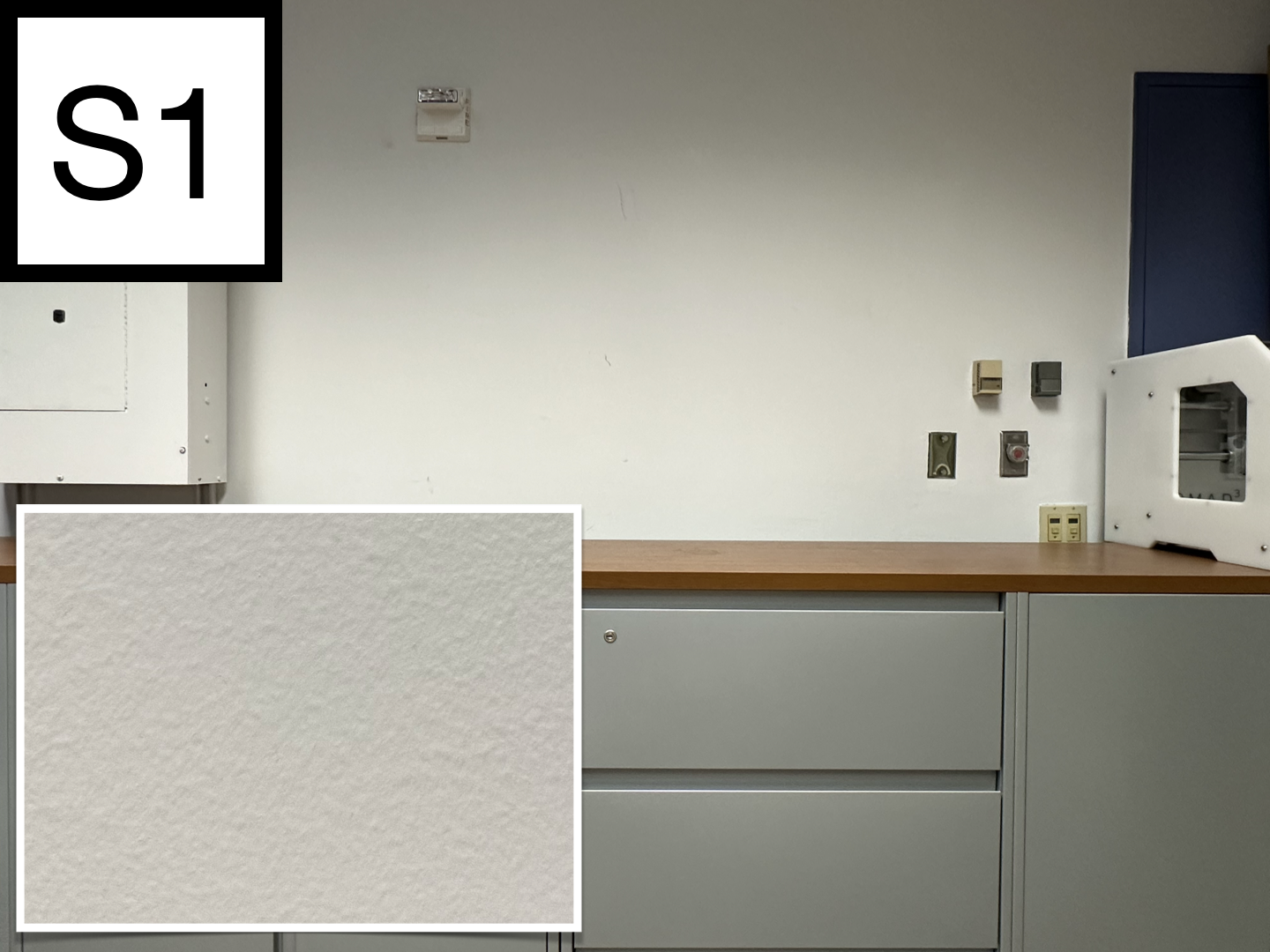}
        \end{subfigure}
         \begin{subfigure}[t]{0.25\textwidth}
            \includegraphics[width=\textwidth]{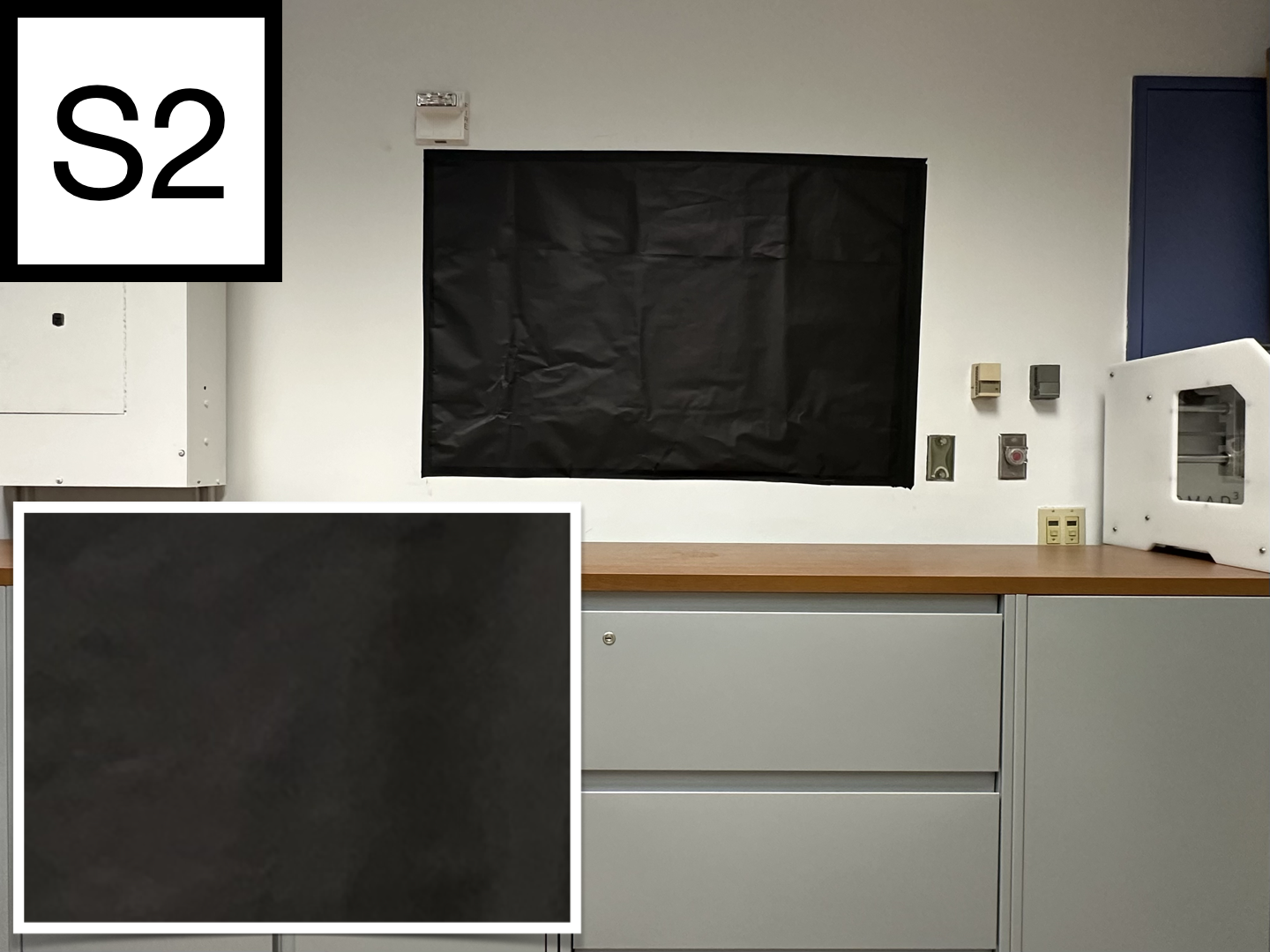}
        \end{subfigure}
         \begin{subfigure}[t]{0.25\textwidth}
            \includegraphics[width=\textwidth]{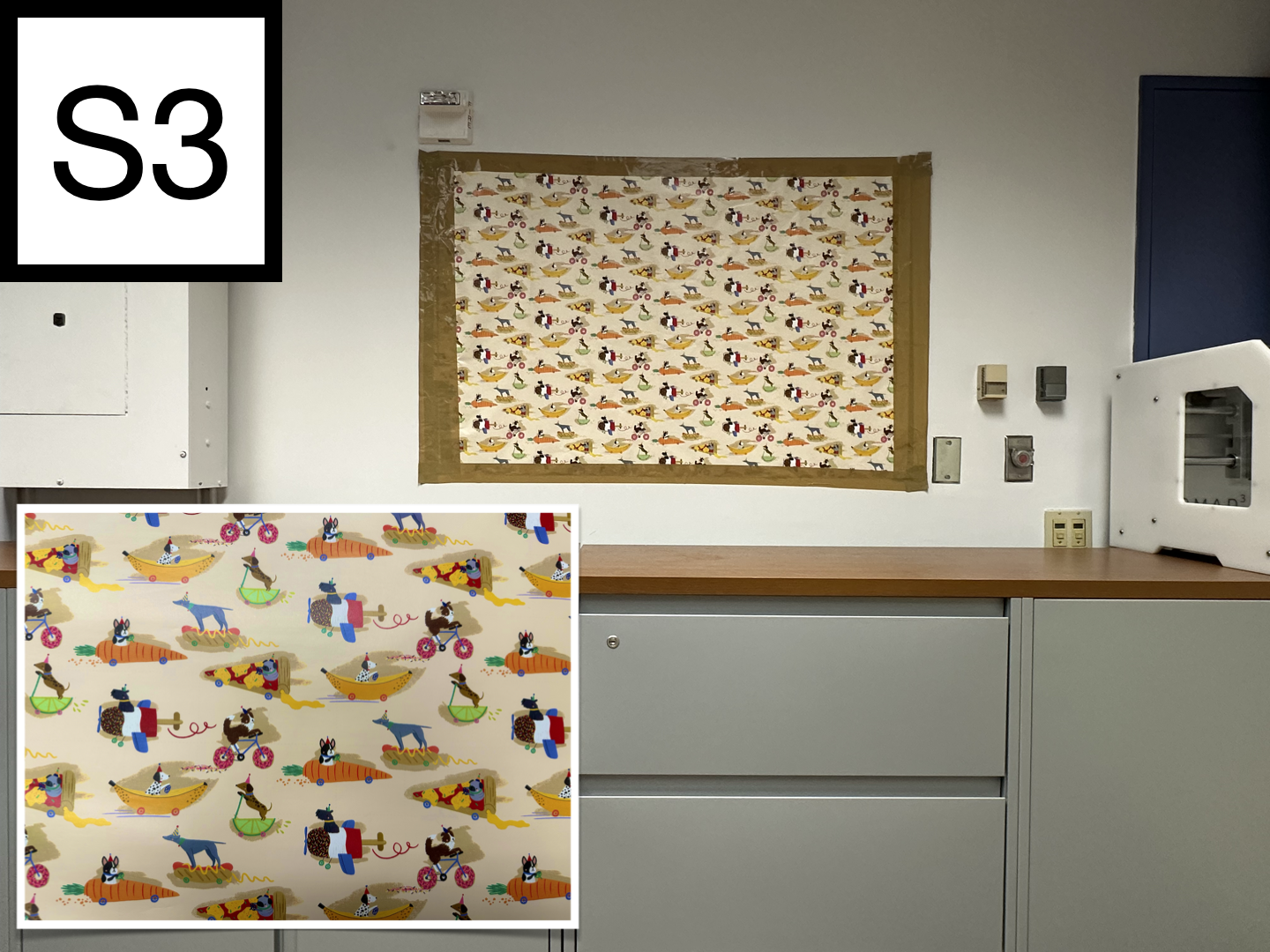}
        \end{subfigure}

        \begin{subfigure}[b]{0.24\textwidth}
            \includegraphics[width=\textwidth]{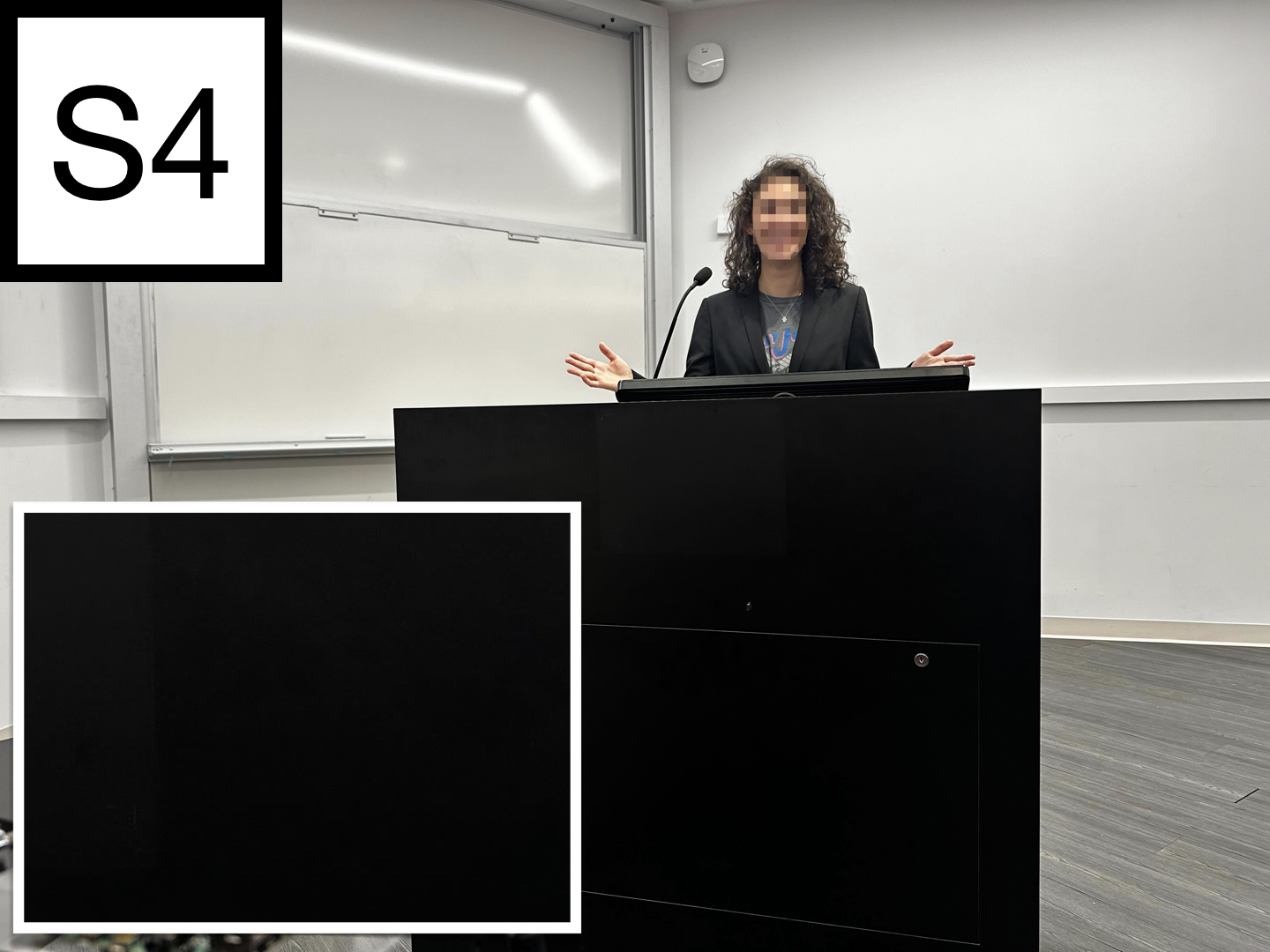}
        \end{subfigure}    
        \begin{subfigure}[b]{0.24\textwidth}
            \includegraphics[width=\textwidth]{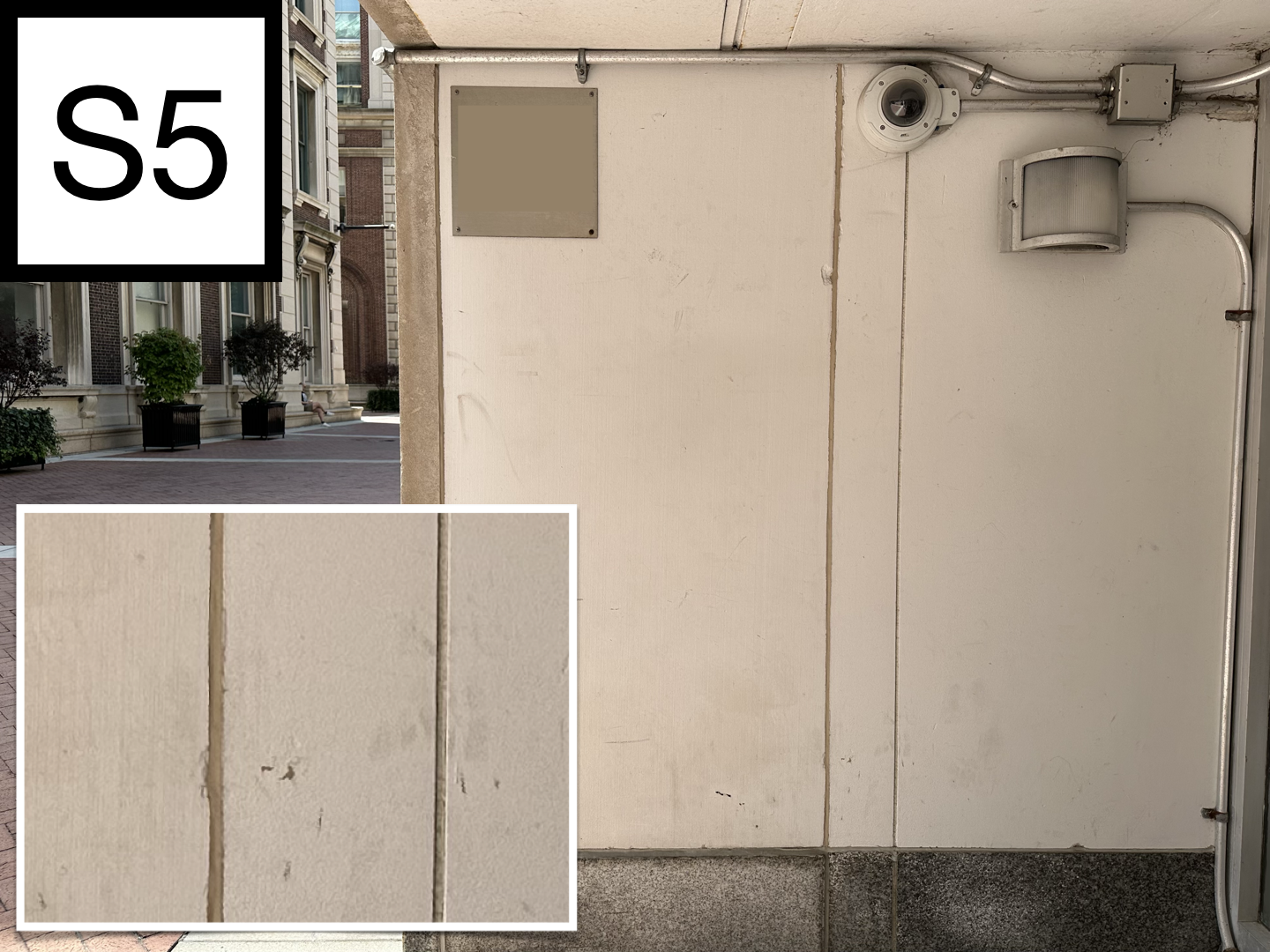}
        \end{subfigure}
        \begin{subfigure}[b]{0.24\textwidth}
            \includegraphics[width=\textwidth]{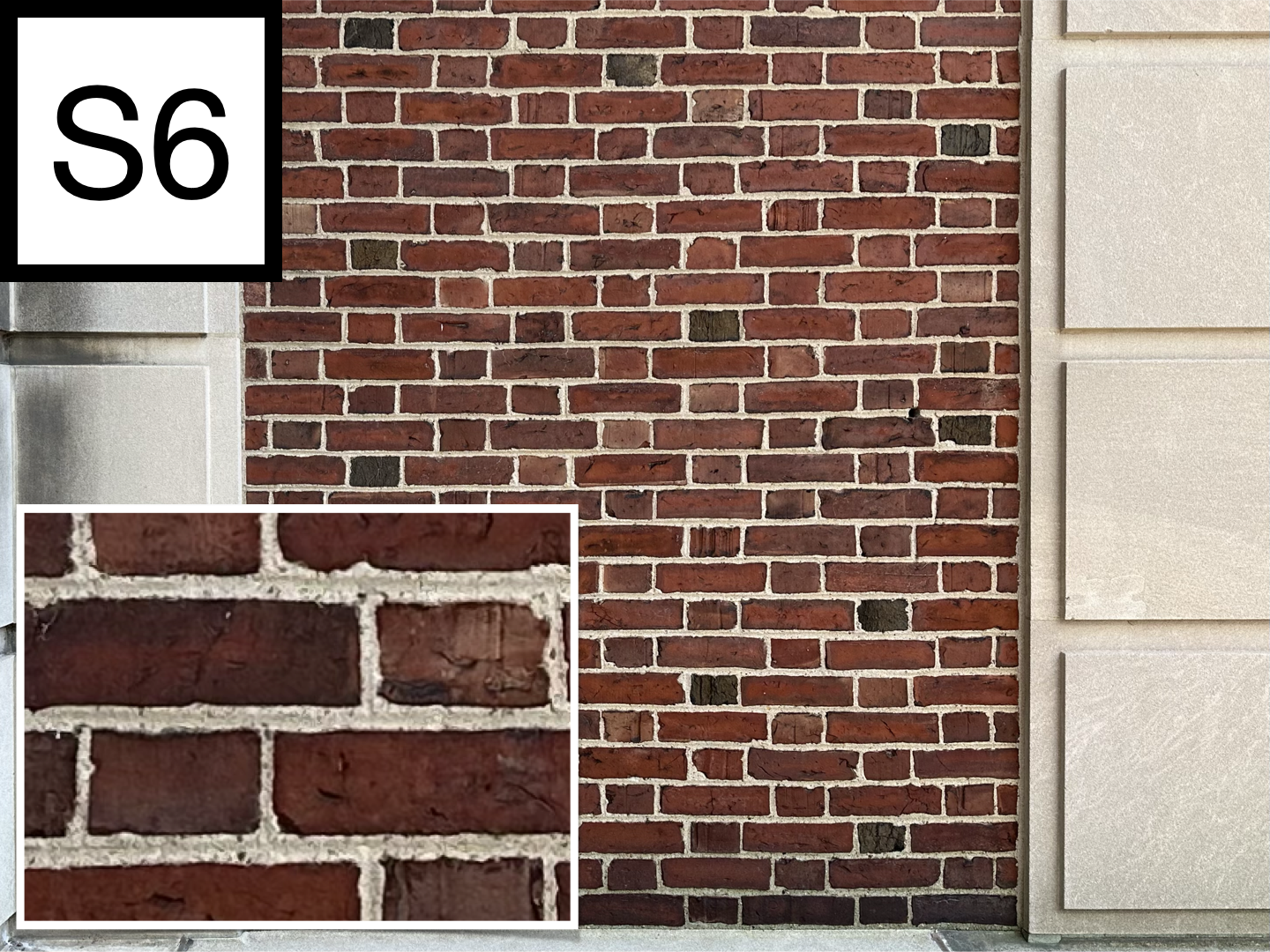}
        \end{subfigure}
        \begin{subfigure}[b]{0.24\textwidth}
            \includegraphics[width=\textwidth]{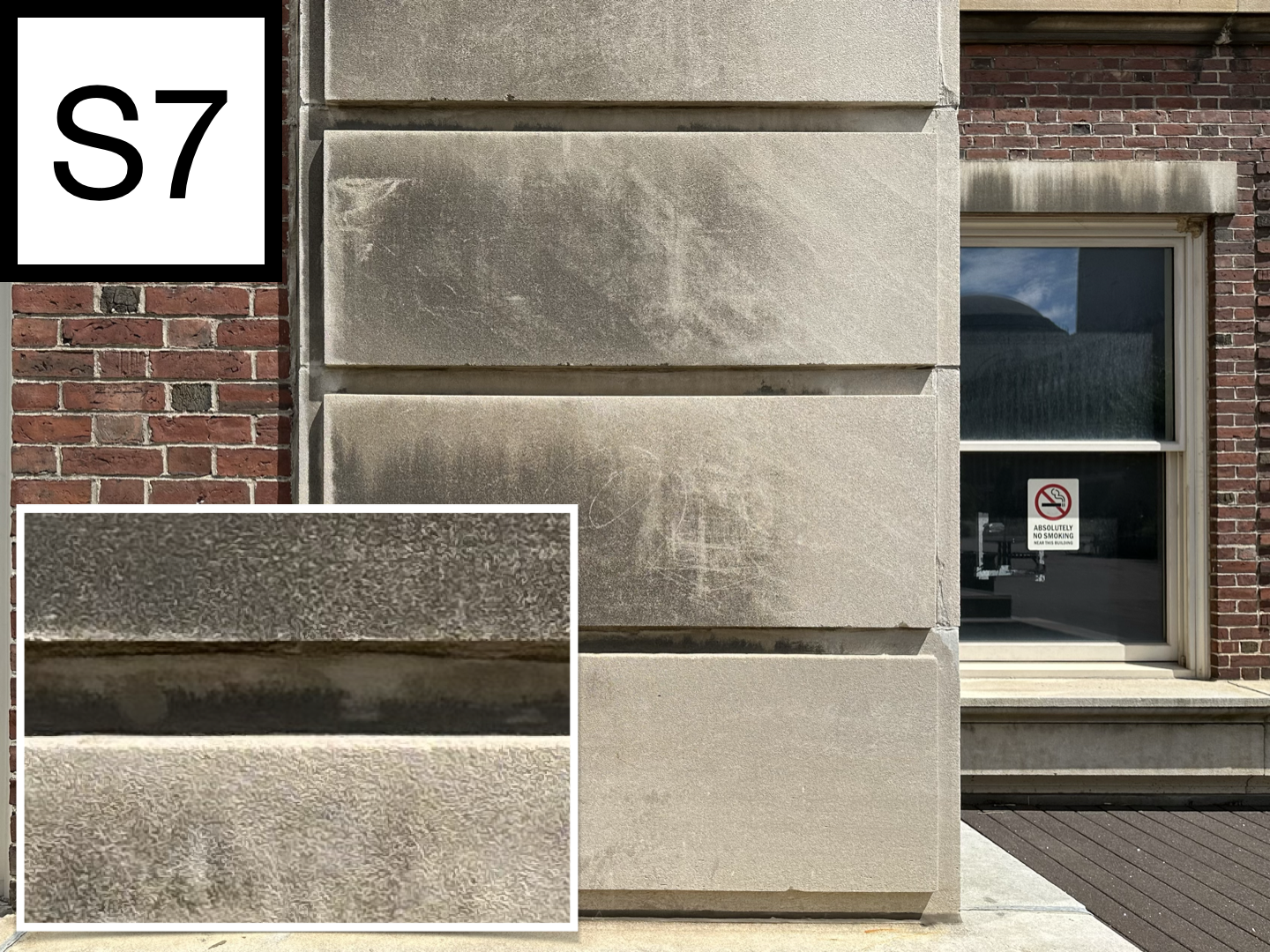}
        \end{subfigure}
        \vspace{-.1in}
        \caption{Environments and respective surfaces (S1-7) tested in ~\S\ref{subsec:embedding_robustness}. Insets show close-ups of the projection surfaces, varying in their textures and coloration. Surfaces 5, 6, and 7 are outdoors.}
        \label{fig:surfaces}
    \end{minipage}
    \vspace{-.1in}
\end{figure}

\para{Generalization across identities} Our identity hashes score an AUC of 0.99 in differentiating all 1,680 LFW subjects. Dynamic feature hashes are similarly robust across speakers, with errors in detecting reenactment deepfakes distributed evenly across subjects.

\para{Hash size} The AUC drop-off between 150- and 50-bit dynamic feature hash sizes (Figure \ref{fig:dyn_hist}) validates our choice of 150-bit hashes.
Hashing ArcFace vectors to 150 bits lowers their AUC by just 0.0016.

\para{Recording distance and angle} Descriptor performance is consistent across recording positions up to 3~m and 60\degree\ off-axis from the speaker. Amongst videos captured at the harsh 3~m, 60\degree\ position, dynamic feature hashes have an AUC of 0.96 in detecting fully falsified windows. Note that we record all dataset videos with no zoom for consistency; using zoom naturally boosts supported range.

\para{Video post-processing} We do not observe significant changes in descriptor performance upon compression, transcoding, or filtering, likely due to the diversity of ArcFace and FaceMesh training data.

\subsection{Signature Embedding Robustness}
\label{subsec:embedding_robustness}

\begin{table}[t]
    \centering
    \label{tab:extraction_robustness}
    \begin{subtable}[t]{1\columnwidth}
        \small
        \caption{\small Distance and viewing angle.}
        \vspace{-0.08in}
        \label{tab:positions}
        \begin{tabularx}{\columnwidth}{@{}r|YYY|YYY|YYY@{}}
            \hlineB{2}
            & \multicolumn{3}{c|}{\bf{2m}} & \multicolumn{3}{c|}{\bf{3.5m}} & \multicolumn{3}{c}{\bf{5m}} \\
            & \bf{0$\degree$} & \bf{45$\degree$} & \bf{60$\degree$}
            & \bf{0$\degree$} & \bf{45$\degree$} & \bf{60$\degree$}
            & \bf{0$\degree$} & \bf{45$\degree$} & \bf{60$\degree$} \\
            \hline
            R & 0.01 & 0.02 & 0.02
            & 0.05 & 0.04 & 0.05
            & 0.13 & 0.13 & 0.12 \\
            V & 0.00 & 0.00 & 0.00
            & 0.00 & 0.00 & 0.00
            & 0.04 & 0.04 & 0.03 \\
            F & 0.00 & 0.00 & 0.00
            & 0.00 & 0.00 & 0.00
            & 0.03 & 0.03 & 0.01 \\
            \hlineB{2}
        \end{tabularx}
    \end{subtable}
     \begin{subtable}[t]{1\columnwidth}
        \small
        \caption{\small In-video cell resolution.}
        \vspace{-0.08in}
        \label{tab:cell_dim}
        \begin{tabularx}{\columnwidth}{@{}r|Y|Y|Y|Y@{}}
            \hlineB{2}
            & \bf{30 x 30 px} & \bf{35 x 35 px} & \bf{40 x 40 px} & \bf{50 x 50 px}\\
            \hline
            R  & 0.12 & 0.05 & 0.04 & 0.02\\
            V & 0.03 & 0.00 & 0.00 & 0.00 \\
            F & 0.01 & 0.00 & 0.00 & 0.00 \\ 
            \hlineB{2}
        \end{tabularx}
    \end{subtable}
    \begin{subtable}[t]{1\columnwidth}
        \small
        \captionsetup{justification=centering}
        \caption{\small Ambient light intensity (lx) and projection surface.}
        \vspace{-0.08in}
        \label{tab:surfs}
        \addtolength{\tabcolsep}{-0.4em}
        \begin{tabularx}{\columnwidth}{@{}r|YYY|YYY|Y|Y|Y|Y@{}}
            \hlineB{2}
              & \multicolumn{3}{c|}{\bf{320}} & \multicolumn{3}{c|}{\bf{750}} & \bf{530}
            & \bf{3k} &  \bf{3.5k} &  \bf{2.6k}\\
            & \bf{S1} & \bf{S2} & \bf{S3}
            & \bf{S1} & \bf{S2} & \bf{S3}
            & \bf{S4}
            & \bf{S5} & \bf{S6} & \bf{S7} \\
            \hline
             R & 0.01 & 0.07 & 0.01 & 0.01 & 0.08 & 0.01 & 0.70 & 0.01 & 0.03 & 0.01 \\
            V & 0.00 & 0.04 & 0.00 & 0.00 & 0.02 & 0.00 & 0.05 & 0.00 & 0.00 & 0.00\\
            F & 0.00 & 0.00 & 0.00 & 0.00 & 0.00 & 0.00 & 0.00 & 0.00 & 0.00 & 0.00\\
            \hlineB{2}
        \end{tabularx}
        \label{tab:bers_surf}
    \end{subtable}
    \begin{subtable}[t]{1\columnwidth}
        \small
        \caption{\small Device type.}
        \vspace{-0.08in}
        \label{tab:device}
        \begin{tabularx}{\columnwidth}{@{}r|Y|Y|Y|Y|Y@{}}
        \hlineB{2}
            & \bf{Google Pixel} & \bf{iPhone (HD)} & \bf{iPhone (ProRes)} & \bf{Webcam} & \bf{DSLR} \\
            \hline
            R & 0.01 & 0.01 & 0.01 & 0.01 & 0.00 \\
            V & 0.00 & 0.00 & 0.00 & 0.00 & 0.00 \\
            F & 0.00 & 0.00 & 0.00 & 0.00 & 0.00 \\
            \hlineB{2}
        \end{tabularx}
    \end{subtable}
    \begin{subtable}[t]{1\columnwidth}
        \small
        \caption{\small Video transcoding and compression.}
        \vspace{-0.08in}
        \label{tab:compression}
          \addtolength{\tabcolsep}{-0.4em}
        \begin{tabularx}{\columnwidth}{@{}r|YYc|YYYYY@{}}
            \hlineB{2}
            & \multicolumn{3}{c|}{\bf{Transcoding}} & \multicolumn{5}{c}{\bf{Bitrate Decrease}} \\
            & \bf{None} & \bf{H.264} & \bf{MPEG4} & \bf{10\%} & \bf{30\%} & \bf{50\%} & \bf{70\%} & \bf{90\%} \\
            \hline
             R & 0.01 & 0.02 & 0.04 & 0.03 & 0.02 & 0.03 & 0.03 & 0.02 \\
            V & 0.00 & 0.00 & 0.00 & 0.00 & 0.00 & 0.00 & 0.00 & 0.00 \\
            F & 0.00 & 0.00 & 0.00 & 0.00 & 0.00 & 0.00 & 0.00 & 0.00 \\
            \hlineB{2}
        \end{tabularx}
    \end{subtable}
    \begin{subtable}[t]{1\columnwidth}
        \small
        \captionsetup{justification=centering}
        \caption{\small Contrast (C), exposure (E), auto-enhance and monochrome edits.}
        \vspace{-0.08in}
        \label{tab:aesthetic}
        \begin{tabularx}{\columnwidth}{@{}r|Y|Y|Y|Y|Y|Y@{}}
            \hlineB{2}
            & \bf{C-50\%} & \bf{C+50\%} & \bf{E-50\%} & \bf{E+50\%} & \bf{Auto} & \bf{Mono} \\
             \hline
              R & 0.04 & 0.01 & 0.01 & 0.02 & 0.01 & 0.01 \\
              V & 0.00 & 0.00 & 0.00 & 0.00 & 0.00 & 0.00 \\
              F & 0.00 & 0.00 & 0.00 & 0.00 & 0.00 & 0.00 \\
             \hlineB{2}
        \end{tabularx}
    \end{subtable}   
    \caption{Signature embedding robustness across recording factors. We report the BER at each stage of error correction (\S\ref{subsec:code}): raw (R), before error correction; after Viterbi decoding (V); and final (F), after RS error correction.}
    \vspace{-0.3in}
\end{table}

We evaluate our optical signature embedding scheme's robustness with respect to \name's ability to extract embedded data from videos. We consider extensive practical factors, from ambient lighting and projection surfaces to video post-processing.

\para{Experimental setup} To assess the effects of projection surface and ambient lighting, we project onto seven surfaces (Figure \ref{fig:surfaces}), including three outdoors under dynamic cloud coverage. We evaluate two lighting conditions for S1-3, for a total of 10 environments. To assess the remaining factors, we project onto S1. For each scenario, we embed a random bitstream for 100~s and extract the data from the recording. By default, we record with the core unit setup detailed in \S\ref{subsec:overall_performance} and a Google Pixel at 2~m from the projection surface. We quantify robustness in terms of the bit error rate (BER) at each stage of decoding: raw, post-Viterbi, and final, upon full error correction. 

\para{Recording distance and angle} \name's embedding supports recording up to 3.5~m and 60$\degree$ from the projection surface (Table \ref{tab:positions}). We find embedding robustness is primarily constrained not by camera position, but rather the resolution of cells in recordings, as this determines their SNR. The BER increase at 5~m is fundamentally due to inadequate cell resolution. 
Table \ref{tab:cell_dim} shows final BER is zero so long as cells occupy $\geq$ 35 x 35 pixels. The full projection region corresponding to this cell resolution is just 16\% of a 1080p frame.

\para{Recording environment and projection surface} \name\ achieves error-free embedding in all evaluated scenes (Table \ref{tab:surfs}), including dynamic outdoor environments and surfaces ranging from red brick (S6) to irregularly patterned, glossy paper (S3). We attribute this to \name's adaptive embedding procedure, which continually ensures sufficient SNR. We find \name\ achieves a final BER of zero in any scene where ambient light intensity does not dominate SLM-projected light. 
We measure this threshold value to be 4~klx (roughly 20x brighter than a typical indoor setting~\cite{luxlevels}).

\para{Recording device} Because \name\ encodes data as simple light intensity changes, it is naturally compatible with any modern RGB camera. Signatures are reliably extracted from videos captured on all five tested $\geq$ 1080p devices, from webcams to DSLRs (Table \ref{tab:device}).

\para{Video post-processing} Signatures remain decodable after varied forms of video post-processing, including compression (via reducing bitrate from the original 19k kbps), transcoding from the original H.265 codec (Table \ref{tab:compression}), and application of aesthetic filters and lighting adjustments in the iPhone Photos app (Table \ref{tab:aesthetic}).

\section{Countermeasures}

We explore countermeasures an attacker may employ in an attempt to create falsified content that nonetheless passes verification or otherwise disrupt \name\ operation. We assume the attacker has white-box access to all \name\ algorithms and models.

\subsection{Spoofing via Adversarial Examples}

Extensive prior works show that deep neural networks (DNNs) are vulnerable to adversarial examples -- carefully crafted inputs that look normal to the naked eye but cause models to make incorrect predictions~\cite{carlini2017towards}. An attacker may try to craft adversarial examples against the FaceMesh and ArcFace DNNs that \name\ uses to extract feature vectors (\S\ref{subsec:feature_extraction}). Concretely, their goal is to take a \name-protected video and create a falsified version that, although to the naked eye portrays a different identity or facial motion, elicits identity and dynamic features highly similar to those of the real video. If they achieve this, their fake video's feature vectors will possess locality-sensitive hashes similar to those in the original video signature. They can thus simply retain this signature in their fake video, and it will pass verification, \emph{spoofing} \name.

We demonstrate two approaches to creating such adversarial examples -- one in which the attacker specially trains their deepfake model and the other in which they perturb their videos post-factum -- and find neither succeeds. While we are under no illusion this means creating adversarial examples against \name\ is fundamentally impossible, we show it is in practice highly challenging. This significantly raises the bar for attack execution.

Below, we use the VoxCeleb video dataset~\cite{Nagrani19} for training and tests. Note that the LFW dataset, while favorable for our evaluations in \S\ref{subsec:digest_robustness} due to its greater identity and pose diversity, cannot be used in the following video-focused studies because it is image-based.

\para{Adversarial deepfake generation} In the first method, 
with white-box access to the FaceMesh and ArcFace models, the attacker directly incorporates the above signature spoofing objective into the loss function of their deepfake model. Specifically, during training, she can extract the \name\ feature vectors from generated videos and apply a penalty if those feature vectors are dissimilar from the original video feature vectors. If our descriptors are vulnerable to adversarial examples, the deepfake model should learn to satisfy the attack objective while still achieving the intended falsification. Note that adversarial examples must meet \emph{both} of these criteria. If those generated videos that pass verification simply resemble the real videos, \name\ is providing the expected protections.

We test this method using the FSGAN identity swap model and DaGAN reenactment model. We choose DaGAN as our representative reenactment model because its outputs were the hardest for passive detectors to detect (Table \ref{tab:aucs}). %
We modify the DaGAN and FSGAN loss functions to include a \emph{\name-spoofing term}, which applies the aforementioned penalty based on generated videos' dynamic (DaGAN) or identity (FSGAN) feature vectors. We train both models with three different weights on this term (empirically set to optimize attack success), and then use each version to generate 50 fake videos. We report the rate at which each model's generated videos pass \name\ verification. Details on our model implementations can be found in \ref{subsec:adv_training}.

\begin{figure}[t]
    \centering
    \includegraphics[width=\columnwidth]{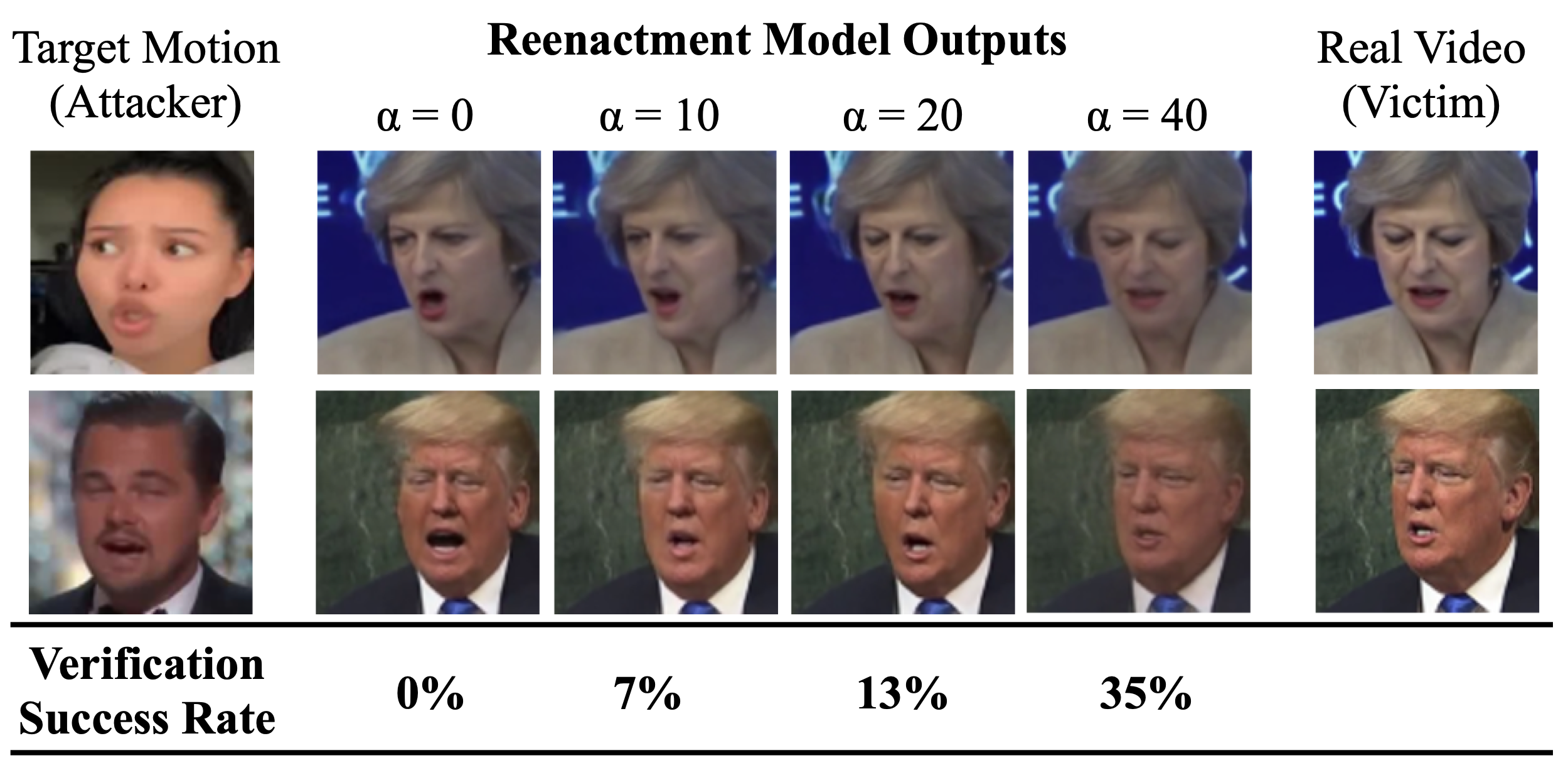}
        \vspace{-0.25in}
        \caption{Example DaGAN outputs after training with various weights $\alpha$ on the spoofing objective. Higher weights increase generated videos' verification success rates, but cause the model to reproduce the semantic content of the real video.} 
     \label{fig:dagan}
    \vspace{-0.2in}
\end{figure}

We find that none of our attack models can produce adversarial examples. When trained with sufficiently high weight on the \name-spoofing term, DaGAN learns to output content that passes \name\ verification by simply retaining the face and lip motion of the real video. Figure \ref{fig:dagan} illustrates this effect. We can see that while the original DaGAN model (spoofing weight $\alpha=0$) modifies the victim's facial movement according to the attacker-provided target, the outputs of the adversarial model with the highest success rate ($\alpha=40$) largely portray the same expressions as the real video frames, with some perceptual degradations. The FSGAN spoofing rate remained at 0\% across all tested weights on our term.

These behaviors arise from a clear contradiction between the \name-spoofing and original deepfake loss function components: while the former enforces similarity between the real and fake videos' content, the latter explicitly rewards real videos' modification. During training, we observe one component strictly dominates the other, depending on their relative weights. Even after extensive testing, we cannot find a weight at which both components simultaneously converge. Thus, \name's descriptors are adversarially robust; even when deeepfake models can directly backpropagate through its feature extractors in training, they cannot find loopholes enabling generation of adversarial examples.

\para{Adversarial perturbation of frames} The attacker may also add adversarial perturbations to video frames as a post-processing step, inspired by other perturbation-based attacks on vision DNNs~\cite{carlini2017towards}. 

We first apply this method to our identity features by adapting Fawkes~\cite{shan2020fawkes}, a white-box attack on face recognition models. Given a structural dis-similarity (DSSIM)~\cite{wang2004image} budget configuring permitted perturbation visibility, Fawkes perturbs a \emph{source} face image to shift its feature space representation towards that of a desired \emph{target} identity. 
We replace the Fawkes feature extractor with our identity feature extractor, consisting of the face detector and ArcFace model. 

We then randomly choose 22 source-target pairs of identities from VoxCeleb and use Fawkes to perturb all source video frames toward their targets.\footnote{We compute perturbations using the same parameters as the original Fawkes implementation~\cite{shan2020fawkes}: Adam optimizer for 1000 iterations with a learning rate of 0.5.} Because frames must be perturbed independently, this attack is highly computationally expensive (over 1~min \emph{per frame} for moderate GPUs). 
We report the rate at which perturbed source videos are successfully verified. We perform the experiment under four DSSIM budgets: 0.003, 0.005, 0.007, and 0.009. A larger budget enables larger feature space shifts -- boosting chances of success -- but introduces noticeable artifacts exposing the attack. Prior works perturbing face images use budgets from 0.003 to 0.007~\cite{wang2018great, shan2020fawkes} to remain invisible. Note, however, recent work~\cite{passananti2024disrupting} suggests perturbations at these budgets may be visible in \emph{videos} due to temporal incoherence across independently-perturbed frames.

We observe a verification success rate of zero for all budgets $\leq$ 0.007. Though Fawkes successfully perturbs a larger portion of frames per video as the budget approaches 0.007, it fails to succeed on \emph{all} frames. This is necessary for the video to pass verification, as \name\ validates all frames' identity features (\S\ref{sec:verification}). At the highest budget of 0.009, the success rate rises to 4.5\%. However, the frames exhibit visible artifacts, causing a distinct flickering effect when played. Thus, even with a budget exceeding prior perceptibility thresholds, the perturbation-based attack on identity features fails.

These results can be attributed to \name's particularly stringent identity feature verification threshold (configured in \S\ref{sec:prototype}), which forces perturbations to shift features by larger amounts to produce a spoof. \name's identity feature verification threshold is particularly strict because videos recorded at a speech site necessarily vary only in their viewpoint of the speaker, with other appearance variations that ArcFace is trained to accommodate (e.g., makeup, lighting) naturally constant. As a result, legitimate video identity features are generally highly similar to those disseminated by the core unit. This phenomenon is also reflected in \name's perfect AUC in detecting identity swap deepfakes in \S\ref{subsec:overall_performance} (Table \ref{tab:aucs}).

Beyond this empirical validation, several works show that with sufficient perturbed and unperturbed images of a subject, a defender can train a highly accurate adversarial perturbation detector (AUC > .997) that generalizes across perturbation methods~\cite{radiya2021data, metzen2017on}. For high-profile speakers, images for such training are abundant. Thus, \name\ can incorporate a detector to preemptively detect and reject perturbed videos. Given these findings, we leave exploration of perturbation-based attacks on dynamic features to future work.

\subsection{Other Countermeasures}
\label{subsec:other_countermeasures}

\para{Screen recording} An adversary may launch the following attack based on screen recording: they place the core unit in front of a screen displaying a fake video, record the outputted optical modulations, and then digitally overlay them on the fake video. This attack can be simply addressed by equipping the core unit with an existing depth-sensing tool~\cite{apple-facescan} to differentiate a 2D screen from a speaker's physical presence, which we leave to future work.

\para{Environment-level interference} An attacker present at a speech could interfere with \name\ by injecting light onto its projection surface. For this to be effective, interfering light must dominate SLM illumination (i.e., measure roughly 4~klx, based on  \S\ref{subsec:embedding_robustness}). Light of this intensity is quite visible; thus the attack can be detected and stopped at the scene. As further defenses, \name\ can project onto multiple surfaces and periodically randomize projection surfaces.

\section{Perceptibility Evaluation}
\label{sec:perc_eval}

We evaluate the perceptibility of \name's modulations both live and in video via a user study and perceptual metrics. We summarize findings below, with further details in \ref{subsec:extra_perc_eval}.

For our user study, we invited participants to each scene in Figure~\ref{fig:surfaces}. At each, they were asked to assess the projection surface during trials in which the core unit either performed embedding or was powered off as a control case. For each trial, they reported whether they believed optical modulations were present, and rated the obtrusiveness of any perceived modulations. We repeated the study with videos of the scenes. As shown in the bottom panels of Figure \ref{fig:perceptibility}, participants overwhelmingly performed no better than random at detecting \name\ operation, indicating its effective imperceptibility.  In the few cases users accurately detected operation, they uniformly reported low obtrusiveness. 

We also analyze videos using the learned perceptual loss (LPIPS) ~\cite{zhang2018unreasonable} metric. All videos possess LPIPS scores over ten times lower than the established LPIPS perceptibility threshold (Figure \ref{fig:perceptibility}).

\section{Discussion and Future Work}
\label{sec:limitations_future}

\para{Alternative embedding mechanisms}
\name\ is compatible with varied embedding methods. Its modulation scheme supports any cell shapes and layouts and can be realized via existing projectors or screens at events.
Future work will explore acoustic methods as well as optical embedding on non-planar and non-stationary surfaces, with the ultimate goal of projecting onto the speaker face.

\para{Alternative dynamic features} \name\ is compatible with diverse dynamic features. One alternative is the cryptographic hash of a window's script, extracted via real-time speech-to-text.
This would provide key semantic information and aid detection of fine-grained content falsifications (e.g., changing "do" to "do not"), though failing to capture speech speed or visual cues. Future work will pursue LSH-compatible features capturing additional audio characteristics (e.g., tone) and visual attributes. Ultimately, dynamic features should be chosen based on anticipated attacks, and LSH aids in including multiple, complementary features in a signature.

\para{Verifying the last speech window} To ensure all speech content is protected, a video must have a window of downtime at its end for embedding of the final signature. Minimizing window durations can mitigate this overhead. This can be achieved by increasing embedding bandwidth, in turn reducing the time needed to embed each signature. The current scheme's bandwidth can be boosted by increasing projection region size or adjusting the imperceptibility-SNR trade-off. Developing acoustic embedding methods for joint use with optical modulation is a further promising direction.

\para{Camera movement}  
The current \name\ implementation assumes recordings are taken on a still camera, a constraint imposed by the verification module's assumption that embedded signals are carried by the same pixels throughout a video. Future efforts will integrate established video stabilization~\cite{liu2011subspace} and inter-frame alignment~\cite{okuma2004automatic} methods into verification to allow camera motion.

\section*{Acknowledgments}
We thank our reviewers for their insightful feedback. This work is supported in part by the SEAS-KFAI Generative AI and Public Discourse Research program at Columbia University. 
\bibliographystyle{ACM-Reference-Format}
\balance
\bibliography{references}


\begin{thebibliography}{119}


\ifx \showCODEN    \undefined \def \showCODEN     #1{\unskip}     \fi
\ifx \showDOI      \undefined \def \showDOI       #1{#1}\fi
\ifx \showISBNx    \undefined \def \showISBNx     #1{\unskip}     \fi
\ifx \showISBNxiii \undefined \def \showISBNxiii  #1{\unskip}     \fi
\ifx \showISSN     \undefined \def \showISSN      #1{\unskip}     \fi
\ifx \showLCCN     \undefined \def \showLCCN      #1{\unskip}     \fi
\ifx \shownote     \undefined \def \shownote      #1{#1}          \fi
\ifx \showarticletitle \undefined \def \showarticletitle #1{#1}   \fi
\ifx \showURL      \undefined \def \showURL       {\relax}        \fi
\providecommand\bibfield[2]{#2}
\providecommand\bibinfo[2]{#2}
\providecommand\natexlab[1]{#1}
\providecommand\showeprint[2][]{arXiv:#2}

\bibitem[nan(2019)]%
        {nancy19}
 \bibinfo{year}{2019}\natexlab{}.
\newblock \bibinfo{title}{{Doctored Nancy Pelosi video highlights threat of
  "deepfake" tech}}.
\newblock \bibinfo{howpublished}{{CBS News}}.
\newblock


\bibitem[ult(2022)]%
        {ultra_light_face_detector}
 \bibinfo{year}{2022}\natexlab{}.
\newblock \bibinfo{title}{Ultra-Light-Fast-Generic-Face-Detector-1MB}.
\newblock \bibinfo{howpublished}{GitHub repository}.
\newblock
\urldef\tempurl%
\url{https://github.com/Linzaer/Ultra-Light-Fast-Generic-Face-Detector-1MB}
\showURL{%
\tempurl}


\bibitem[dag(2023)]%
        {dagan_imp}
 \bibinfo{year}{2023}\natexlab{}.
\newblock \bibinfo{title}{{CVPR2022-DaGAN}}.
\newblock
  \bibinfo{howpublished}{{\url{https://github.com/harlanhong/CVPR2022-DaGAN/tree/master?tab=readme-ov-file}}}.
\newblock


\bibitem[bid(2023)]%
        {biden23_draft}
 \bibinfo{year}{2023}\natexlab{}.
\newblock \bibinfo{title}{{Fact Check: Video of Joe Biden calling for a
  military draft was created with AI}}.
\newblock \bibinfo{howpublished}{{Reuters}}.
\newblock


\bibitem[app(2024)]%
        {apple-facescan}
 \bibinfo{year}{2024}\natexlab{}.
\newblock \bibinfo{title}{{Apple 3DFaceScan}}.
\newblock
  \bibinfo{howpublished}{{\url{https://apps.apple.com/us/app/3dfacescan-structure-sdk/id6473282888}}}.
\newblock


\bibitem[ado(2024)]%
        {adobe_cai}
 \bibinfo{year}{2024}\natexlab{}.
\newblock \bibinfo{title}{{Content Authenticity Initiative}}.
\newblock \bibinfo{howpublished}{\url{https://contentauthenticity.org/}}.
\newblock


\bibitem[har(2024a)]%
        {harris24_slur}
 \bibinfo{year}{2024}\natexlab{a}.
\newblock \bibinfo{title}{{Deepfake Kamala Harris slurs her lines}}.
\newblock \bibinfo{howpublished}{AI, Algorithmic and Automation Incidents and
  Controversies}.
\newblock


\bibitem[med(2024)]%
        {mediapipe}
 \bibinfo{year}{2024}\natexlab{}.
\newblock \bibinfo{title}{{Face landmark detection guide}}.
\newblock
  \bibinfo{howpublished}{\url{https://developers.google.com/mediapipe/solutions/vision/face_landmarker}}.
\newblock


\bibitem[arc(2024)]%
        {arcface_impl}
 \bibinfo{year}{2024}\natexlab{}.
\newblock \bibinfo{title}{{InsightFace: 2D and 3D Face Analysis Project}}.
\newblock
  \bibinfo{howpublished}{{\url{https://github.com/deepinsight/insightface/tree/master}}}.
\newblock


\bibitem[har(2024b)]%
        {harris24_slow}
 \bibinfo{year}{2024}\natexlab{b}.
\newblock \bibinfo{title}{{Old Kamala Harris footage manipulated to slow her
  speech}}.
\newblock \bibinfo{howpublished}{{AFP Factcheck}}.
\newblock


\bibitem[syn(2024)]%
        {synthid}
 \bibinfo{year}{2024}\natexlab{}.
\newblock \bibinfo{title}{{SynthID}}.
\newblock
  \bibinfo{howpublished}{\url{https://deepmind.google/technologies/synthid/}}.
\newblock


\bibitem[pyt(2025)]%
        {pytorchlightning_lpips}
 \bibinfo{year}{2025}\natexlab{}.
\newblock \bibinfo{title}{{Learned Perceptual Image Patch Similarity (LPIPS)}}.
\newblock
  \bibinfo{howpublished}{{\url{https://lightning.ai/docs/torchmetrics/stable/image/learned_perceptual_image_patch_similarity.html}}}.
\newblock


\bibitem[onn(2025)]%
        {onnx}
 \bibinfo{year}{2025}\natexlab{}.
\newblock \bibinfo{title}{{Open Neural Network Exchange}}.
\newblock \bibinfo{howpublished}{\url{https://onnx.ai/}}.
\newblock


\bibitem[tru(2025)]%
        {truepic}
 \bibinfo{year}{2025}\natexlab{}.
\newblock \bibinfo{title}{{Truepic}}.
\newblock \bibinfo{howpublished}{{\url{https://truepicvision.com/}}}.
\newblock


\bibitem[ver(2025a)]%
        {verilight_github}
 \bibinfo{year}{2025}\natexlab{a}.
\newblock \bibinfo{title}{{VeriLight Github}}.
\newblock
  \bibinfo{howpublished}{{\url{https://github.com/MobileX-CU/verilight}}}.
\newblock


\bibitem[ver(2025b)]%
        {verilight_zenodo}
 \bibinfo{year}{2025}\natexlab{b}.
\newblock \bibinfo{title}{{VeriLight Zenodo}}.
\newblock \bibinfo{howpublished}{{\url{https://zenodo.org/records/17063747}}}.
\newblock


\bibitem[Afchar et~al\mbox{.}(2018)]%
        {afchar2018mesonet}
\bibfield{author}{\bibinfo{person}{Darius Afchar}, \bibinfo{person}{Vincent
  Nozick}, \bibinfo{person}{Junichi Yamagishi}, {and} \bibinfo{person}{Isao
  Echizen}.} \bibinfo{year}{2018}\natexlab{}.
\newblock \showarticletitle{Mesonet: a compact facial video forgery detection
  network}. In \bibinfo{booktitle}{\emph{IEEE International Workshop on
  Information Forensics and Security}}.
\newblock


\bibitem[Agarwal et~al\mbox{.}(2005)]%
        {agarwal2005survey}
\bibfield{author}{\bibinfo{person}{Anubhav Agarwal}, \bibinfo{person}{CV
  Jawahar}, {and} \bibinfo{person}{PJ Narayanan}.}
  \bibinfo{year}{2005}\natexlab{}.
\newblock \showarticletitle{A survey of planar homography estimation
  techniques}.
\newblock \bibinfo{journal}{\emph{Centre for Visual Information Technology,
  Tech. Rep}} (\bibinfo{year}{2005}).
\newblock


\bibitem[Agarwal and Farid(2021)]%
        {agarwal2021auraldynamics}
\bibfield{author}{\bibinfo{person}{Shruti Agarwal} {and} \bibinfo{person}{Hany
  Farid}.} \bibinfo{year}{2021}\natexlab{}.
\newblock \showarticletitle{{Detecting Deep-Fake Videos from Aural and Oral
  Dynamics}}. In \bibinfo{booktitle}{\emph{Proc. of {CVPR} Workshops}}.
\newblock


\bibitem[Agarwal et~al\mbox{.}(2020a)]%
        {agarwal2020appearancebehavior}
\bibfield{author}{\bibinfo{person}{Shruti Agarwal}, \bibinfo{person}{Hany
  Farid}, \bibinfo{person}{Tarek El-Gaaly}, {and} \bibinfo{person}{Ser-Nam
  Lim}.} \bibinfo{year}{2020}\natexlab{a}.
\newblock \showarticletitle{{Detecting Deep-Fake Videos from Appearance and
  Behavior}}. In \bibinfo{booktitle}{\emph{IEEE International Workshop on
  Information Forensics and Security}}.
\newblock


\bibitem[Agarwal et~al\mbox{.}(2020b)]%
        {agarwal2020phoneviseme}
\bibfield{author}{\bibinfo{person}{Shruti Agarwal}, \bibinfo{person}{Hany
  Farid}, \bibinfo{person}{Ohad Fried}, {and} \bibinfo{person}{Maneesh
  Agrawala}.} \bibinfo{year}{2020}\natexlab{b}.
\newblock \showarticletitle{Detecting Deep-Fake Videos from Phoneme-Viseme
  Mismatches}. In \bibinfo{booktitle}{\emph{Proc. of {CVPR} Workshops}}.
  \bibinfo{pages}{2814--2822}.
\newblock


\bibitem[Agarwal et~al\mbox{.}(2019)]%
        {agarwal2019protecting}
\bibfield{author}{\bibinfo{person}{Shruti Agarwal}, \bibinfo{person}{Hany
  Farid}, \bibinfo{person}{Yuming Gu}, \bibinfo{person}{Mingming He},
  \bibinfo{person}{Koki Nagano}, {and} \bibinfo{person}{Hao Li}.}
  \bibinfo{year}{2019}\natexlab{}.
\newblock \showarticletitle{{Protecting World Leaders Against Deep Fakes}}. In
  \bibinfo{booktitle}{\emph{Proc. of {CVPR} Workshops}}.
\newblock


\bibitem[Agarwal et~al\mbox{.}(2023)]%
        {agarwal2023watch}
\bibfield{author}{\bibinfo{person}{Shruti Agarwal}, \bibinfo{person}{Liwen Hu},
  \bibinfo{person}{Evonne Ng}, \bibinfo{person}{Trevor Darrell},
  \bibinfo{person}{Hao Li}, {and} \bibinfo{person}{Anna Rohrbach}.}
  \bibinfo{year}{2023}\natexlab{}.
\newblock \showarticletitle{Watch those words: Video falsification detection
  using word-conditioned facial motion}. In \bibinfo{booktitle}{\emph{IEEE/CVF
  Winter Conference on Applications of Computer Vision}}.
\newblock


\bibitem[Air(2023)]%
        {luxlevels}
\bibfield{author}{\bibinfo{person}{Prana Air}.}
  \bibinfo{year}{2023}\natexlab{}.
\newblock \bibinfo{title}{{Illuminance Levels Indoors: Your Standard Lux Level
  Chart}}.
\newblock
  \bibinfo{howpublished}{\url{https://www.pranaair.com/blog/illuminance-levels-indoors-the-standard-lux-levels}}.
\newblock


\bibitem[{Arsha Nagrani and Joon~Son Chung and Weidi Xie and Andrew
  Zisserman}(2019)]%
        {Nagrani19}
\bibfield{author}{\bibinfo{person}{{Arsha Nagrani and Joon~Son Chung and Weidi
  Xie and Andrew Zisserman}}.} \bibinfo{year}{2019}\natexlab{}.
\newblock \showarticletitle{{VoxCeleb: Large-scale speaker verification in the
  wild}}.
\newblock \bibinfo{journal}{\emph{Computer Science and Language}}
  (\bibinfo{year}{2019}).
\newblock


\bibitem[Bonettini et~al\mbox{.}(2021)]%
        {bonettini2021video}
\bibfield{author}{\bibinfo{person}{Nicolo Bonettini},
  \bibinfo{person}{Edoardo~Daniele Cannas}, \bibinfo{person}{Sara Mandelli},
  \bibinfo{person}{Luca Bondi}, \bibinfo{person}{Paolo Bestagini}, {and}
  \bibinfo{person}{Stefano Tubaro}.} \bibinfo{year}{2021}\natexlab{}.
\newblock \showarticletitle{{Video Face Manipulation Detection Through Ensemble
  of CNNs}}. In \bibinfo{booktitle}{\emph{International Conference on Pattern
  Recognition}}.
\newblock


\bibitem[Cai et~al\mbox{.}(2022)]%
        {cai2022you}
\bibfield{author}{\bibinfo{person}{Zhixi Cai}, \bibinfo{person}{Kalin
  Stefanov}, \bibinfo{person}{Abhinav Dhall}, {and} \bibinfo{person}{Munawar
  Hayat}.} \bibinfo{year}{2022}\natexlab{}.
\newblock \showarticletitle{{Do You Really Mean That? Content Driven
  Audio-Visual Deepfake Dataset and Multimodal Method for Temporal Forgery
  Localization}}. In \bibinfo{booktitle}{\emph{International Conference on
  Digital Image Computing: Techniques and Applications}}.
\newblock


\bibitem[Cao et~al\mbox{.}(2022)]%
        {cao2022end}
\bibfield{author}{\bibinfo{person}{Junyi Cao}, \bibinfo{person}{Chao Ma},
  \bibinfo{person}{Taiping Yao}, \bibinfo{person}{Shen Chen},
  \bibinfo{person}{Shouhong Ding}, {and} \bibinfo{person}{Xiaokang Yang}.}
  \bibinfo{year}{2022}\natexlab{}.
\newblock \showarticletitle{End-to-end reconstruction-classification learning
  for face forgery detection}. In \bibinfo{booktitle}{\emph{Proc. of {CVPR}}}.
\newblock


\bibitem[Caria et~al\mbox{.}(2019)]%
        {caria2019exploring}
\bibfield{author}{\bibinfo{person}{Maria Caria}, \bibinfo{person}{Gabriele
  Sara}, \bibinfo{person}{Giuseppe Todde}, \bibinfo{person}{Marco Polese},
  {and} \bibinfo{person}{Antonio Pazzona}.} \bibinfo{year}{2019}\natexlab{}.
\newblock \showarticletitle{Exploring smart glasses for augmented reality: A
  valuable and integrative tool in precision livestock farming}.
\newblock \bibinfo{journal}{\emph{Animals}} \bibinfo{volume}{9},
  \bibinfo{number}{11} (\bibinfo{year}{2019}), \bibinfo{pages}{903}.
\newblock


\bibitem[Carlini and Farid(2020)]%
        {carlini2020evading}
\bibfield{author}{\bibinfo{person}{Nicholas Carlini} {and}
  \bibinfo{person}{Hany Farid}.} \bibinfo{year}{2020}\natexlab{}.
\newblock \showarticletitle{{Evading Deepfake-Image Detectors with White- and
  Black-Box Attacks}}.
\newblock \bibinfo{journal}{\emph{arXiv preprint arXiv:2004.00622}}
  (\bibinfo{year}{2020}).
\newblock


\bibitem[Carlini and Wagner(2017)]%
        {carlini2017towards}
\bibfield{author}{\bibinfo{person}{Nicholas Carlini} {and}
  \bibinfo{person}{David Wagner}.} \bibinfo{year}{2017}\natexlab{}.
\newblock \showarticletitle{Towards evaluating the robustness of neural
  networks}. In \bibinfo{booktitle}{\emph{IEEE Symposium on Security and
  Privacy}}.
\newblock


\bibitem[Charikar(2002)]%
        {charikar2002similarity}
\bibfield{author}{\bibinfo{person}{Moses~S Charikar}.}
  \bibinfo{year}{2002}\natexlab{}.
\newblock \showarticletitle{{Similarity estimation techniques from rounding
  algorithms}}. In \bibinfo{booktitle}{\emph{{Proc. of ACM Symposium on Theory
  of Computing}}}. \bibinfo{pages}{380--388}.
\newblock


\bibitem[Chen et~al\mbox{.}(2021)]%
        {chen2021defakehop}
\bibfield{author}{\bibinfo{person}{Hong-Shuo Chen}, \bibinfo{person}{Mozhdeh
  Rouhsedaghat}, \bibinfo{person}{Hamza Ghani}, \bibinfo{person}{Shuowen Hu},
  \bibinfo{person}{Suya You}, {and} \bibinfo{person}{C-C~Jay Kuo}.}
  \bibinfo{year}{2021}\natexlab{}.
\newblock \showarticletitle{{Defakehop: A light-weight high-performance
  deepfake detector}}. In \bibinfo{booktitle}{\emph{{IEEE International
  Conference on Multimedia and Expo}}}.
\newblock


\bibitem[Chen et~al\mbox{.}(2022)]%
        {pulsedit}
\bibfield{author}{\bibinfo{person}{Mingliang Chen}, \bibinfo{person}{Xin Liao},
  {and} \bibinfo{person}{Min Wu}.} \bibinfo{year}{2022}\natexlab{}.
\newblock \showarticletitle{{PulseEdit: Editing Physiological Signals in Facial
  Videos for Privacy Protection}}.
\newblock \bibinfo{journal}{\emph{IEEE Transactions on Information Forensics
  and Security}}  \bibinfo{volume}{17} (\bibinfo{year}{2022}),
  \bibinfo{pages}{457--471}.
\newblock


\bibitem[Chugh et~al\mbox{.}(2020)]%
        {chugh2020not}
\bibfield{author}{\bibinfo{person}{Komal Chugh}, \bibinfo{person}{Parul Gupta},
  \bibinfo{person}{Abhinav Dhall}, {and} \bibinfo{person}{Ramanathan
  Subramanian}.} \bibinfo{year}{2020}\natexlab{}.
\newblock \showarticletitle{Not made for each other-audio-visual
  dissonance-based deepfake detection and localization}. In
  \bibinfo{booktitle}{\emph{{ACM} International Conference on Multimedia}}.
\newblock


\bibitem[Ciftci et~al\mbox{.}(2020)]%
        {ciftci2020fakecatcher}
\bibfield{author}{\bibinfo{person}{Umur~Aybars Ciftci}, \bibinfo{person}{Ilke
  Demir}, {and} \bibinfo{person}{Lijun Yin}.} \bibinfo{year}{2020}\natexlab{}.
\newblock \showarticletitle{Fakecatcher: Detection of synthetic portrait videos
  using biological signals}.
\newblock \bibinfo{journal}{\emph{IEEE Transactions on Pattern Analysis and
  Machine Intelligence}} (\bibinfo{year}{2020}).
\newblock


\bibitem[Cotting et~al\mbox{.}(2004)]%
        {cotting2004embedding}
\bibfield{author}{\bibinfo{person}{Daniel Cotting}, \bibinfo{person}{Martin
  Naef}, \bibinfo{person}{Markus Gross}, {and} \bibinfo{person}{Henry Fuchs}.}
  \bibinfo{year}{2004}\natexlab{}.
\newblock \showarticletitle{{Embedding imperceptible patterns into projected
  images for simultaneous acquisition and display}}. In
  \bibinfo{booktitle}{\emph{{IEEE} International Symposium on Mixed and
  Augmented Reality}}.
\newblock


\bibitem[Cozzolino et~al\mbox{.}(2021)]%
        {cozzolino2021id}
\bibfield{author}{\bibinfo{person}{Davide Cozzolino}, \bibinfo{person}{Andreas
  R{\"o}ssler}, \bibinfo{person}{Justus Thies}, \bibinfo{person}{Matthias
  Nie{\ss}ner}, {and} \bibinfo{person}{Luisa Verdoliva}.}
  \bibinfo{year}{2021}\natexlab{}.
\newblock \showarticletitle{{ID-reveal: Identity-aware Deepfake Video
  Detection}}. In \bibinfo{booktitle}{\emph{International Conference on
  Computer Vision}}. \bibinfo{pages}{15108--15117}.
\newblock


\bibitem[Critch(2022)]%
        {critch2022wordsig}
\bibfield{author}{\bibinfo{person}{Andrew Critch}.}
  \bibinfo{year}{2022}\natexlab{}.
\newblock \showarticletitle{{WordSig: QR streams enabling platform-independent
  self-identification that's impossible to deepfake}}.
\newblock \bibinfo{journal}{\emph{arXiv preprint arXiv:2207.10806}}
  (\bibinfo{year}{2022}).
\newblock


\bibitem[Cui et~al\mbox{.}(2019)]%
        {unseencode}
\bibfield{author}{\bibinfo{person}{Hao Cui}, \bibinfo{person}{Huanyu Bian},
  \bibinfo{person}{Weiming Zhang}, {and} \bibinfo{person}{Nenghai Yu}.}
  \bibinfo{year}{2019}\natexlab{}.
\newblock \showarticletitle{{UnseenCode: Invisible On-screen Barcode with
  Image-based Extraction}}. In \bibinfo{booktitle}{\emph{Proc. of {INFOCOM}}}.
\newblock


\bibitem[D.~Cooper(2008)]%
        {cert_pki}
\bibfield{author}{\bibinfo{person}{S.~Farrell S. Boeyen R. Housley W.~Polk
  D.~Cooper, S.~Santesson}.} \bibinfo{year}{2008}\natexlab{}.
\newblock \bibinfo{title}{{RFC 5208 - Internet X.509 Public Key Infrastructure
  Certificate and Certificate Revocation List (CRL) Profile}}.
\newblock
  \bibinfo{howpublished}{\url{https://datatracker.ietf.org/doc/html/rfc5280}}.
\newblock


\bibitem[Deng et~al\mbox{.}(2019)]%
        {deng2019arcface}
\bibfield{author}{\bibinfo{person}{Jiankang Deng}, \bibinfo{person}{Jia Guo},
  \bibinfo{person}{Niannan Xue}, {and} \bibinfo{person}{Stefanos Zafeiriou}.}
  \bibinfo{year}{2019}\natexlab{}.
\newblock \showarticletitle{Arcface: Additive angular margin loss for deep face
  recognition}. In \bibinfo{booktitle}{\emph{Proc. of {CVPR}}}.
\newblock


\bibitem[Diffie and Hellman(1976)]%
        {diffie2022new}
\bibfield{author}{\bibinfo{person}{Whitfield Diffie} {and}
  \bibinfo{person}{Martin~E Hellman}.} \bibinfo{year}{1976}\natexlab{}.
\newblock \showarticletitle{New directions in cryptography}.
\newblock In \bibinfo{booktitle}{\emph{Democratizing Cryptography: The Work of
  Whitfield Diffie and Martin Hellman}}.
\newblock


\bibitem[Donovan and Scherer(2005)]%
        {Donovan_Scherer_2005}
\bibfield{author}{\bibinfo{person}{Robert~J. Donovan} {and}
  \bibinfo{person}{Ray Scherer}.} \bibinfo{year}{2005}\natexlab{}.
\newblock \bibinfo{booktitle}{\emph{Unsilent Revolution: Television News and
  American Public Life, 1948-1991}}.
\newblock \bibinfo{publisher}{Woodrow Wilson International Center for Scholars;
  Cambridge University Press}.
\newblock


\bibitem[Fang et~al\mbox{.}(2022)]%
        {tera}
\bibfield{author}{\bibinfo{person}{Han Fang}, \bibinfo{person}{Dongdong Chen},
  \bibinfo{person}{Feng Wang}, \bibinfo{person}{Zehua Ma},
  \bibinfo{person}{Honggu Liu}, \bibinfo{person}{Wenbo Zhou},
  \bibinfo{person}{Weiming Zhang}, {and} \bibinfo{person}{Nenghai Yu}.}
  \bibinfo{year}{2022}\natexlab{}.
\newblock \showarticletitle{{TERA: Screen-to-Camera Image Code With
  Transparency, Efficiency, Robustness and Adaptability}}.
\newblock \bibinfo{journal}{\emph{{IEEE} Transactions on Multimedia}}
  \bibinfo{volume}{24} (\bibinfo{year}{2022}), \bibinfo{pages}{955--967}.
\newblock


\bibitem[Farrukh et~al\mbox{.}(2020)]%
        {facerevelio}
\bibfield{author}{\bibinfo{person}{Habiba Farrukh},
  \bibinfo{person}{Reham~Mohamed Aburas}, \bibinfo{person}{Siyuan Cao}, {and}
  \bibinfo{person}{He Wang}.} \bibinfo{year}{2020}\natexlab{}.
\newblock \showarticletitle{{FaceRevelio: A Face Liveness Detection System for
  Smartphones with a Single Front Camera}}. In \bibinfo{booktitle}{\emph{Proc.
  of {MobiCom}}}.
\newblock


\bibitem[Ferri et~al\mbox{.}(1994)]%
        {FERRI1994403}
\bibfield{author}{\bibinfo{person}{F.J. Ferri}, \bibinfo{person}{P. Pudil},
  \bibinfo{person}{M. Hatef}, {and} \bibinfo{person}{J. Kittler}.}
  \bibinfo{year}{1994}\natexlab{}.
\newblock \showarticletitle{Comparative study of techniques for large-scale
  feature selection}.
\newblock In \bibinfo{booktitle}{\emph{Pattern Recognition in Practice IV}}.
  Vol.~\bibinfo{volume}{16}. \bibinfo{pages}{403--413}.
\newblock


\bibitem[Frank et~al\mbox{.}(2020)]%
        {spatial_freq_det}
\bibfield{author}{\bibinfo{person}{Joel Frank}, \bibinfo{person}{Thorsten
  Eisenhofer}, \bibinfo{person}{Lea Sch\"{o}nherr}, \bibinfo{person}{Asja
  Fischer}, \bibinfo{person}{Dorothea Kolossa}, {and} \bibinfo{person}{Thorsten
  Holz}.} \bibinfo{year}{2020}\natexlab{}.
\newblock \showarticletitle{{Leveraging Frequency Analysis for Deep Fake Image
  Recognition}}. In \bibinfo{booktitle}{\emph{International Conference on
  Machine Learning}}.
\newblock


\bibitem[Garofolo(1993)]%
        {garofolo1993timit}
\bibfield{author}{\bibinfo{person}{John~S Garofolo}.}
  \bibinfo{year}{1993}\natexlab{}.
\newblock \showarticletitle{{Timit Acoustic Phonetic Continuous Speech
  Corpus}}.
\newblock \bibinfo{journal}{\emph{Linguistic Data Consortium}}
  (\bibinfo{year}{1993}).
\newblock


\bibitem[Gerstner and Farid(2022)]%
        {gerstner2022detecting}
\bibfield{author}{\bibinfo{person}{Candice~R. Gerstner} {and}
  \bibinfo{person}{Hany Farid}.} \bibinfo{year}{2022}\natexlab{}.
\newblock \showarticletitle{{Detecting real-time deep-fake videos using active
  illumination}}. In \bibinfo{booktitle}{\emph{Proc. of {CVPR}}}.
  \bibinfo{pages}{53--60}.
\newblock


\bibitem[H.~Krawczyk(1997)]%
        {hmac_rfc}
\bibfield{author}{\bibinfo{person}{R.~Canetti H.~Krawczyk, M.~Bellare}.}
  \bibinfo{year}{1997}\natexlab{}.
\newblock \bibinfo{title}{{HMAC: Keyed-Hashing for Message Authentication}}.
\newblock
  \bibinfo{howpublished}{\url{https://datatracker.ietf.org/doc/html/rfc2104\#section-5}}.
\newblock


\bibitem[Haliassos et~al\mbox{.}(2021)]%
        {haliassos2021lips}
\bibfield{author}{\bibinfo{person}{Alexandros Haliassos},
  \bibinfo{person}{Konstantinos Vougioukas}, \bibinfo{person}{Stavros
  Petridis}, {and} \bibinfo{person}{Maja Pantic}.}
  \bibinfo{year}{2021}\natexlab{}.
\newblock \showarticletitle{Lips don't lie: A generalisable and robust approach
  to face forgery detection}. In \bibinfo{booktitle}{\emph{Proc. of {CVPR}}}.
\newblock


\bibitem[Hong et~al\mbox{.}(2022)]%
        {hong2022depth}
\bibfield{author}{\bibinfo{person}{Fa-Ting Hong}, \bibinfo{person}{Longhao
  Zhang}, \bibinfo{person}{Li Shen}, {and} \bibinfo{person}{Dan Xu}.}
  \bibinfo{year}{2022}\natexlab{}.
\newblock \showarticletitle{Depth-Aware Generative Adversarial Network for
  Talking Head Video Generation}.
\newblock \bibinfo{journal}{\emph{Proc. of {CVPR}}}.
\newblock


\bibitem[Hu et~al\mbox{.}(2021)]%
        {hu2021exposing}
\bibfield{author}{\bibinfo{person}{Shu Hu}, \bibinfo{person}{Yuezun Li}, {and}
  \bibinfo{person}{Siwei Lyu}.} \bibinfo{year}{2021}\natexlab{}.
\newblock \showarticletitle{Exposing GAN-generated faces using inconsistent
  corneal specular highlights}. In \bibinfo{booktitle}{\emph{IEEE International
  Conference on Acoustics, Speech and Signal Processing}}.
\newblock


\bibitem[Huang et~al\mbox{.}(2008)]%
        {huang2008labeled}
\bibfield{author}{\bibinfo{person}{Gary~B Huang}, \bibinfo{person}{Marwan
  Mattar}, \bibinfo{person}{Tamara Berg}, {and} \bibinfo{person}{Eric
  Learned-Miller}.} \bibinfo{year}{2008}\natexlab{}.
\newblock \showarticletitle{Labeled faces in the wild: A database forstudying
  face recognition in unconstrained environments}. In
  \bibinfo{booktitle}{\emph{{Workshop on faces in 'Real-Life' Images:
  detection, alignment, and recognition}}}.
\newblock


\bibitem[Jo et~al\mbox{.}(2016)]%
        {disco}
\bibfield{author}{\bibinfo{person}{Kensei Jo}, \bibinfo{person}{Mohit Gupta},
  {and} \bibinfo{person}{Shree~K. Nayar}.} \bibinfo{year}{2016}\natexlab{}.
\newblock \showarticletitle{{DisCo: Display-Camera Communication Using Rolling
  Shutter Sensors}}.
\newblock \bibinfo{journal}{\emph{ACM Trans. Graph.}} \bibinfo{volume}{35},
  \bibinfo{number}{5} (\bibinfo{year}{2016}).
\newblock


\bibitem[Khelifi and Bouridane(2017)]%
        {khelifi2017perceptual}
\bibfield{author}{\bibinfo{person}{Fouad Khelifi} {and} \bibinfo{person}{Ahmed
  Bouridane}.} \bibinfo{year}{2017}\natexlab{}.
\newblock \showarticletitle{Perceptual video hashing for content identification
  and authentication}.
\newblock \bibinfo{journal}{\emph{IEEE Transactions on Circuits and Systems for
  Video Technology}} \bibinfo{volume}{29}, \bibinfo{number}{1}
  (\bibinfo{year}{2017}), \bibinfo{pages}{50--67}.
\newblock


\bibitem[Laidlaw et~al\mbox{.}(2020)]%
        {laidlaw2020perceptual}
\bibfield{author}{\bibinfo{person}{Cassidy Laidlaw}, \bibinfo{person}{Sahil
  Singla}, {and} \bibinfo{person}{Soheil Feizi}.}
  \bibinfo{year}{2020}\natexlab{}.
\newblock \showarticletitle{Perceptual adversarial robustness: Defense against
  unseen threat models}.
\newblock \bibinfo{journal}{\emph{arXiv preprint arXiv:2006.12655}}
  (\bibinfo{year}{2020}).
\newblock


\bibitem[Lee et~al\mbox{.}(2015)]%
        {rollinglight}
\bibfield{author}{\bibinfo{person}{Hui-Yu Lee}, \bibinfo{person}{Hao-Min Lin},
  \bibinfo{person}{Yu-Lin Wei}, \bibinfo{person}{Hsin-I Wu},
  \bibinfo{person}{Hsin-Mu Tsai}, {and} \bibinfo{person}{Kate Ching-Ju Lin}.}
  \bibinfo{year}{2015}\natexlab{}.
\newblock \showarticletitle{{RollingLight: Enabling Line-of-Sight
  Light-to-Camera Communications}}. In \bibinfo{booktitle}{\emph{Proc. of
  {MobiSys}}}.
\newblock


\bibitem[Li et~al\mbox{.}(2019)]%
        {blending_artifacts}
\bibfield{author}{\bibinfo{person}{Lingzhi Li}, \bibinfo{person}{Jianmin Bao},
  \bibinfo{person}{Ting Zhang}, \bibinfo{person}{Hao Yang},
  \bibinfo{person}{Dong Chen}, \bibinfo{person}{Fang Wen}, {and}
  \bibinfo{person}{Baining Guo}.} \bibinfo{year}{2019}\natexlab{}.
\newblock \showarticletitle{{Face X-Ray for More General Face Forgery
  Detection}}.
\newblock \bibinfo{journal}{\emph{Proc. of {CVPR}}} (\bibinfo{year}{2019}).
\newblock


\bibitem[Li et~al\mbox{.}(2015)]%
        {hilight}
\bibfield{author}{\bibinfo{person}{Tianxing Li}, \bibinfo{person}{Chuankai An},
  \bibinfo{person}{Xinran Xiao}, \bibinfo{person}{Andrew~T. Campbell}, {and}
  \bibinfo{person}{Xia Zhou}.} \bibinfo{year}{2015}\natexlab{}.
\newblock \showarticletitle{{Real-Time Screen-Camera Communication Behind Any
  Scene}}. In \bibinfo{booktitle}{\emph{Proc. of {MobiSys}}}.
\newblock


\bibitem[Li et~al\mbox{.}(2018)]%
        {li2018ictu}
\bibfield{author}{\bibinfo{person}{Yuezun Li}, \bibinfo{person}{Ming-Ching
  Chang}, {and} \bibinfo{person}{Siwei Lyu}.} \bibinfo{year}{2018}\natexlab{}.
\newblock \showarticletitle{In ictu oculi: Exposing ai created fake videos by
  detecting eye blinking}. In \bibinfo{booktitle}{\emph{IEEE International
  Workshop on Information Forensics and Security}}.
\newblock


\bibitem[Liu et~al\mbox{.}(2011)]%
        {liu2011subspace}
\bibfield{author}{\bibinfo{person}{Feng Liu}, \bibinfo{person}{Michael
  Gleicher}, \bibinfo{person}{Jue Wang}, \bibinfo{person}{Hailin Jin}, {and}
  \bibinfo{person}{Aseem Agarwala}.} \bibinfo{year}{2011}\natexlab{}.
\newblock \showarticletitle{Subspace video stabilization}.
\newblock \bibinfo{journal}{\emph{{ACM} Trans. on Graphics}}
  \bibinfo{volume}{30}, \bibinfo{number}{1} (\bibinfo{year}{2011}),
  \bibinfo{pages}{1--10}.
\newblock


\bibitem[Liu et~al\mbox{.}(2021)]%
        {liu2021spatial}
\bibfield{author}{\bibinfo{person}{Honggu Liu}, \bibinfo{person}{Xiaodan Li},
  \bibinfo{person}{Wenbo Zhou}, \bibinfo{person}{Yuefeng Chen},
  \bibinfo{person}{Yuan He}, \bibinfo{person}{Hui Xue},
  \bibinfo{person}{Weiming Zhang}, {and} \bibinfo{person}{Nenghai Yu}.}
  \bibinfo{year}{2021}\natexlab{}.
\newblock \showarticletitle{Spatial-phase shallow learning: rethinking face
  forgery detection in frequency domain}. In \bibinfo{booktitle}{\emph{Proc. of
  {CVPR}}}.
\newblock


\bibitem[Liu et~al\mbox{.}(2020)]%
        {livescreen}
\bibfield{author}{\bibinfo{person}{Hongbo Liu}, \bibinfo{person}{Zhihua Li},
  \bibinfo{person}{Yucheng Xie}, \bibinfo{person}{Ruizhe Jiang},
  \bibinfo{person}{Yan Wang}, \bibinfo{person}{Xiaonan Guo}, {and}
  \bibinfo{person}{Yingying Chen}.} \bibinfo{year}{2020}\natexlab{}.
\newblock \showarticletitle{{LiveScreen: Video Chat Liveness Detection
  Leveraging Skin Reflection}}. In \bibinfo{booktitle}{\emph{Proc. of
  {INFOCOM}}}.
\newblock


\bibitem[Liu et~al\mbox{.}(2022)]%
        {liu2022vronicle}
\bibfield{author}{\bibinfo{person}{Yuxin Liu}, \bibinfo{person}{Yoshimichi
  Nakatsuka}, \bibinfo{person}{Ardalan~Amiri Sani}, \bibinfo{person}{Sharad
  Agarwal}, {and} \bibinfo{person}{Gene Tsudik}.}
  \bibinfo{year}{2022}\natexlab{}.
\newblock \showarticletitle{Vronicle: verifiable provenance for videos from
  mobile devices}. In \bibinfo{booktitle}{\emph{Proc. of {MobiSys}}}.
\newblock


\bibitem[Liu et~al\mbox{.}(2024)]%
        {liu24}
\bibfield{author}{\bibinfo{person}{Yuxin~(Myles) Liu}, \bibinfo{person}{Zhihao
  Yao}, \bibinfo{person}{Mingyi Chen}, \bibinfo{person}{Ardalan Amiri~Sani},
  \bibinfo{person}{Sharad Agarwal}, {and} \bibinfo{person}{Gene Tsudik}.}
  \bibinfo{year}{2024}\natexlab{}.
\newblock \showarticletitle{{ProvCam: A Camera Module with Self-Contained TCB
  for Producing Verifiable Videos}}. In \bibinfo{booktitle}{\emph{Proc. of
  {MobiCom}}}.
\newblock


\bibitem[Lugstein et~al\mbox{.}(2021)]%
        {lugstein2021prnu}
\bibfield{author}{\bibinfo{person}{Florian Lugstein}, \bibinfo{person}{Simon
  Baier}, \bibinfo{person}{Gregor Bachinger}, {and} \bibinfo{person}{Andreas
  Uhl}.} \bibinfo{year}{2021}\natexlab{}.
\newblock \showarticletitle{{PRNU-based deepfake detection}}. In
  \bibinfo{booktitle}{\emph{{Proc. of ACM Workshop on Information Hiding and
  Multimedia Security}}}.
\newblock


\bibitem[Luo et~al\mbox{.}(2021)]%
        {luo2021generalizing}
\bibfield{author}{\bibinfo{person}{Yuchen Luo}, \bibinfo{person}{Yong Zhang},
  \bibinfo{person}{Junchi Yan}, {and} \bibinfo{person}{Wei Liu}.}
  \bibinfo{year}{2021}\natexlab{}.
\newblock \showarticletitle{Generalizing face forgery detection with
  high-frequency features}. In \bibinfo{booktitle}{\emph{Proc. of {CVPR}}}.
  \bibinfo{pages}{16317--16326}.
\newblock


\bibitem[Lyon(2023)]%
        {leica_cam}
\bibfield{author}{\bibinfo{person}{Santiago Lyon}.}
  \bibinfo{year}{2023}\natexlab{}.
\newblock \bibinfo{title}{{Leica Launches World’s First Camera with Content
  Credentials}}.
\newblock
  \bibinfo{howpublished}{\url{https://contentauthenticity.org/blog/leica-launches-worlds-first-camera-with-content-credentials}}.
\newblock


\bibitem[Masi et~al\mbox{.}(2020)]%
        {masi2020two}
\bibfield{author}{\bibinfo{person}{Iacopo Masi}, \bibinfo{person}{Aditya
  Killekar}, \bibinfo{person}{Royston~Marian Mascarenhas},
  \bibinfo{person}{Shenoy~Pratik Gurudatt}, {and} \bibinfo{person}{Wael
  AbdAlmageed}.} \bibinfo{year}{2020}\natexlab{}.
\newblock \showarticletitle{Two-branch recurrent network for isolating
  deepfakes in videos}. In \bibinfo{booktitle}{\emph{European Conference on
  Computer Vision}}.
\newblock


\bibitem[McCarthy(2024)]%
        {biden24_lipsync}
\bibfield{author}{\bibinfo{person}{Bill McCarthy}.}
  \bibinfo{year}{2024}\natexlab{}.
\newblock \bibinfo{title}{{Video of Biden botching Ukraine history is a
  deepfake}}.
\newblock
  \bibinfo{howpublished}{{\url{https://factcheck.afp.com/doc.afp.com.34LY8TK}}}.
\newblock


\bibitem[Metzen et~al\mbox{.}(2017)]%
        {metzen2017on}
\bibfield{author}{\bibinfo{person}{Jan~Hendrik Metzen}, \bibinfo{person}{Tim
  Genewein}, \bibinfo{person}{Volker Fischer}, {and} \bibinfo{person}{Bastian
  Bischoff}.} \bibinfo{year}{2017}\natexlab{}.
\newblock \showarticletitle{On Detecting Adversarial Perturbations}. In
  \bibinfo{booktitle}{\emph{International Conference on Learning
  Representations}}.
\newblock


\bibitem[Michael et~al\mbox{.}(2025)]%
        {michael2025noise}
\bibfield{author}{\bibinfo{person}{Peter Michael}, \bibinfo{person}{Zekun Hao},
  \bibinfo{person}{Serge Belongie}, {and} \bibinfo{person}{Abe Davis}.}
  \bibinfo{year}{2025}\natexlab{}.
\newblock \showarticletitle{Noise-Coded Illumination for Forensic and
  Photometric Video Analysis}.
\newblock \bibinfo{journal}{\emph{{ACM} Trans. on Graphics}}
  \bibinfo{volume}{44}, \bibinfo{number}{5} (\bibinfo{year}{2025}),
  \bibinfo{pages}{1--16}.
\newblock


\bibitem[Mittal et~al\mbox{.}(2020)]%
        {mittal2020emotions}
\bibfield{author}{\bibinfo{person}{Trisha Mittal}, \bibinfo{person}{Uttaran
  Bhattacharya}, \bibinfo{person}{Rohan Chandra}, \bibinfo{person}{Aniket
  Bera}, {and} \bibinfo{person}{Dinesh Manocha}.}
  \bibinfo{year}{2020}\natexlab{}.
\newblock \showarticletitle{Emotions don't lie: An audio-visual deepfake
  detection method using affective cues}. In \bibinfo{booktitle}{\emph{{ACM}
  International Conference on Multimedia}}.
\newblock


\bibitem[Neekhara et~al\mbox{.}(2022)]%
        {neekhara2022facesigns}
\bibfield{author}{\bibinfo{person}{Paarth Neekhara}, \bibinfo{person}{Shehzeen
  Hussain}, \bibinfo{person}{Xinqiao Zhang}, \bibinfo{person}{Ke Huang},
  \bibinfo{person}{Julian McAuley}, {and} \bibinfo{person}{Farinaz
  Koushanfar}.} \bibinfo{year}{2022}\natexlab{}.
\newblock \bibinfo{title}{{FaceSigns: Semi-Fragile Neural Watermarks for Media
  Authentication and Countering Deepfakes}}.
\newblock
\newblock
\showeprint[arxiv]{2204.01960}


\bibitem[Nguyen et~al\mbox{.}(2019)]%
        {nguyen2019capsule}
\bibfield{author}{\bibinfo{person}{Huy~H Nguyen}, \bibinfo{person}{Junichi
  Yamagishi}, {and} \bibinfo{person}{Isao Echizen}.}
  \bibinfo{year}{2019}\natexlab{}.
\newblock \showarticletitle{Capsule-forensics: Using capsule networks to detect
  forged images and videos}. In \bibinfo{booktitle}{\emph{IEEE International
  Conference on Acoustics, Speech and Signal Processing}}.
\newblock


\bibitem[Nirkin et~al\mbox{.}(2019)]%
        {nirkin2019fsgan}
\bibfield{author}{\bibinfo{person}{Yuval Nirkin}, \bibinfo{person}{Yosi
  Keller}, {and} \bibinfo{person}{Tal Hassner}.}
  \bibinfo{year}{2019}\natexlab{}.
\newblock \showarticletitle{{FSGAN: Subject agnostic face swapping and
  reenactment}}. In \bibinfo{booktitle}{\emph{Proc. of {CVPR}}}.
\newblock


\bibitem[Okuma et~al\mbox{.}(2004)]%
        {okuma2004automatic}
\bibfield{author}{\bibinfo{person}{Kenji Okuma}, \bibinfo{person}{James~J
  Little}, {and} \bibinfo{person}{David~G Lowe}.}
  \bibinfo{year}{2004}\natexlab{}.
\newblock \showarticletitle{Automatic Rectification of Long Image Sequences}.
  In \bibinfo{booktitle}{\emph{Asian Conference on Computer Vision}}.
\newblock


\bibitem[Passananti et~al\mbox{.}(2024)]%
        {passananti2024disrupting}
\bibfield{author}{\bibinfo{person}{Josephine Passananti},
  \bibinfo{person}{Stanley Wu}, \bibinfo{person}{Shawn Shan},
  \bibinfo{person}{Haitao Zheng}, {and} \bibinfo{person}{Ben~Y Zhao}.}
  \bibinfo{year}{2024}\natexlab{}.
\newblock \showarticletitle{Disrupting style mimicry attacks on video imagery}.
\newblock \bibinfo{journal}{\emph{arXiv preprint arXiv:2405.06865}}
  (\bibinfo{year}{2024}).
\newblock


\bibitem[Qi et~al\mbox{.}(2020)]%
        {qi2020deeprhythm}
\bibfield{author}{\bibinfo{person}{Hua Qi}, \bibinfo{person}{Qing Guo},
  \bibinfo{person}{Felix Juefei-Xu}, \bibinfo{person}{Xiaofei Xie},
  \bibinfo{person}{Lei Ma}, \bibinfo{person}{Wei Feng}, \bibinfo{person}{Yang
  Liu}, {and} \bibinfo{person}{Jianjun Zhao}.} \bibinfo{year}{2020}\natexlab{}.
\newblock \showarticletitle{Deeprhythm: Exposing deepfakes with attentional
  visual heartbeat rhythms}. In \bibinfo{booktitle}{\emph{{ACM} International
  Conference on Multimedia}}.
\newblock


\bibitem[Qureshi et~al\mbox{.}(2021)]%
        {qureshi2021detecting}
\bibfield{author}{\bibinfo{person}{Amna Qureshi}, \bibinfo{person}{David
  Meg{\'\i}as}, {and} \bibinfo{person}{Minoru Kuribayashi}.}
  \bibinfo{year}{2021}\natexlab{}.
\newblock \showarticletitle{Detecting deepfake videos using digital
  watermarking}. In \bibinfo{booktitle}{\emph{Asia-Pacific Signal and
  Information Processing Association Annual Summit and Conference}}.
  \bibinfo{pages}{1786--1793}.
\newblock


\bibitem[Radiya-Dixit et~al\mbox{.}(2021)]%
        {radiya2021data}
\bibfield{author}{\bibinfo{person}{Evani Radiya-Dixit},
  \bibinfo{person}{Sanghyun Hong}, \bibinfo{person}{Nicholas Carlini}, {and}
  \bibinfo{person}{Florian Tram{\`e}r}.} \bibinfo{year}{2021}\natexlab{}.
\newblock \showarticletitle{Data poisoning won't save you from facial
  recognition}. In \bibinfo{booktitle}{\emph{International Conference on
  Learning Representations}}.
\newblock


\bibitem[Rossler et~al\mbox{.}(2019)]%
        {rossler2019faceforensics++}
\bibfield{author}{\bibinfo{person}{Andreas Rossler}, \bibinfo{person}{Davide
  Cozzolino}, \bibinfo{person}{Luisa Verdoliva}, \bibinfo{person}{Christian
  Riess}, \bibinfo{person}{Justus Thies}, {and} \bibinfo{person}{Matthias
  Nie{\ss}ner}.} \bibinfo{year}{2019}\natexlab{}.
\newblock \showarticletitle{Faceforensics++: Learning to Detect Manipulated
  Facial Images}. In \bibinfo{booktitle}{\emph{International Conference on
  Computer Vision}}.
\newblock


\bibitem[Sankaranarayanan et~al\mbox{.}(2021)]%
        {sankaranarayanan2021presidential}
\bibfield{author}{\bibinfo{person}{Aruna Sankaranarayanan},
  \bibinfo{person}{Matthew Groh}, \bibinfo{person}{Rosalind Picard}, {and}
  \bibinfo{person}{Andrew Lippman}.} \bibinfo{year}{2021}\natexlab{}.
\newblock \showarticletitle{The presidential deepfakes dataset}. In
  \bibinfo{booktitle}{\emph{{CEUR Workshop Proceedings}}}.
\newblock


\bibitem[Shahid and Roy(2023)]%
        {shahid2023my}
\bibfield{author}{\bibinfo{person}{Irtaza Shahid} {and}
  \bibinfo{person}{Nirupam Roy}.} \bibinfo{year}{2023}\natexlab{}.
\newblock \showarticletitle{{" Is this my president speaking?" Tamper-proofing
  Speech in Live Recordings}}. In \bibinfo{booktitle}{\emph{Proc. of
  {MobiSys}}}.
\newblock


\bibitem[Shan et~al\mbox{.}(2020)]%
        {shan2020fawkes}
\bibfield{author}{\bibinfo{person}{Shawn Shan}, \bibinfo{person}{Emily Wenger},
  \bibinfo{person}{Jiayun Zhang}, \bibinfo{person}{Huiying Li},
  \bibinfo{person}{Haitao Zheng}, {and} \bibinfo{person}{Ben~Y Zhao}.}
  \bibinfo{year}{2020}\natexlab{}.
\newblock \showarticletitle{{Fawkes: Protecting privacy against unauthorized
  deep learning models}}. In \bibinfo{booktitle}{\emph{29th USENIX Security
  Symposium}}.
\newblock


\bibitem[Shang and Wu(2020)]%
        {shangrealtime}
\bibfield{author}{\bibinfo{person}{Jiacheng Shang} {and} \bibinfo{person}{Jie
  Wu}.} \bibinfo{year}{2020}\natexlab{}.
\newblock \showarticletitle{{Protecting Real-time Video Chat against Fake
  Facial Videos Generated by Face Reenactment}}. In
  \bibinfo{booktitle}{\emph{International Conference on Distributed Computing
  Systems}}.
\newblock


\bibitem[Sharma et~al\mbox{.}(2005)]%
        {sharma2005ciede2000}
\bibfield{author}{\bibinfo{person}{Gaurav Sharma}, \bibinfo{person}{Wencheng
  Wu}, {and} \bibinfo{person}{Edul~N Dalal}.} \bibinfo{year}{2005}\natexlab{}.
\newblock \showarticletitle{{The CIEDE2000 color-difference formula:
  Implementation notes, supplementary test data, and mathematical
  observations}}.
\newblock \bibinfo{journal}{\emph{Color Research \& Application}}
  \bibinfo{volume}{30}, \bibinfo{number}{1} (\bibinfo{year}{2005}),
  \bibinfo{pages}{21--30}.
\newblock


\bibitem[Siarohin et~al\mbox{.}(2019)]%
        {siarohin2019first}
\bibfield{author}{\bibinfo{person}{Aliaksandr Siarohin},
  \bibinfo{person}{St{\'e}phane Lathuili{\`e}re}, \bibinfo{person}{Sergey
  Tulyakov}, \bibinfo{person}{Elisa Ricci}, {and} \bibinfo{person}{Nicu Sebe}.}
  \bibinfo{year}{2019}\natexlab{}.
\newblock \showarticletitle{First order motion model for image animation}. In
  \bibinfo{booktitle}{\emph{Advances in Neural Information Processing
  Systems}}.
\newblock


\bibitem[Siegel(1965)]%
        {colordiscriminationvtime}
\bibfield{author}{\bibinfo{person}{Michael~H. Siegel}.}
  \bibinfo{year}{1965}\natexlab{}.
\newblock \showarticletitle{Color Discrimination as a Function of Exposure
  Time}.
\newblock \bibinfo{journal}{\emph{Journal of the Optical Society of America}}
  \bibinfo{volume}{55}, \bibinfo{number}{5} (\bibinfo{date}{May}
  \bibinfo{year}{1965}), \bibinfo{pages}{566--568}.
\newblock


\bibitem[Sundar et~al\mbox{.}(2021)]%
        {sudar2021}
\bibfield{author}{\bibinfo{person}{S~Shyam Sundar}, \bibinfo{person}{Maria~D
  Molina}, {and} \bibinfo{person}{Eugene Cho}.}
  \bibinfo{year}{2021}\natexlab{}.
\newblock \showarticletitle{{Seeing Is Believing: Is Video Modality More
  Powerful in Spreading Fake News via Online Messaging Apps?}}
\newblock \bibinfo{journal}{\emph{Journal of Computer-Mediated Communication}}
  \bibinfo{volume}{26}, \bibinfo{number}{6} (\bibinfo{date}{08}
  \bibinfo{year}{2021}), \bibinfo{pages}{301--319}.
\newblock


\bibitem[Suzuki(1985)]%
        {suzuki1985topological}
\bibfield{author}{\bibinfo{person}{Satoshi Suzuki}.}
  \bibinfo{year}{1985}\natexlab{}.
\newblock \showarticletitle{Topological structural analysis of digitized binary
  images by border following}.
\newblock \bibinfo{journal}{\emph{Computer Vision, Graphics, and Image
  Processing}} \bibinfo{volume}{30}, \bibinfo{number}{1}
  (\bibinfo{year}{1985}), \bibinfo{pages}{32--46}.
\newblock


\bibitem[Szafir(2017)]%
        {colordiscriminationvsize}
\bibfield{author}{\bibinfo{person}{Danielle Szafir}.}
  \bibinfo{year}{2017}\natexlab{}.
\newblock \showarticletitle{Effects of Size and Shape on Perceived Color
  Differences}.
\newblock \bibinfo{journal}{\emph{Journal of Vision}} \bibinfo{volume}{17},
  \bibinfo{number}{10} (\bibinfo{year}{2017}), \bibinfo{pages}{1189}.
\newblock


\bibitem[Tan and Le(2019)]%
        {tan2019efficientnet}
\bibfield{author}{\bibinfo{person}{Mingxing Tan} {and} \bibinfo{person}{Quoc
  Le}.} \bibinfo{year}{2019}\natexlab{}.
\newblock \showarticletitle{Efficientnet: Rethinking model scaling for
  convolutional neural networks}. In \bibinfo{booktitle}{\emph{International
  Conference on Machine Learning}}.
\newblock


\bibitem[Unno and Uehira(2020)]%
        {kazutake_opticalwatermarking3}
\bibfield{author}{\bibinfo{person}{Hiroshi Unno} {and}
  \bibinfo{person}{Kazutake Uehira}.} \bibinfo{year}{2020}\natexlab{}.
\newblock \showarticletitle{{Lighting Technique for Attaching Invisible
  Information Onto Real Objects Using Temporally and Spatially Color-Intensity
  Modulated Light}}.
\newblock \bibinfo{journal}{\emph{IEEE Transactions on Industry Applications}}
  \bibinfo{volume}{56}, \bibinfo{number}{6} (\bibinfo{year}{2020}),
  \bibinfo{pages}{7202--7207}.
\newblock


\bibitem[Vlku(2002)]%
        {first_video_speeches}
\bibfield{author}{\bibinfo{person}{Nick Vlku}.}
  \bibinfo{year}{2002}\natexlab{}.
\newblock \bibinfo{title}{{The History of Television}}.
\newblock
  \bibinfo{howpublished}{\url{https://www.cs.cornell.edu/~pjs54/Teaching/AutomaticLifestyle-S02/Projects/Vlku/history.html}}.
\newblock


\bibitem[Wade(2023)]%
        {hadid23}
\bibfield{author}{\bibinfo{person}{Natalie Wade}.}
  \bibinfo{year}{2023}\natexlab{}.
\newblock \bibinfo{title}{{Deepfake of Bella Hadid misrepresents her statements
  on Israel}}.
\newblock \bibinfo{howpublished}{{AFP Fact Check}}.
\newblock


\bibitem[Wang et~al\mbox{.}(2015)]%
        {inframe++}
\bibfield{author}{\bibinfo{person}{Anran Wang}, \bibinfo{person}{Zhuoran Li},
  \bibinfo{person}{Chunyi Peng}, \bibinfo{person}{Guobin Shen},
  \bibinfo{person}{Gan Fang}, {and} \bibinfo{person}{Bing Zeng}.}
  \bibinfo{year}{2015}\natexlab{}.
\newblock \showarticletitle{{InFrame++: Achieve Simultaneous Screen-Human
  Viewing and Hidden Screen-Camera Communication}}. In
  \bibinfo{booktitle}{\emph{Proc. of {MobiSys}}}.
\newblock


\bibitem[Wang et~al\mbox{.}(2018b)]%
        {wang2018great}
\bibfield{author}{\bibinfo{person}{Bolun Wang}, \bibinfo{person}{Yuanshun Yao},
  \bibinfo{person}{Bimal Viswanath}, \bibinfo{person}{Haitao Zheng}, {and}
  \bibinfo{person}{Ben~Y Zhao}.} \bibinfo{year}{2018}\natexlab{b}.
\newblock \showarticletitle{With great training comes great vulnerability:
  Practical attacks against transfer learning}. In
  \bibinfo{booktitle}{\emph{27th USENIX Security Symposium}}.
\newblock


\bibitem[Wang et~al\mbox{.}(2023b)]%
        {wang2023talklip}
\bibfield{author}{\bibinfo{person}{Jiadong Wang}, \bibinfo{person}{Xinyuan
  Qian}, \bibinfo{person}{Malu Zhang}, \bibinfo{person}{Robby~T Tan}, {and}
  \bibinfo{person}{Haizhou Li}.} \bibinfo{year}{2023}\natexlab{b}.
\newblock \showarticletitle{{Seeing What You Said: Talking Face Generation
  Guided by a Lip Reading Expert}}. In \bibinfo{booktitle}{\emph{Proc. of
  {CVPR}}}.
\newblock


\bibitem[Wang et~al\mbox{.}(2018a)]%
        {wang2018attention}
\bibfield{author}{\bibinfo{person}{Qiongqiong Wang}, \bibinfo{person}{Koji
  Okabe}, \bibinfo{person}{Kong~Aik Lee}, \bibinfo{person}{Hitoshi Yamamoto},
  {and} \bibinfo{person}{Takafumi Koshinaka}.}
  \bibinfo{year}{2018}\natexlab{a}.
\newblock \showarticletitle{Attention Mechanism in Speaker Recognition: What
  Does it Learn in Deep Speaker Embedding?}. In \bibinfo{booktitle}{\emph{2018
  IEEE Spoken Language Technology Workshop}}. \bibinfo{pages}{1052--1059}.
\newblock


\bibitem[Wang et~al\mbox{.}(2021)]%
        {wang2021faketagger}
\bibfield{author}{\bibinfo{person}{Run Wang}, \bibinfo{person}{Felix
  Juefei-Xu}, \bibinfo{person}{Meng Luo}, \bibinfo{person}{Yang Liu}, {and}
  \bibinfo{person}{Lina Wang}.} \bibinfo{year}{2021}\natexlab{}.
\newblock \showarticletitle{Faketagger: Robust safeguards against deepfake
  dissemination via provenance tracking}. In \bibinfo{booktitle}{\emph{{ACM}
  International Conference on Multimedia}}.
\newblock


\bibitem[Wang et~al\mbox{.}(2020)]%
        {spotfornow}
\bibfield{author}{\bibinfo{person}{Sheng{-}Yu Wang}, \bibinfo{person}{Oliver
  Wang}, \bibinfo{person}{Richard Zhang}, \bibinfo{person}{Andrew Owens}, {and}
  \bibinfo{person}{Alexei~A. Efros}.} \bibinfo{year}{2020}\natexlab{}.
\newblock \showarticletitle{{CNN-Generated Images Are Surprisingly Easy to
  Spot... for Now}}. In \bibinfo{booktitle}{\emph{Proc. of {CVPR}}}.
\newblock


\bibitem[Wang et~al\mbox{.}(2023a)]%
        {Wang_2023_CVPR}
\bibfield{author}{\bibinfo{person}{Zhendong Wang}, \bibinfo{person}{Jianmin
  Bao}, \bibinfo{person}{Wengang Zhou}, \bibinfo{person}{Weilun Wang}, {and}
  \bibinfo{person}{Houqiang Li}.} \bibinfo{year}{2023}\natexlab{a}.
\newblock \showarticletitle{AltFreezing for More General Video Face Forgery
  Detection}. In \bibinfo{booktitle}{\emph{Proc. of {CVPR}}}.
\newblock


\bibitem[Wang et~al\mbox{.}(2004)]%
        {wang2004image}
\bibfield{author}{\bibinfo{person}{Zhou Wang}, \bibinfo{person}{Alan~C Bovik},
  \bibinfo{person}{Hamid~R Sheikh}, {and} \bibinfo{person}{Eero~P Simoncelli}.}
  \bibinfo{year}{2004}\natexlab{}.
\newblock \showarticletitle{Image quality assessment: from error visibility to
  structural similarity}.
\newblock \bibinfo{journal}{\emph{IEEE Transactions on Image Processing}}
  \bibinfo{volume}{13}, \bibinfo{number}{4} (\bibinfo{year}{2004}),
  \bibinfo{pages}{600--612}.
\newblock


\bibitem[Wiles et~al\mbox{.}(2018)]%
        {Wiles18a}
\bibfield{author}{\bibinfo{person}{O. Wiles}, \bibinfo{person}{A.S. Koepke},
  {and} \bibinfo{person}{A. Zisserman}.} \bibinfo{year}{2018}\natexlab{}.
\newblock \showarticletitle{{Self-supervised learning of a facial attribute
  embedding from video}}. In \bibinfo{booktitle}{\emph{British Machine Vision
  Conference}}.
\newblock


\bibitem[Woo et~al\mbox{.}(2012)]%
        {vrcode}
\bibfield{author}{\bibinfo{person}{Grace Woo}, \bibinfo{person}{Andy Lippman},
  {and} \bibinfo{person}{Ramesh Raskar}.} \bibinfo{year}{2012}\natexlab{}.
\newblock \showarticletitle{{VRCodes: Unobtrusive and active visual codes for
  interaction by exploiting rolling shutter}}. In
  \bibinfo{booktitle}{\emph{{IEEE} International Symposium on Mixed and
  Augmented Reality}}.
\newblock


\bibitem[{Xander Elliards}(2024)]%
        {swinney2024}
\bibfield{author}{\bibinfo{person}{{Xander Elliards}}.}
  \bibinfo{year}{2024}\natexlab{}.
\newblock \bibinfo{title}{{AI deepfake video of John Swinney on Sky News goes
  viral on Twitter}}.
\newblock \bibinfo{howpublished}{{The National}}.
\newblock


\bibitem[Xu et~al\mbox{.}(2023)]%
        {xu2023tall}
\bibfield{author}{\bibinfo{person}{Yuting Xu}, \bibinfo{person}{Jian Liang},
  \bibinfo{person}{Gengyun Jia}, \bibinfo{person}{Ziming Yang},
  \bibinfo{person}{Yanhao Zhang}, {and} \bibinfo{person}{Ran He}.}
  \bibinfo{year}{2023}\natexlab{}.
\newblock \showarticletitle{TALL: Thumbnail Layout for Deepfake Video
  Detection}. In \bibinfo{booktitle}{\emph{International Conference on Computer
  Vision}}.
\newblock


\bibitem[Yan et~al\mbox{.}(2023a)]%
        {yan2023ucf}
\bibfield{author}{\bibinfo{person}{Zhiyuan Yan}, \bibinfo{person}{Yong Zhang},
  \bibinfo{person}{Yanbo Fan}, {and} \bibinfo{person}{Baoyuan Wu}.}
  \bibinfo{year}{2023}\natexlab{a}.
\newblock \showarticletitle{{UCF: Uncovering Common Features for Generalizable
  Deepfake Detection}}.
\newblock \bibinfo{journal}{\emph{arXiv preprint arXiv:2304.13949}}
  (\bibinfo{year}{2023}).
\newblock


\bibitem[Yan et~al\mbox{.}(2023b)]%
        {DeepfakeBench_YAN_NEURIPS2023}
\bibfield{author}{\bibinfo{person}{Zhiyuan Yan}, \bibinfo{person}{Yong Zhang},
  \bibinfo{person}{Xinhang Yuan}, \bibinfo{person}{Siwei Lyu}, {and}
  \bibinfo{person}{Baoyuan Wu}.} \bibinfo{year}{2023}\natexlab{b}.
\newblock \showarticletitle{DeepfakeBench: A Comprehensive Benchmark of
  Deepfake Detection}. In \bibinfo{booktitle}{\emph{Advances in Neural
  Information Processing Systems}}.
\newblock


\bibitem[Yang et~al\mbox{.}(2004)]%
        {yang2004hierarchical}
\bibfield{author}{\bibinfo{person}{Zixiang Yang}, \bibinfo{person}{Wei~Tsang
  Ooi}, {and} \bibinfo{person}{Qibin Sun}.} \bibinfo{year}{2004}\natexlab{}.
\newblock \showarticletitle{Hierarchical, non-uniform locality sensitive
  hashing and its application to video identification}. In
  \bibinfo{booktitle}{\emph{IEEE International Conference on Multimedia and
  Expo}}, Vol.~\bibinfo{volume}{1}. \bibinfo{pages}{743--746}.
\newblock


\bibitem[Zhang et~al\mbox{.}(2018b)]%
        {chromacode}
\bibfield{author}{\bibinfo{person}{Kai Zhang}, \bibinfo{person}{Chenshu Wu},
  \bibinfo{person}{Chaofan Yang}, \bibinfo{person}{Yi Zhao},
  \bibinfo{person}{Kehong Huang}, \bibinfo{person}{Chunyi Peng},
  \bibinfo{person}{Yunhao Liu}, {and} \bibinfo{person}{Zheng Yang}.}
  \bibinfo{year}{2018}\natexlab{b}.
\newblock \showarticletitle{{ChromaCode: A Fully Imperceptible Screen-Camera
  Communication System}}. In \bibinfo{booktitle}{\emph{Proc. of {MobiCom}}}.
\newblock


\bibitem[Zhang et~al\mbox{.}(2018a)]%
        {zhang2018unreasonable}
\bibfield{author}{\bibinfo{person}{Richard Zhang}, \bibinfo{person}{Phillip
  Isola}, \bibinfo{person}{Alexei~A Efros}, \bibinfo{person}{Eli Shechtman},
  {and} \bibinfo{person}{Oliver Wang}.} \bibinfo{year}{2018}\natexlab{a}.
\newblock \showarticletitle{The unreasonable effectiveness of deep features as
  a perceptual metric}. In \bibinfo{booktitle}{\emph{Proc. of {CVPR}}}.
\newblock


\bibitem[Zhang et~al\mbox{.}(2023)]%
        {zhang2023sadtalker}
\bibfield{author}{\bibinfo{person}{Wenxuan Zhang}, \bibinfo{person}{Xiaodong
  Cun}, \bibinfo{person}{Xuan Wang}, \bibinfo{person}{Yong Zhang},
  \bibinfo{person}{Xi Shen}, \bibinfo{person}{Yu Guo}, \bibinfo{person}{Ying
  Shan}, {and} \bibinfo{person}{Fei Wang}.} \bibinfo{year}{2023}\natexlab{}.
\newblock \showarticletitle{{SadTalker: Learning Realistic 3D Motion
  Coefficients for Stylized Audio-Driven Single Image Talking Face Animation}}.
  In \bibinfo{booktitle}{\emph{Proc. of {CVPR}}}.
\newblock


\bibitem[Zhang et~al\mbox{.}(2019)]%
        {artifact_det_and_sim}
\bibfield{author}{\bibinfo{person}{Xu Zhang}, \bibinfo{person}{Svebor Karaman},
  {and} \bibinfo{person}{Shih-Fu Chang}.} \bibinfo{year}{2019}\natexlab{}.
\newblock \showarticletitle{{Detecting and Simulating Artifacts in GAN Fake
  Images}}. In \bibinfo{booktitle}{\emph{IEEE International Workshop on
  Information Forensics and Security}}.
\newblock


\bibitem[Zhao et~al\mbox{.}(2021)]%
        {multiattentional_nn}
\bibfield{author}{\bibinfo{person}{Hanqing Zhao}, \bibinfo{person}{Wenbo Zhou},
  \bibinfo{person}{Dongdong Chen}, \bibinfo{person}{Tianyi Wei},
  \bibinfo{person}{Weiming Zhang}, {and} \bibinfo{person}{Nenghai Yu}.}
  \bibinfo{year}{2021}\natexlab{}.
\newblock \showarticletitle{{Multi-attentional Deepfake Detection}}. In
  \bibinfo{booktitle}{\emph{Proc. of {CVPR}}}.
\newblock


\bibitem[Zhou and Lim(2021)]%
        {zhou2021joint}
\bibfield{author}{\bibinfo{person}{Yipin Zhou} {and} \bibinfo{person}{Ser-Nam
  Lim}.} \bibinfo{year}{2021}\natexlab{}.
\newblock \showarticletitle{Joint audio-visual deepfake detection}. In
  \bibinfo{booktitle}{\emph{Proc. of {CVPR}}}.
\newblock


\end{thebibliography}

\appendix
\renewcommand{\thesubsection}{A.\arabic{subsection}}

\section*{Appendix}
\label{sec:appendix}

\subsection{Perceptibility Evaluation Details}
\label{subsec:extra_perc_eval}

Here, we detail the setup of our perceptibility user study and LPIPS evaluation and further discuss the studies' results.

\para{User study}
We invite groups of 9 participants (recruited via email, with ages ranging from 18 to 34) to each of the test environments (Figure~\ref{fig:surfaces}). At each environment participants were asked to assess the projection surface during four 45~s trials in which the core unit either performed embedding of a random bitstream or was powered off as a control scenario (done for two randomly selected trials). Participants were informed that light may be projected and shown the projection region boundaries. During each trial they were allowed to walk freely up to 1.5~m of the surface. They were asked to answer two questions once ready:
\begin{itemize}
     \item[\emph{Q1}] \emph{Do you believe the light pattern is present in the video?(Y/N)}
     \item[\emph{Q2}] \emph{How obtrusive do you find the pattern? (Low/Medium/High Obtrusiveness)}. 
\end{itemize}
Finally, we invite 20 participants (recruited via email, ages 18-65) to assess \emph{videos} of the projection surface. For each video, the boundaries of the projection region were marked to enable participants to accurately examine it. Participants were permitted to freely zoom into any portion of frames, pause, and rewind.

\begin{figure}[hbt!]
    \includegraphics[width=\columnwidth]{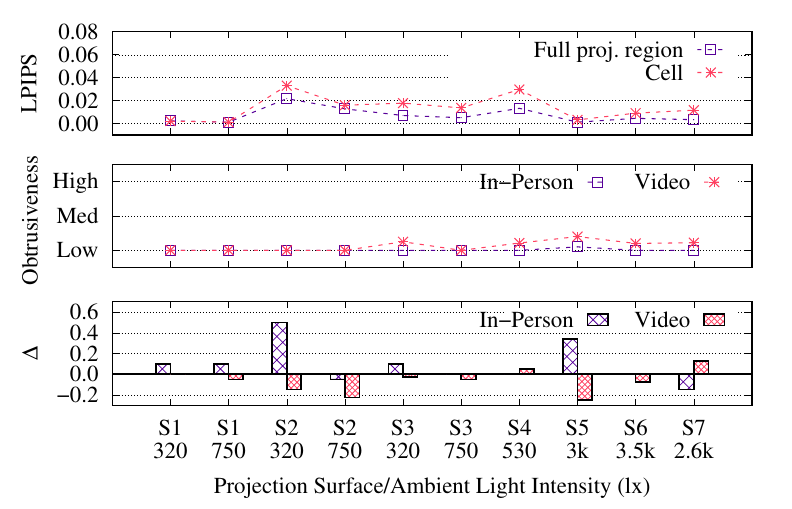}
    \caption{Embedding perceptibility results. Top: Mean LPIPS for pairs of video frames captured w/ and w/o \name\ operation. All scores are significantly below the established LPIPS perceptibility threshold of 0.5~\cite{laidlaw2020perceptual}.
    Middle: mean obtrusiveness reported by users study participants. Bottom: user study $\Delta = TPR - FPR$, normalizing true positive identifications of embedding occurrence to false positive identifications during control trials. A nonpositive value indicates imperceptibility.}
    \label{fig:perceptibility}
\end{figure}

As shown in the bottom two figures of Figure~\ref{fig:perceptibility}, for each environment we plot (1) $\Delta = TPR - FPR$, where TPR is the rate at which participants responded "Yes" to Q1 when embedding was occurring, and FPR is the rate at which they incorrectly responded "Yes"; and (2) the average response to Q2, excluding those given with a Q1 false positive. A $\Delta$ value $\leq 0$ indicates that embedding is effectively imperceptible, as participants perform no better than random at detecting it. We observe this in all but two video cases. In person, $\Delta \leq 0.2$ in all but two environments. Respondents uniformly report low obtrusiveness in video and live. The primary factors influencing perceptibility are the surface's texture and color. Darker, homogeneous surfaces (i.e., S2, S4, S5) fundamentally contrast with impinging light, whereas brighter-colored, more complex ones (e.g, S1, S3, S5) provide a camouflage. Increasing ambient light intensity can counteract this (as with S4 and S2 at 750~lx) by increasing the baseline brightness of a surface's appearance.

\para{Perceptual metrics} We additionally evaluate optical signature perceptibility in video using the learned perceptual loss (LPIPS)~\cite{zhang2018unreasonable, pytorchlightning_lpips}  metric. This metric has been shown to have superior correlation with human perception, especially in the context of fine-grained changes such as those introduced by \name's optical modulations. LPIPS takes as input two images and outputs a score ranging from 0 to 1, where a lower value indicates the inputs are more perceptually similar. In particular, prior works show humans cannot sense differences when LPIPS is below 0.5~\cite{laidlaw2020perceptual}. To apply these image-level metrics to our videos, we compute the mean LPIPS between crops of the full projection region and individual cells in 5,000 pairs of frames captured with and without \name\ operating. We compute scores at both the full projection region and cell level to understand perceptibility at both scales viewers may notice differences.

The top panel of Figure~\ref{fig:perceptibility} shows that full region and cell-level scores are highly correlated. Further, all LPIPS scores are over ten times lower than then established LPIPS perceptibility threshold. 

\begin{table*}[t!]
    \small
    \addtolength{\tabcolsep}{-0.4em}
     \begin{tabularx}{\textwidth}{@{}p{1.6cm}|YYYYYYYYYYYY@{}}
        \hlineB{2}
        \vspace{0.25em}
        & \multicolumn{11}{c}{\bf{Detector}} \\
          \bf{\% Window\newline Modified} &  
         Meso4\newline\cite{afchar2018mesonet}&
         Xception\newline\cite{rossler2019faceforensics++} & Capsule\newline\cite{nguyen2019capsule} & Efficient\newline\cite{tan2019efficientnet}  & \phantom{a}SRM\newline\cite{luo2021generalizing} & \phantom{a}SPSL\newline\cite{liu2021spatial}  &  
         \phantom{a}Recce\newline\cite{cao2022end} & 
         \phantom{a}UCF\newline\cite{yan2023ucf}  & 
         \phantom{a}TALL\newline\cite{xu2023tall}  & 
         AltFreeze.\newline\cite{Wang_2023_CVPR} & 
         \bf{Ours}\\
         \hline
        10-20 & 0.53 & 0.59 & 0.57 & 0.57 & 0.55 & 0.55  & 0.55 & 0.60 & 0.55 & 0.47 & \bf{0.72}\\
        20-30 & 0.58 & 0.62 & 0.55 & 0.58 & 0.55 & 0.58 & 0.55 & 0.62 & 0.57 & 0.47 & \bf{0.84}\\
        30-40 & 0.60 & 0.64 & 0.58 & 0.60 & 0.58 & 0.60 & 0.56 & 0.63 & 0.58 & 0.47 & \bf{0.90}\\
        40-50 & 0.59 & 0.67 & 0.59 & 0.63 & 0.58 & 0.61 & 0.59 & 0.66 & 0.61 & 0.48 & \bf{0.92}\\
        100 & 0.75 & 0.90 & 0.80 & 0.81 & 0.76 & 0.76 & 0.83 & 0.84 & 0.84 & 0.26 & \bf{0.98}\\
        \hlineB{2}
    \end{tabularx}
    \caption{Comparison of AUC scores achieved by passive detectors and \name's 150-bit dynamic feature hashes on our multi-posed, fine-grained reenactment dataset (\S\ref{subsec:digest_robustness}), in which modifications of various granularities (by percentage duration modified) are applied to each 4.5~s window. Best performing method is bolded.}
    \label{tab:dyn_aucs}
\end{table*}

\subsection{Adversarial Deepfake Model Training}
\label{subsec:adv_training}
Here we provide details on the training, implementation, and evaluation of our DaGAN and FSGAN models. 

\para{DaGAN} The DaGAN reenactment model forward pass takes as input two face images: a source image of the victim  and an attacker-provided "driving" image. It synthesizes a fake image where the victim possesses a new facial expression and pose, matching those in the driving image. A video can be formed by performing a forward pass of the model for each pair of frames in an original video and driving video.

During training of our modified DaGAN model, we extract the 16 FaceMesh features considered by \name's dynamic feature vectors (\S\ref{subsec:feature_extraction}) from each \emph{source} and generated \emph{fake} image. Our loss function $L_{adv}$ is then defined as follows:
\begin{equation*}
    L_{adv} = L + \alpha\Theta(Dyn_{src}, Dyn_{fake})
\end{equation*}
where $L$ is the original DaGAN loss function, $Dyn \in \mathbb{R}^{16}$ are vectors containing the FaceMesh results, and $\Theta$ is the cosine similarity function. The coefficient $\alpha$ weights the importance of our adversarial spoofing objective.

To implement this modified version of DaGAN, we first develop a fully differentiable PyTorch implementation of FaceMesh.\footnote{We release our MediaPipe FaceMesh implementation with our other code, as linked on our project website} This is needed because FaceMech is only distributed as a LiteRT module, which is only designed for inference and does not have the necessary mechanisms for backpropagation. 
We use ONNX-converted~\cite{onnx} ports of each of the three neural networks underlying FaceMesh and integrate them to the best of our abilities by referencing MediaPipe's public model cards~\cite{mediapipe}. Our implementation outputs facial landmarks with an average difference of 2 pixels and blendshape scores with an average difference of  0.11 (arbitrary units, ranging from 0 to 1) from their official implementation counterparts when run on DaGAN's training dataset. 

We train our adversarial DaGAN model by fine-tuning with our loss component added. We start training from the checkpoint released by the authors~\cite{dagan_imp}, applying early stopping based on validation loss. We use the same learning rate and parameters employed in their original implementation, as well as their same training data, sourced from VoxCeleb. We create evaluation videos by randomly choosing 55 pairs of videos from the VoxCeleb test split and using one to drive the other.

\para{FSGAN} The FSGAN identity swap model takes as input a source image of the victim and an attacker-provided target face image. The model synthesizes an image in which the victim's face is supplanted with the target's face, effectively modifying the portrayed identity. 

During training of our modified FSGAN models, we extract the ArcFace embedding from both the source and generated fake images. Similar to the case of DaGAN, our loss function $L_{adv}$ is then defined as follows:
\begin{equation*}
    L_{adv} = L + \alpha\Theta(Arc_{src}, Arc_{fake})
\end{equation*}
where $L$ is the original FSGAN loss function, $Arc \in \mathbb{R}^{512}$ are the extracted ArcFace embeddings, and $\Theta$ is the cosine similarity function. The coefficient $\alpha$ again weights the importance of our adversarial spoofing objective.

We train the FSGAN model from scratch using data from the VoxCeleb dataset, because the authors do not release all weights necessary for fine-tuning. We use the same learning parameters employed in the original version and apply early stopping based on validation loss. For testing, we randomly choose 55 pairs of videos from the VoxCeleb dataset (ensuring they portray different identities) and generate an identity swap deepfake for each.

\subsection{Proofs and Definitions: Locality-Sensitive Hashing}
\label{subsec:lsh_proofs}
Here we formalize our approach to using cosine similarity-based LSH for verification. We include both definitions and discussions of the methodology (Definition \ref{def:lsh_ver}) and derive an equation for the relationship between hash size and verification performance (Theorem \ref{theorem:k_perf}). This equation illustrates that verification performance has no dependence on input vector dimensionality.

\begin{definition}[Cosine Similarity LSH Scheme]
\label{def:lsh_cos}
\end{definition}

Recall from \S4.1 that the LSH scheme for cosine similarity, $H_{cos}$, outputs $k$-bit hashes such that $D(H_{cos}(\vec{u}), H_{cos}(\vec{v})$ estimates $\Theta(\vec{u}, \vec{v})$. $\Theta(\vec{u}, \vec{v})$ is the angle between $\vec{u}$ and $\vec{v}$, and $D$ is the Hamming distance. As defined in \cite{charikar2002similarity}, $H_{cos}$ is composed of multiple \emph{random projection} hash functions $h$, each outputting one bit.

Specifically, let $\vec{r}$ be a vector in $\mathbb{R}^n$ chosen randomly from the $n$-dimensional Gaussian distribution (i.e., each coordinate is drawn from a Gaussian distribution). Let the hash function $h_{\vec{r}} : \vec{u} \in \mathbb{R}^n \mapsto \{0, 1\}$ be defined as follows:
\begin{equation*}
   h_{\vec{r}}(\vec{u}) =
  \begin{cases}
        1 & \text{if $\vec{r} \cdot \vec{u} \geq 0$ } \\
        0 & \text{if $\vec{r} \cdot \vec{u} < 0$ } 
  \end{cases}
\end{equation*}

The locality sensitive hashing scheme $H$ is defined as $H(\vec{u}) = \{h_1(\vec{u}), h_2(\vec{u}), ..., h_k(\vec{u})\}$, where $h_i$  are independently and randomly chosen hash functions of the form above. Thus, given an input $\vec{u} \in \mathbb{R}^n$, $H$ outputs a $k$-bit vector, formed by concatenating the single bit outputs of each of its hash functions $h_i$.

The key idea of this scheme is that the sign of a vector $\vec{u}$'s projection onto $\vec{r}_i$ is fundamentally related to the angle between $\vec{u}$ and $\vec{r}_i$. Thus if $\vec{u}$ and another vector $\vec{v}$ have a high cosine similarity, their projections onto $\vec{r}_i$ are more likely to have the same sign. Each bit in the hash is an additional instance of this process and “sample” to aid in approximating $\Theta$; including more bits increases the probability that $D$ correctly reflects $\Theta$. This captures the cosine similarity with extreme space efficiency.

\begin{definition}[Verification using LSH]
\label{def:lsh_ver}
\end{definition}
We depart from the definition of a traditional verification problem: given two feature vectors, we would like to confirm that they represent the same source (e.g., face embeddings corresponding to the same identity). Two feature vectors $u, v$ are said to correspond to the source if $\Theta(\vec{u}, \vec{v}) \leq \theta_{th}$, where $\theta_{th}$ is a decision threshold. Otherwise, the vectors are said to correspond to different sources.

We can similarly formalize the verification problem on hashed feature vectors: the hashes of two feature vectors, $H_{cos}(u), H_{cos}(v)$  are said to correspond to the same source if $D(H(\vec{v}), H(\vec{u})) \leq d_{th}$, where $d_{th}$ is a decision threshold for verification on hashed vectors. If $\theta_{th}$ is the optimal decision threshold for verification on the raw feature vectors, intuitively, the optimal value of $d_{th}$ should be the \emph{expected value} of $D(H(\vec{v}), H(\vec{u}))$  for two vectors $u, v$ s.t. $\Theta(\vec{u}, \vec{v}) = \theta_{th}$. From Theorem \ref{theorem:lsh_expval} below, this is $\frac{k\theta_{th}}{\pi}$.

\begin{theorem}[Expected Value of Hamming Distance]
\label{theorem:lsh_expval}
Let $H$ be a locality sensitive hashing scheme defined according to Definition \ref{def:lsh_cos} and $D$ be the Hamming distance function. The expected value of $D(H(\vec{v}), H(\vec{u}))$  is $\frac{k\Theta(\vec{u}, \vec{v})}{\pi}$.
\end{theorem}

\begin{proof}
Recall that $D$, the Hamming distance function, gives the number of positions at which the values of two bitstrings differ. Since the value of each bit of $H$'s output is determined by a hash function $h_i$, the expected value of $D(H(\vec{v}), H(\vec{u}))$ is the expected number of $H$'s $k$ hash functions $h_i$ for which  $h_i(\vec{u}) \ne h_i(\vec{v})$. From \cite{charikar2002similarity}, 
\begin{equation*}
    \bold{Pr}[h(\vec{u}) \ne h(\vec{v})] = \frac{\Theta(\vec{u}, \vec{v})}{\pi}.
\end{equation*} 
Therefore, the probability that $D(H(\vec{v}), H(\vec{u})) = n$, (i.e., the outputs of exactly $n$ of the hash functions $h_i$ differ) is 
\begin{equation}\label{eq:idiffer}
\begin{aligned}
    {k \choose n} \bold{Pr}[h(\vec{u}) \ne h(\vec{v})]^n \bold{Pr}[h(\vec{u}) = h(\vec{v})]^{n-k} \\
    = {k \choose n}\left(\frac{\Theta(\vec{u}, \vec{v})}{\pi}\right)^n(1 - \frac{\Theta(\vec{u}, \vec{v})}{\pi})^{k-n}   
\end{aligned}
\end{equation}
The expected value of $D(H(\vec{v}), H(\vec{u}))$ is thus
\begin{equation}
    \sum_{n=0}^k n{k \choose n}\left(\frac{\Theta(\vec{u}, \vec{v})}{\pi}\right)^n\left(1 - \frac{\Theta(\vec{u}, \vec{v})}{\pi}\right)^{k-n} = \frac{k\Theta(\vec{u}, \vec{v})}{\pi} 
\end{equation}
\end{proof}

\begin{theorem}[Impact of $k$ on Verification Performance]\label{theorem:k_perf} \emph{To assess the impact of the hash size $k$ on verification performance, we seek the probability $P_{\theta_{th}}(k)$ that \emph{all} decisions obtained from verification on $k$-bit hashed feature vectors are the \emph{same} as those obtained from performing verification on the raw vectors with a decision threshold of $\theta_{th}$. This event indicates that the hashed feature vectors perfectly preserve verification performance. Thus we can view the probability of its occurrence as a measure of the hashed vectors' performance relative to that of the raw vectors.} $P_{\theta_{th}}(k)$ is given by the equation
\begin{equation*}
\begin{aligned}
P_{\theta_{th}}(k) = \exp\left(\int_{0}^{\theta_{th}} \ln \left( \sum_{n=0}^{\frac{k\theta_{th}}{\pi}} {k \choose n}\left(\frac{\theta}{\pi}\right)^n\left(1 - \frac{\theta}{\pi}\right)^{k-n}\right)d\theta\right)  * \\ 
\exp\left(\int_{\theta_{th}}^{\pi} \ln \left( \sum_{n=\frac{k\theta_{th}}{\pi}}^{k} {k \choose n}\left(\frac{\theta}{\pi}\right)^n\left(1 - \frac{\theta}{\pi}\right)^{k-n}\right)d\theta\right)
\end{aligned}
\end{equation*}
\end{theorem}

\begin{figure}[t!]
    \includegraphics[width=0.9\columnwidth]{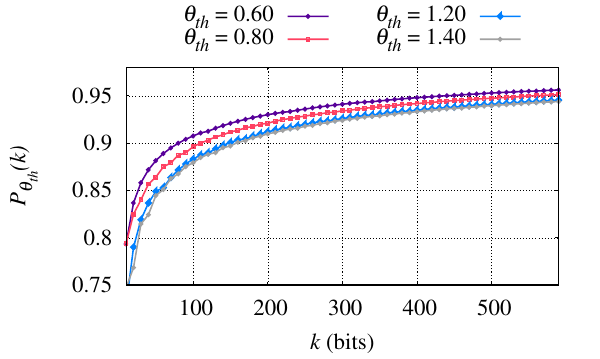}
    \vspace{-0.15in}
    \caption{$\bm{P_{\theta_{th}}(k)}$ (derived in Theorem 2) plotted for various cosine similarity decision thresholds $\bm{\theta_{th}}$. Our results are obtained using a cosine similarity threshold of 1.17 for our raw dynamic feature vectors and 0.88 for our ArcFace identity feature vectors (\S4.2). Notably, all curves' elbows occur before our chosen hash size $\bm{k=150}$.}
    \label{fig:pk}
\end{figure}

Evidently, none of the equation's terms depend on the dimensionality of the input vectors. $P_{\theta_{th}}(k)$ is only a function of the decision threshold $\theta_{th}$ and $k$.

\begin{proof}

Based on Definition \ref{def:lsh_ver}, verification decisions are obtained from raw or hashed feature vectors using the criteria $\Theta(\vec{u}, \vec{v}) \leq \theta_{th}$ or $D(H(\vec{v}), H(\vec{u})) \leq \frac{k\theta_{th}}{\pi}$, respectively. Thus we have
\begin{equation*}
\begin{aligned}
P_{\theta_{th}}(k) = \bold{Pr}[\hspace{6cm}\\
\forall (\vec{u}, \vec{v}) \in \{(\vec{u}, \vec{v}) : \Theta(\vec{u}, \vec{v}) \leq \theta_{th}\},  \hspace{0.3cm} D(H(\vec{v}), H(\vec{u})) \leq \frac{k\theta_{th}}{\pi}  \\
 \cap\hspace{4cm} \\
\hspace{-1cm}\forall (\vec{u}, \vec{v}) \in \{(\vec{u}, \vec{v}) : \Theta(\vec{u}, \vec{v}) > \theta_{th}\} \hspace{0.3cm} D(H(\vec{v}), H(\vec{u})) > \frac{k\theta_{th}}{\pi}\\ 
]\hspace{6cm}
\end{aligned}
\end{equation*}

Intuitively,
\begin{equation*}
\bold{Pr}[ \forall (\vec{u}, \vec{v}) \in \{(\vec{u}, \vec{v}) : \Theta(\vec{u}, \vec{v}) \leq \theta_{th}\},  \hspace{0.3cm} D(H(\vec{v}), H(\vec{u})) \leq \frac{k\theta_{th}}{\pi}]
\end{equation*} 
is the probability that for all $\vec{u}, \vec{v}$ satisfying $\Theta(\vec{u}, \vec{v}) \leq \theta_{th}$, \emph{at most} $\frac{k\theta_{th}}{\pi}$ bits of $H(\vec{u})$ and $H(\vec{v})$ differ.

Using Equation \ref{eq:idiffer} and the independence of each comparison of $(\vec{u}, \vec{v}) \in \{(\vec{u}, \vec{v}) : \Theta(\vec{u}, \vec{v}) \leq \theta_{th}\}$, 
\begin{equation*}
\begin{aligned}
\bold{Pr}[ \forall (\vec{u}, \vec{v}) \in \{(\vec{u}, \vec{v}) : \Theta(\vec{u}, \vec{v}) \leq \theta_{th}\},  \hspace{0.3cm} D(H(\vec{v}), H(\vec{u})) \leq \frac{k\theta_{th}}{\pi}] \\
= \exp\left(\int_{0}^{\theta_{th}} \ln \left( \sum_{n=0}^{\frac{k\theta_{th}}{\pi}} {k \choose n}\left(\frac{\theta}{\pi}\right)^n\left(1 - \frac{\theta}{\pi}\right)^{k-n}\right)d\theta\right)
\end{aligned}
\end{equation*}

By the same logic, 
\begin{equation*}
\begin{aligned}
\bold{Pr}[ \forall (\vec{u}, \vec{v}) \in \{(\vec{u}, \vec{v}) : \Theta(\vec{u}, \vec{v}) > \theta_{th}\},  \hspace{0.3cm} D(H(\vec{v}), H(\vec{u})) > \frac{k\theta_{th}}{\pi}]  \\
= \exp\left(\int_{\theta_{th}}^{\pi} \ln \left( \sum_{n=\frac{k\theta_{th}}{\pi}}^{k} {k \choose n}\left(\frac{\theta}{\pi}\right)^n\left(1 - \frac{\theta}{\pi}\right)^{k-n}\right)d\theta\right)
\end{aligned}
\end{equation*}

We thus have:

\begin{equation*}
\begin{aligned}
P_{\theta_{th}}(k) = \exp\left(\int_{0}^{\theta_{th}} \ln \left( \sum_{n=0}^{\frac{k\theta_{th}}{\pi}} {k \choose n}\left(\frac{\theta}{\pi}\right)^n\left(1 - \frac{\theta}{\pi}\right)^{k-n}\right)d\theta\right)  * \\ 
\exp\left(\int_{\theta_{th}}^{\pi} \ln \left( \sum_{n=\frac{k\theta_{th}}{\pi}}^{k} {k \choose n}\left(\frac{\theta}{\pi}\right)^n\left(1 - \frac{\theta}{\pi}\right)^{k-n}\right)d\theta\right)
\end{aligned}
\end{equation*}

\end{proof}

\end{document}